\documentclass{article}

\usepackage[preprint]{icml2026}

\usepackage[table]{xcolor}
\usepackage{colortbl}
\usepackage{booktabs}
\usepackage{multirow}
\usepackage{enumitem}

\usepackage{algorithm}

\usepackage{amsmath,amssymb,mathtools,amsthm}

\usepackage{hyperref}
\usepackage[capitalize,noabbrev]{cleveref}

\DeclareUnicodeCharacter{00A0}{ }

\theoremstyle{plain}

\theoremstyle{definition}

\theoremstyle{remark}

\usepackage[textsize=tiny]{todonotes}

\icmltitlerunning{Order Is Not Layout: Order-to-Space Bias in Image Generation}

\begin{document}

\twocolumn[
  \icmltitle{Order Is Not Layout: Order-to-Space Bias in Image Generation}



  \icmlsetsymbol{equal}{*}

  \begin{icmlauthorlist}
    \icmlauthor{Yongkang Zhang}{equal,sch,sch1}
    \icmlauthor{Zonglin Zhao}{equal,sch,sch2}
    \icmlauthor{Yuechen Zhang}{equal,sch,sch3}
    \icmlauthor{Fei Ding}{sch4}
    \icmlauthor{Pei Li}{sch1}
    \icmlauthor{Wenxuan Wang}{sch}
  \end{icmlauthorlist}

  \icmlaffiliation{sch}{Renmin University of China, China}
  \icmlaffiliation{sch1}{Huazhong Agricultural University, China}
  \icmlaffiliation{sch2}{Huazhong University of Science and Technology, China}
  \icmlaffiliation{sch3}{Jiangnan University, China}
  \icmlaffiliation{sch4}{Nanchang University, China}
  
  \icmlcorrespondingauthor{Wenxuan Wang}{wangwenxuan@ruc.edu.cn}

  \icmlkeywords{Machine Learning, ICML}

  \vskip 0.3in
]



 \printAffiliationsAndNotice{\icmlEqualContribution}

\begin{abstract}
We study a systematic bias in modern image generation models: the \emph{mention order of entities} in text spuriously determines spatial layout and entity--role binding. We term this phenomenon \textit{Order-to-Space Bias (OTS)} and show that it arises in both text-to-image and image-to-image generation, often overriding grounded cues and causing incorrect layouts or swapped assignments.
To quantify OTS, we introduce \textsc{OTS-Bench}, which isolates order effects with paired prompts differing only in entity order and evaluates models along two dimensions: \emph{homogenization} and \emph{correctness}. Experiments show that Order-to-Space Bias (OTS) is widespread in modern image generation models, and provide evidence that it is primarily data-driven and manifests during the early stages of layout formation. Motivated by this insight, we show that both targeted fine-tuning and early-stage intervention strategies can substantially reduce OTS, while preserving generation quality.
\end{abstract}

\section{Introduction}
\begin{figure}[t]
  \centering
  \includegraphics[width=\columnwidth]{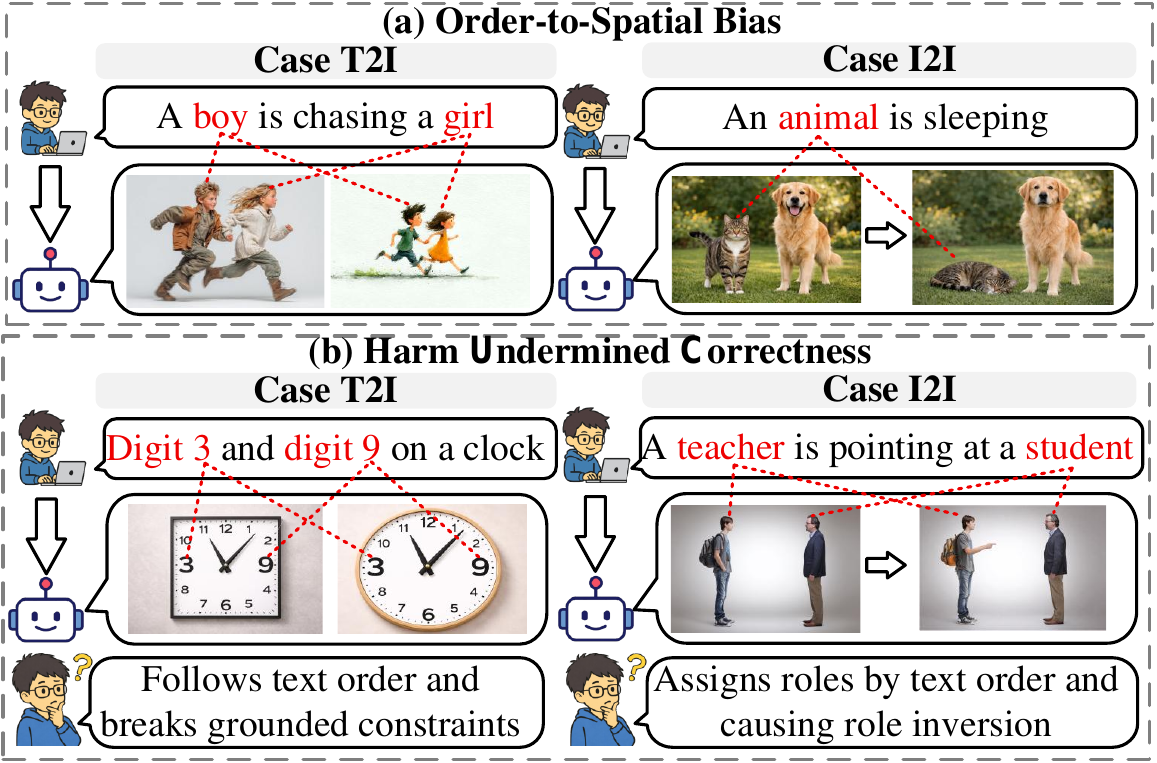}
  \caption{\textbf{Order-to-Space Bias} makes models treat mention order as a cue for spatial layout and role binding.
(a) With neutral prompts, the first-mentioned entity tends to appear on the left, and under-specified edits are often applied to a default side.
(b) When order conflicts with grounded cues, the model follows order instead, producing incorrect layouts or role inversions.}
  \label{fig:introduction}
\end{figure}

The rapid progress of text-to-image (T2I) and image-to-image (I2I) diffusion models has fundamentally transformed modern image synthesis, with models like DALL-E 3~\cite{betker2023improving}, Midjourney v7~\cite{midjourney2025}, Stable Diffusion 3~\cite{esser2024scaling}, and NanoBanana~\cite{geminiFlashImage2025} now powering millions of daily creative workflows. These diffusion-based generative systems synthesize images by iterative denoising~\cite{ho2020denoising}, achieving highly diverse generations and photorealistic quality. However, beneath this impressive capability lies a concerning systematic flaw: \textit{Do models follow grounded layout/identity cues, or do they follow text order—placing the first-mentioned entity on the left and binding roles/actions accordingly?}

As illustrated in Figure~\ref{fig:introduction}(a), modern generative models exhibit a simple but persistent pattern: they often treat the order of mentions in the prompt as a left-to-right layout rule.
In T2I, prompts with no positional language (e.g., ``a boy is chasing a girl'') are nevertheless rendered with a consistent spatial ordering, where the first-mentioned entity appears on the left and the second on the right.
In I2I, a related effect emerges when the prompt is under-specified (e.g., ``an animal is sleeping'') and the input image contains two animals: models tend to attach the requested attribute or action to the left entity by default.
We term this phenomenon \textbf{Order-to-Space Bias (OTS)}—a systematic failure in which generative models treat textual order as a spatial or semantic assignment rule—mapping “A and B” to a fixed left–right layout, or binding attributes and actions according to mention order rather than grounded visual or real-world constraints.

While recent work has made significant strides in identifying and mitigating demographic biases, gender stereotypes, and cultural misrepresentations in text-guided image generation models~\cite{bianchi2023easily,luccioni2023stable,struppek2023exploiting,friedrich2023fair}, the spatial reasoning failures induced by OTS remain largely unexplored. Existing compositional generation benchmarks~\cite{huang2023t2i,huang2025t2i} mainly test \textit{whether} entities appear, but often miss \textit{where} they are placed and \textit{why} a particular layout emerges without spatial cues.
Recent spatial reasoning benchmarks (e.g., SPHERE~\cite{zhang2025sphere} and OmniSpatial~\cite{jia2025omnispatial}) systematically evaluate vision--language models on spatial perception and multi-step spatial reasoning across diverse tasks and difficulty levels, but they do not isolate \emph{order} as a generative shortcut---i.e., whether mention order is spuriously mapped to left--right layout or role/action binding under spatially neutral prompts.

This gap is critical because Order-to-Space Bias (OTS) leads to systematically incorrect generation outcomes. 
As illustrated in Figure~\ref{fig:introduction}(b), models produce erroneous spatial layouts when textual order overrides grounded or semantic constraints—for example, placing ``3'' to the left of ``9'' on a clock, despite the expected clockwise arrangement. 
Beyond layout errors, the same bias propagates to more severe semantic failures in image-to-image generation. 
Models often exhibit \textbf{action misattribution}: given an input image and a prompt such as ``\textit{a teacher is pointing at a student}'', actions are assigned according to textual role order rather than visual grounding, resulting in role inversions that violate expected social and semantic relationships.

To quantify Order-to-Space Bias (OTS), we introduce \textsc{OTS-Bench}, a benchmark consisting of 4{,}300 test cases for both text-to-image and image-to-image generation.
The benchmark is constructed from combinations of 138 entities and 172 actions/states, enabling controlled evaluation of order-induced effects across subjects and interactions.
We evaluate OTS along two dimensions: (1) \textit{Homogenization}, which measures the extent to which models follow \textit{prompt order} in left--right layout (T2I) or action/attribute assignment (I2I); and (2) \textit{Correctness}, which assesses whether model generation \textit{respects} real-world constraints (T2I) or \textit{preserves} grounding from the \textit{input image} (I2I), rather than relying on an order-based shortcut.

Our evaluation across nine state-of-the-art models shows that OTS is pervasive.
In both T2I and I2I, model generation often defaults to an order-following layout or role/action binding: under neutral or under-specified prompts, T2I homogenization is often above 70\%, and when mention order contradicts grounding constraints, T2I correctness can drop from $\sim$90\% in Aligned to $\sim$20\% in Reverse.
We further observe a strong first-mentioned-left prior in web-scale caption--image data, suggesting a largely data-driven origin of OTS.
Moreover, temporal interventions during diffusion sampling localize OTS to the early, layout-forming stage, revealing when order cues dominate generation.
Finally, we show that both generation-time scheduling and targeted fine-tuning can effectively mitigate this bias while preserving image quality.

\noindent\textbf{Contributions.}
(1) We identify \textit{Order-to-Space Bias (OTS)} in modern T2I and I2I generators, where mention order spuriously dictates spatial layout and entity--role/action binding.
(2) We introduce \textsc{OTS-Bench}, a controlled benchmark that isolates order effects with paired prompts and evaluates models along \textit{homogenization} and \textit{correctness}.
(3) We conduct a large-scale evaluation across state-of-the-art models, showing that OTS is pervasive and causes substantial grounding failures.
(4) We trace OTS to data bias and early layout formation, and mitigate it via generation-time scheduling and targeted fine-tuning without degrading image quality.

\section{Related Work}

\noindent\textbf{Text-to-Image and Image-to-Image Models.}
Recent T2I models such as Stable Diffusion 3~\cite{esser2024scaling}, FLUX-dev~\cite{flux2024}, 
and PixArt~\cite{chen2023pixart} leverage diffusion transformers~\cite{peebles2023scalable} 
with vision--language alignment built on CLIP~\cite{radford2021learning} or T5 encoders~\cite{raffel2020exploring}. Instruction-guided 
I2I models such as InstructPix2Pix~\cite{brooks2023instructpix2pix}, ControlNet~\cite{zhang2023adding}, 
and IP-Adapter~\cite{ye2023ip} further enable text-controlled editing of existing images. Both T2I and I2I systems process prompts in a left-to-right fashion, but 
how such sequential encoding introduces systematic spatial behavior during generation has been largely overlooked.

\noindent\textbf{Compositional Generation Benchmarks.}
Most prior benchmarks evaluate compositional correctness under \emph{explicit} spatial specifications.
T2I-CompBench~\cite{huang2023t2i,huang2025t2i} tests 3D relations with detector-based scoring, while GenEval~\cite{ghosh2023geneval} and VQA-style metrics such as TIFA~\cite{hu2023tifa} and VPEval~\cite{cho2023visual} provide broader object-level verification.
Spatial benchmarks such as SPHERE~\cite{zhang2025sphere} and OmniSpatial~\cite{jia2025omnispatial} further diagnose spatial perception and reasoning in structured settings.
In contrast, we study a different failure mode: under \emph{spatially neutral} prompts, models may rely on an order-based shortcut mapping entity mention order to left--right layout or role/action binding.
\begin{figure*}[t]
    \centering
    \includegraphics[width=1\linewidth]{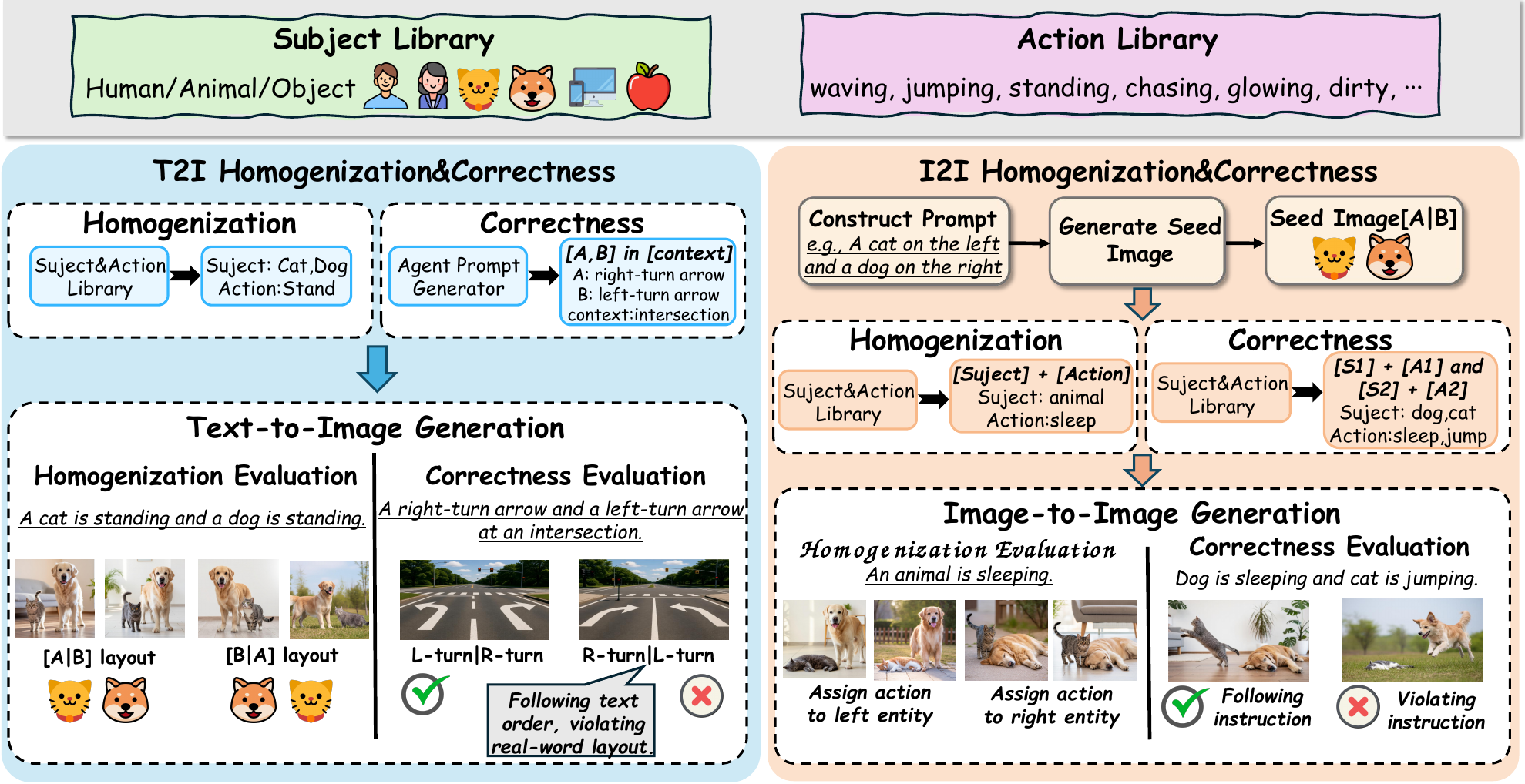}
    \caption{Overview of \textsc{OTS-Bench}. We evaluate OTS in four settings: T2I and I2I, each with homogenization and correctness.}
    \label{fig:overview}
\end{figure*}

\noindent\textbf{Biases in Generative and Vision--Language Models.}
Prior work on generative models primarily targets demographic/cultural biases and fairness evaluation~\cite{friedrich2024multilingual,jha2024visage,wan2024survey,luo2024faintbench,chinchure2024tibet}, while studies on discriminative VLMs examine positional effects such as recency bias and embedding-induced spatial bias~\cite{tian2025identifying,alam2026spatial}. Data analyses (e.g., ConceptMix~\cite{wu2024conceptmix}) further highlight limitations of web captions, including under-specified spatial relations in LAION-5B~\cite{schuhmann2022laion}. However, these lines of work focus on \textit{what} content is generated or \textit{how} positional signals affect recognition, rather than \textit{where} entities are placed or \textit{which} entity receives a role/action during generation. Instead, we study \textbf{Order-to-Space Bias (OTS)}---a generation-time shortcut that maps mention order to spatial layout or role assignment in both T2I and I2I---and link it to ordering asymmetries in caption data, suggesting a data-driven origin.

\section{OTS-Bench}
Figure~\ref{fig:overview} summarizes the design of \textsc{OTS-Bench}: the subject/action libraries and the four settings used to measure homogenization and correctness in T2I and I2I.

\subsection{Benchmark Composition}
As shown in Table~\ref{tab:composition}, \textsc{OTS-Bench} contains 138 subjects and 172 action/state types. Subjects are organized into a three-tier taxonomy: humans (64 across 8 occupations with balanced gender), animals (25 across 5 biological categories by body size), and objects (49 across 7 functional categories).
The action/state inventory includes 35 human actions, 55 animal actions, and 82 object states.
We compile the inventories by referencing existing datasets and lexicons, and consolidate categories to ensure consistency and coverage.
Full lists of entities, actions, and grounding references are provided in Appendix~\ref{app:entity_taxonomy}.

\begin{table}[h]
\centering
\caption{OTS-Bench taxonomy overview. Subject and action/state inventories are curated with reference to COCO~\cite{lin2014microsoft}, ImageNet~\cite{deng2009imagenet}, OpenImages~\cite{kuznetsova2020open}, AVA~\cite{gu2018ava}, VerbNet~\cite{schuler2005verbnet}, Animal Kingdom~\cite{ng2022animal}, and Something-Something~\cite{goyal2017something}.}

\label{tab:composition}
\small
\begin{tabular}{lrl}
\toprule
\textbf{Component} & \textbf{Count} & \textbf{Source} \\
\midrule
\textit{Subjects} & & \\
Humans  & 64 & SOC 2018 \\
Animals & 25 & COCO, ImageNet \\
Objects & 49 & COCO, OpenImages \\
\cmidrule(lr){1-3}
\textit{Subjects Total} & \textbf{138} & \\
\midrule
\textit{Actions \& States} & & \\
Human actions  & 35 & AVA, VerbNet \\
Animal actions & 55 & Animal Kingdom \\
Object states  & 82 & Something-Something \\
\cmidrule(lr){1-3}
\textit{Actions \& States Total} & \textbf{172} & \\
\midrule
\end{tabular}
\end{table}

\begin{table}[t]
\centering
\caption{\textsc{OTS-Bench} summary.}
\label{tab:case_stats}
\small
\setlength{\tabcolsep}{6pt}
\begin{tabular}{lrrr}
\toprule
\textbf{Modality} & \textbf{Homogenization} & \textbf{Correctness} & \textbf{Total} \\
\midrule
T2I & 1{,}700 & 400     & 2{,}100 \\
I2I & 1{,}100 & 1{,}100 & 2{,}200 \\
\midrule
\textbf{Total} & 2{,}800 & 1{,}500 & \textbf{4{,}300} \\
\bottomrule
\end{tabular}
\end{table}

\subsection{Benchmark Construction}

\noindent
\textsc{OTS-Bench} contains \textbf{4,300} test cases spanning two modalities (T2I/I2I) and two evaluation dimensions (homogenization/correctness), summarized in Table~\ref{tab:case_stats}.
The full breakdown by entity-pair type is deferred to Appendix~\ref{app:case_stats_full}.
\label{sec:benchmark_construction}

\subsubsection{Text-to-Image Benchmark}

\paragraph{Homogenization Evaluation.}

We test whether models default to placing the first-mentioned entity on the left and the second on the right, even when the prompt contains no spatial cues.
For example, for ``\textit{a cat is standing and a dog is standing}'', a low-homogenization model should generate both
$[\textit{cat}|\textit{dog}]$ and $[\textit{dog}|\textit{cat}]$ layouts with comparable frequency, whereas an order-biased model
consistently produces $[\textit{cat}|\textit{dog}]$.

\paragraph{Correctness Evaluation.}
We test whether models follow grounded real-world left--right constraints when they conflict with order-implied mappings.
For each constrained pair, we evaluate two matched prompt variants:
\textbf{Aligned (Ali)}, where mention order matches the real-world convention, and
\textbf{Reverse (Rev)}, where mention order is swapped while the intended convention remains unchanged.
This design isolates reliance on textual order as the only difference between the two variants.
For example (Figure~\ref{fig:overview}), at an intersection the correct layout is \textit{left-turn$|$right-turn}:
\textbf{Ali} uses ``left-turn then right-turn'', whereas \textbf{Rev} swaps the order (``right-turn then left-turn'').we then check whether the generated layout still matches the grounded convention.

\begin{figure}[t]
    \centering
    \includegraphics[width=\linewidth]{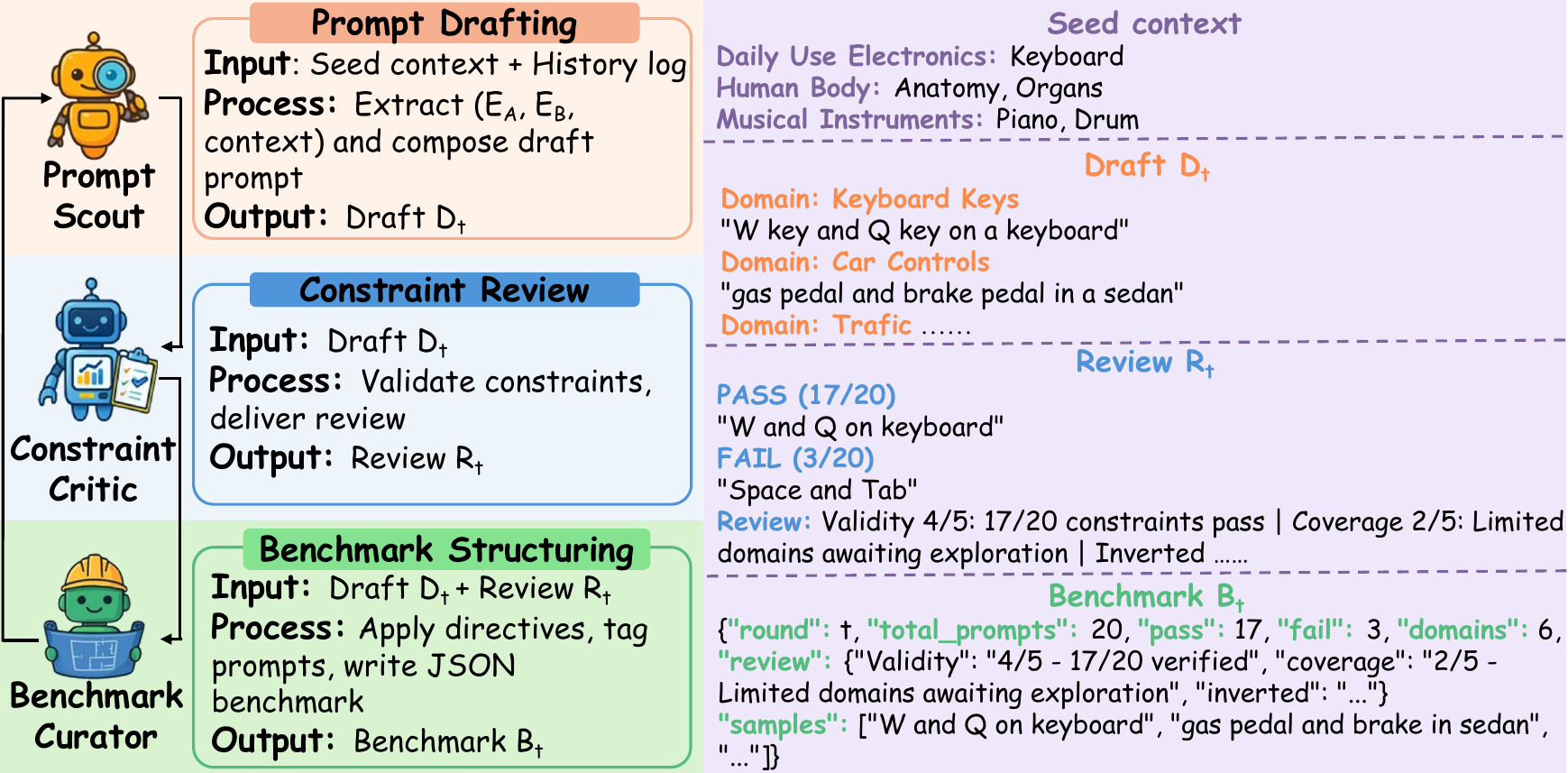}
    \caption{Multi-agent pipeline for generating constrained test cases. Task Explorer proposes candidates, Insight Synthesizer validates real-world constraints, and Diversity Architect produces structured data.}
    \label{fig:agent_workflow}
\end{figure}

We construct 400 such constrained pairs using a multi-agent pipeline (Figure~\ref{fig:agent_workflow}) that proposes candidates and validates verifiable left--right conventions, spanning functional constraints, temporal/positional conventions, and conventional arrangements.

\subsubsection{Image-to-Image Benchmark}
\paragraph{Homogenization Evaluation.}
We test whether models default to a consistent action assignment
when the generation instruction is under-specified.
We first generate a seed image with explicit left--right grounding (e.g., ``\textit{a cat on the left and a dog on the right}''),
then apply an instruction that does not specify the acting entity (e.g., ``\textit{one animal is sleeping}''). 
A low-homogenization model should assign the action to either entity with similar frequency, whereas a biased model consistently assigns it to one side (typically the left/first entity).

\paragraph{Correctness Evaluation.}
We test whether models follow the prompt’s entity--action bindings \emph{without swapping which entity receives which action}.
We start from a seed image with two clearly separated subjects (e.g., a \emph{cat} on the left and a \emph{dog} on the right).
We then use a two-entity prompt that assigns different actions to the two subjects
(e.g., ``\textit{dog is sleeping and cat is jumping}'').
The output is correct only if the \emph{dog} exhibits ``sleeping'' and the \emph{cat} exhibits ``jumping''.

To isolate OTS from general I2I generation difficulty, we evaluate each test case under two matched variants.
In \textbf{Aligned}, the prompt lists entities in the same left--right identity order as in the seed image (left entity first, right entity second);
in \textbf{Reverse}, we flip the entity order in the prompt while keeping the intended entity--action bindings unchanged.
We include (1) non-interactive actions where entities act independently, and (2) directed interactions that specify who acts on whom.

\section{Evaluation Protocol} 
We assess Order-to-Space Bias in two settings: T2I generation and I2I generation, along two evaluation dimensions: \emph{homogenization} and \emph{correctness}.
For each setting, we generate outputs from prompts, verify entity presence and left--right clarity, and label outcomes according to the corresponding criteria.
Homogenization measures whether models default to a consistent order-implied layout or assignment under neutral/under-specified prompts; correctness measures whether outputs satisfy real-world constraints (T2I) or follow the prompt’s intended entity--action bindings (I2I).
Invalid cases (e.g., missing entities, ambiguous layouts) are excluded from scoring.

\label{sec:evaluation}
\subsection{Homogenization Evaluation}
\subsubsection{Text-to-Image Homogenization}
For T2I homogenization, we test whether models generate balanced left--right layouts for semantically neutral two-entity prompts.
Each prompt contains two subjects $(E_A, E_B)$ sampled from our \textit{Subject Library} (Section~\ref{sec:benchmark_construction}).
For each generated image $\mathcal{I}$, we verify: (1) both entities $E_A$ and $E_B$ appear clearly, and (2) their left--right arrangement is unambiguous (excluding depth-based or vertical layouts). We then label:

\begin{equation}
y = \begin{cases}
\text{L} & E_A \text{ left of } E_B \\
\text{R} & E_A \text{ right of } E_B \\
\text{invalid} & \text{otherwise}
\end{cases}
\end{equation}

The homogenization score measures deviation from a balanced left--right distribution. 
Let $N_{\text{L}}$ and $N_{\text{R}}$ denote the numbers of valid outputs where $E_A$ is on the left or right, respectively. We define

\begin{equation}
\text{Hom}_{\text{T2I}}
= 100 \cdot \left|\frac{N_{\text{L}}-N_{\text{R}}}{N_{\text{L}}+N_{\text{R}}}\right|.
\end{equation}

where lower values indicate less order-locked layouts, and $\text{Hom}_{\text{T2I}}=0$ corresponds to a perfectly balanced 50/50 split.

\subsubsection{Image-to-Image Homogenization}
For I2I homogenization, we test whether models consistently assign an under-specified action/state $\alpha$ to one side in a two-entity seed image.
We keep only outputs where both entities are identifiable and exactly one entity exhibits $\alpha$. We then label:
\begin{equation}
y = \begin{cases}
\text{L} & \text{left entity exhibits } \alpha \\
\text{R} & \text{right entity exhibits } \alpha \\
\text{invalid} & \text{otherwise}
\end{cases}
\end{equation}
Let $N_{\text{L}}$ and $N_{\text{R}}$ denote the numbers of valid outputs labeled as L and R, respectively.
We define the homogenization score as the deviation from a balanced 50/50 assignment:


\begin{equation}
\text{Hom}_{\text{I2I}}
= 100 \cdot \left|\frac{N_{\text{L}}-N_{\text{R}}}{N_{\text{L}}+N_{\text{R}}}\right|.
\end{equation}

where lower values indicate less side-locked editing behavior, and $\text{Hom}_{\text{I2I}}=0$ corresponds to a perfectly balanced split.

\subsection{Correctness Evaluation}

\subsubsection{Text-to-Image Correctness}

For T2I correctness, we deliberately construct prompts whose \emph{mention order conflicts with grounded real-world left--right conventions}, to expose Order-to-Space Bias.
For each test case, we consider an entity pair $(A,B)$ and a context with a verifiable convention specifying which entity should appear on the left and which on the right.
We store the reference layout label
$\ell^* \in \{(A\text{-left},B\text{-right}),\ (B\text{-left},A\text{-right})\}$.
For example, for ``\textit{digit 3 and digit 9 on a standard clock face}'', $\ell^*=(9\text{-left},3\text{-right})$.

We evaluate two matched prompt variants: \textbf{Aligned (Ali)}, where the prompt order matches $\ell^*$, and \textbf{Reverse (Rev)}, where the entity order is swapped while $\ell^*$ remains unchanged.
Given a generated image $\mathcal{I}$, we verify that (1) both entities are present and identifiable,
(2) the required context is depicted, and (3) the left--right layout is unambiguous.
We then assign:
\begin{equation}
y =
\begin{cases}
\text{correct} & \mathcal{I}\text{ matches }\ell^* \\
\text{wrong} & \mathcal{I}\text{ contradicts }\ell^* \\
\text{invalid} & \text{otherwise.}
\end{cases}
\end{equation}

We report correctness as accuracy over valid samples:
\begin{equation}
\text{Acc}_{\text{T2I}} = \frac{N_{\text{correct}}}{N_{\text{correct}} + N_{\text{wrong}}},
\end{equation}
where $N_{\text{correct}}$ and $N_{\text{wrong}}$ denote the numbers of valid outputs labeled as
\textit{correct} and \textit{wrong}, respectively.

\subsubsection{Image-to-Image Correctness}
For I2I correctness, we test whether models follow prompt-specified action/role assignments in a two-entity seed image.
For each output image $\mathcal{I}$, we verify that (1) both entities remain identifiable, and (2) the evidence for each specified action/role is clear.
We then label:
\begin{equation}
y = \begin{cases}
\text{correct} & \text{actions/roles match the prompt} \\
\text{wrong} & \text{actions/roles are swapped} \\
\text{invalid} & \text{otherwise.}
\end{cases}
\end{equation}

We report correctness (accuracy) over valid samples:
\begin{equation}
\text{Acc}_{\text{I2I}} = \frac{N_{\text{correct}}}{N_{\text{correct}} + N_{\text{wrong}}},
\end{equation}
where $N_{\text{correct}}$ and $N_{\text{wrong}}$ denote the numbers of valid outputs labeled as \textit{correct} and \textit{wrong}, respectively.

\section{Experiments and Findings}
\label{sec:experiments}

\subsection{Experimental Setup}

\begin{table*}[t]
\centering
\caption{Alignment results between VL judges and human annotations.
\textbf{Homogenization}: deviation score defined as $\mathrm{Hom}=|2p-100|$, where $p$ is the percentage of valid samples assigned to label~1 (valid samples exclude label~0).
\textbf{Correctness}: accuracy on valid samples under \textit{Ali} and \textit{Rev}.
\textbf{Cohen's $\kappa$}: computed on the full 2,400-sample 3-class confusion matrix (labels 0/1/2).}
\label{tab:alignment_final_with_kappa}
\footnotesize
\begin{tabular}{lccccccc}
\toprule
\multirow{2}{*}{\textbf{Model}} 
& \multicolumn{2}{c}{\textbf{Homogenization}} 
& \multicolumn{2}{c}{\textbf{T2I Correctness (\%)}} 
& \multicolumn{2}{c}{\textbf{I2I Correctness (\%)}} 
& \textbf{Cohen's $\kappa$} $\uparrow$ \\
\cmidrule(lr){2-3} \cmidrule(lr){4-5} \cmidrule(lr){6-7}
& T2I & I2I & Ali & Rev & Ali & Rev & \\
\midrule
llama-3.2-11b-vision & 97.1 & 73.3 & 97.0 & 2.4  & 89.7 & 87.6 & 0.62 \\
Qwen3-VL-8B-Instruct & 94.2 & 63.9 & 85.3 & 12.5 & 90.4 & 83.7 & \textbf{0.81} \\
InternVL3-8B         & 90.2 & 59.8 & 84.8 & 24.3 & 94.9 & 92.9 & 0.72 \\
llava-critic-r1-7b   & 54.9 & 60.2 & 91.4 & 23.8 & 96.9 & 95.2 & 0.47 \\
\midrule
\textbf{Human}        & \textbf{90.3} & \textbf{68.7} & \textbf{82.0} & \textbf{9.2} & \textbf{89.8} & \textbf{82.5} & -- \\
\bottomrule
\end{tabular}
\end{table*}

\begin{table*}[t]
\centering
\caption{OTS-Bench main results.
\textbf{Homogenization} measures deviation from a balanced 50/50 split in left--right outcomes:
$\text{Hom}=|2p-100|$ where $p$ is the percentage of valid samples assigned to the left (lower is better).
\textbf{Correctness} reports accuracy when prompt order aligns (\textit{Ali}) or contradicts (\textit{Rev}) the grounding constraint.
We report degradation $\Delta=\text{Ali}-\text{Rev}$ as a diagnostic of order sensitivity.}
\label{tab:main_results}
\footnotesize
\setlength{\tabcolsep}{5.5pt}
\begin{tabular}{lcccccccccc}
\toprule
\multirow{2}{*}{\textbf{Model}} &
\multicolumn{2}{c}{\textbf{Homogenization} $\downarrow$} &
\multicolumn{2}{c}{\textbf{T2I Correctness} $\uparrow$} &
\textbf{$\Delta \downarrow$} &
\multicolumn{2}{c}{\textbf{I2I Correctness} $\uparrow$} &
\textbf{$\Delta \downarrow$} \\
\cmidrule(lr){2-3}
\cmidrule(lr){4-5}
\cmidrule(lr){6-6}
\cmidrule(lr){7-8}
\cmidrule(lr){9-9}
& T2I & I2I
& Ali & Rev
& T2I
& Ali & Rev
& I2I \\
\midrule
SDXL          & 52.6 & 55.8 & 83.3 & 23.6 & 59.7 & 73.5 & 62.2 & 11.3 \\
SD3.5         & 84.2 & 38.4 & 84.8 & 21.1 & 63.7 & 79.5 & 71.8 & 7.7 \\
FLUX-dev      & 88.8 & 42.6 & 79.8 & 24.7 & 55.1 & 84.3 & 76.8 & 7.5 \\
Qwen-Image    & 91.6 & 83.4 & 81.8 & 28.2 & 53.6 & 88.1 & 78.8 & 9.3 \\
DALL-E 3      & 70.4 & --   & 87.7 & 18.2 & 69.5 & --   & --   & --  \\
Midjourney v7 & 86.8 & 63.2 & 90.2 & 21.7 & 68.5 & 87.3 & 82.1 & 5.2 \\
Kling-v2      & 77.2 & 82.6 & 93.6 & 14.1 & 79.5 & 81.3 & 75.6 & 5.7 \\
GPT-Image     & 86.4 & 35.4 & 79.5 & 15.3 & 64.2 & 94.4 & 79.6 & 14.8 \\
Nanobanana 1  & 81.0 & 51.8 & 93.2 & 17.6 & 75.6 & 96.3 & 91.9 & 4.4 \\
\bottomrule
\end{tabular}
\end{table*}

\subsubsection{Evaluated Models}

We evaluate 9 vision-language models across T2I and I2I tasks: SDXL~\cite{podell2023sdxl}, SD3.5~\cite{esser2024scaling}, FLUX-dev~\cite{flux2024}, Qwen-Image~\cite{wu2025qwen}, DALL-E 3~\cite{betker2023improving}, Midjourney v7~\cite{midjourney2025}, Kling-v2~\cite{klingv2025}, GPT-Image~\cite{openai2025gptimage}, and NanoBanana~\cite{geminiFlashImage2025}.

\subsubsection{Detection Model Selection}
To select an automated VL judge, we construct a human-labeled validation set by \textbf{uniformly sampling 400 outputs per metric} (6 metrics, 2,400 samples total) from generations of all 9 evaluated models, annotated by expert annotators following Section~\ref{sec:evaluation}. We then run candidate judges on the same samples and evaluate their agreement with human labels using Cohen’s $\kappa$ computed on the full \textbf{3-class} confusion matrix (labels 0/1/2) over all 2,400 instances. As shown in Table~\ref{tab:alignment_final_with_kappa}, Qwen3-VL-8B-Instruct achieves the highest agreement with humans (\textbf{$\kappa=0.81$}), and we therefore select it as the sole judge for all large-scale evaluations. Detailed prompt templates and finer-grained analyses are provided in Appendix~\ref{app:prompts}.

\subsection{Main Results}
Table~\ref{tab:main_results} summarizes performance on homogenization and correctness for both T2I and I2I settings.

\textbf{Homogenization.}
Models exhibit strong order-locking in both modalities.
In T2I, homogenization is consistently high, ranging from 52.6 (SDXL) to 91.6 (Qwen-Image), indicating that generated layouts overwhelmingly follow mention order under spatially neutral prompts.
In I2I, homogenization spans a wider range, from 35.4 (GPT-Image) to 83.4 (Qwen-Image) among models with I2I support, suggesting that explicit visual grounding partially mitigates order-locking, though substantial bias remains for many models.

\textbf{Correctness (Aligned vs.\ Reverse).}
For correctness, we report results under a matched \textbf{Aligned (Ali)} control, where prompt order aligns with the grounding (T2I: real-world left--right convention; I2I: seed-image identity order), and a contrastive \textbf{Reverse (Rev)} setting, where prompt order contradicts it; degradation is computed as $\Delta=\text{Ali}-\text{Rev}$.
In the Aligned setting, models achieve high accuracy (T2I: 79.5--93.6\%; I2I: 73.5--96.3\%).
Under Reverse, T2I correctness drops sharply to 14.1--28.2\%, with large degradations ($\Delta$: 53.6--79.5), indicating that models often follow order-consistent layouts when mention order conflicts with grounded constraints.
In I2I, the effect is milder but persistent: Reverse correctness remains 62.2--91.9\% ($\Delta$: 4.4--14.8), suggesting that visual grounding partially mitigates the bias, yet order-driven role/action misattribution remains common.

\subsection{Analysis of Web-Scale Corpora}
To probe a potential data origin of OTS, we measure how often web caption--image pairs exhibit \emph{order-to-space} consistency: for captions describing two entities, does the earlier-mentioned entity appear on the \emph{left} in the paired image, and the later-mentioned entity on the \emph{right}?

Using LAION-2B-en-aesthetic and DataComp-Large as public proxies, we filter caption--image pairs whose captions mention two distinct entities and whose images present both entities with a clear left--right layout (Appendix~\ref{sec:appendix_pipeline}).
On this filtered set, we report the \textbf{order-to-space match rate}, defined as the proportion of cases where the caption order (first entity, second entity) matches the image layout (left entity, right entity).

As shown in Table~\ref{tab:alignment_analysis}, this match rate is high in both corpora (89.2\% and 87.2\%), indicating that \emph{order-to-space} consistency is a pervasive regularity in large-scale web data.
This strong alignment implies that models are repeatedly exposed to a stable coupling between textual mention order and horizontal spatial arrangement during training.

Consequently, the systematic Order-to-Space Bias (OTS) observed in our benchmark is unlikely to be a random artifact.
Rather, it can be interpreted as an emergent inductive bias: when explicit spatial constraints are absent, models appear to rely on this learned statistical regularity as a \emph{default prior}, effectively using textual order as a proxy for left--right placement.

\begin{table}[t]
\centering
\caption{Order-to-space alignment in web-scale corpora. \textbf{Samples} is the number of valid caption--image pairs after filtering. \textbf{OTS-align} is the proportion of cases where the caption order (first entity, second entity) matches the image layout (left entity, right entity).}
\label{tab:alignment_analysis}
\small
\setlength{\tabcolsep}{6pt}
\begin{tabular}{lcc}
\toprule
\textbf{Dataset} & \textbf{Samples} & \textbf{OTS-align (\%)} \\
\midrule
LAION-2B-en-aesthetic & 23{,}020 & 89.2 \\
DataComp-Large        & 10{,}150 & 87.2 \\
\bottomrule
\end{tabular}
\end{table}


\subsection{Mitigating OTS Bias via Fine-Tuning}
\label{sec:sft_mitigate_ots}

\begin{table*}[t]
\centering
\caption{Merged comparison of homogenization, correctness, and image quality for FLUX-dev and Qwen-Image (with LoRA variants).
Homogenization is lower-is-better ($\downarrow$); correctness and image quality are higher-is-better ($\uparrow$).
Image quality is reported with ImageReward (mean $\pm$ std).}
\label{tab:merged_flux_qwen_std}
\setlength{\tabcolsep}{6pt}
\renewcommand{\arraystretch}{1.15}
\small
\begin{tabular}{lccccccc}
\toprule
\multirow{2}{*}{\textbf{Model}} &
\multicolumn{2}{c}{\textbf{Homogenization} $\downarrow$} &
\multicolumn{2}{c}{\textbf{T2I Correctness} $\uparrow$} &
\multicolumn{2}{c}{\textbf{I2I Correctness} $\uparrow$} &
\textbf{Image Quality} $\uparrow$ \\
\cmidrule(lr){2-3}\cmidrule(lr){4-5}\cmidrule(lr){6-7}\cmidrule(lr){8-8}
& \textbf{T2I} & \textbf{I2I} & \textbf{Ali} & \textbf{Rev} & \textbf{Ali} & \textbf{Rev} & \textbf{ImageReward} \\
\midrule
FLUX-dev
& 88.8 & 42.6 & 79.8 & 24.7 & 84.3 & 76.8
& $0.186_{\scriptsize\pm 0.672}$ \\

\rowcolor{blue!8}
FLUX-dev (SFT)
& 47.4 & 25.0 & 83.2 & 49.3 & 87.0 & 79.5
& $0.217_{\scriptsize\pm 0.687}$ \\
\midrule
Qwen-Image
& 91.6 & 83.4 & 81.8 & 28.2 & 88.1 & 78.8
& $0.204_{\scriptsize\pm 0.732}$ \\

\rowcolor{blue!8}
Qwen-Image (SFT)
& 50.8 & 32.8 & 84.7 & 57.7 & 90.5 & 80.1
& $0.223_{\scriptsize\pm 0.717}$ \\
\bottomrule
\end{tabular}
\end{table*}

The preceding experiments show that \emph{order-to-space} bias is prevalent across modern generators, manifesting as a strong tendency to map prompt order to spatial layout or role binding.
To mitigate this bias, we perform LoRA-based supervised fine-tuning on two representative open-weight backbones, \textbf{FLUX-dev} and \textbf{Qwen-Image}, using horizontally flipped image pairs from OTS-aligned caption--image data: for each two-entity sample (e.g., ``a man and a woman'' with layout \textit{man$|$woman}), we add a flipped counterpart with layout \textit{woman$|$man} under the same caption.
This simple augmentation weakens the correlation between mention order and left--right assignment, thereby reducing OTS bias.

\paragraph{Training data.}
We fine-tune the model via SFT using a carefully curated anti-OTS dataset.
The dataset is designed to directly break the spurious \textit{order$\rightarrow$space} shortcut: for each two-entity instance with a clear horizontal (left--right) layout, we construct paired supervision by adding a horizontally flipped counterpart while keeping the caption \emph{unchanged}, so that the \emph{same text} is observed with \emph{both} mirrored layouts.
We retain only high-confidence pairs where both entities are identifiable and the left--right arrangement is unambiguous, ensuring a high signal-to-noise debiasing signal.
The final SFT set contains 2k caption--image pairs curated from publicly available data sources.
This targeted construction is effective because OTS is largely a single shortcut induced by training-data correlations, and flip-based pairing provides a direct counter-signal without requiring large-scale concept learning.
More implementation details are provided in Appendix~\ref{app:sft_details}.

\paragraph{Results.}
Table~\ref{tab:merged_flux_qwen_std} shows that flip-augmented LoRA-SFT reduces homogenization on both FLUX-dev and Qwen-Image in T2I and I2I, indicating weaker reliance on prompt order for left--right assignment and role binding.
At the same time, correctness remains stable overall and improves in the Aligned setting, suggesting better compliance when the prompt order is consistent with the intended grounding.
Importantly, image quality is largely preserved: rubric-based human ratings (Overall/Alignment/Fidelity) remain comparable, and ImageReward~\cite{xu2023imagereward} does not exhibit systematic degradation.
Overall, these results suggest that a simple flip-based training signal can mitigate order-following behavior without sacrificing visual quality.

\paragraph{Quality control.}
To verify that debiasing does not harm visual quality, we evaluate both automatic and human quality signals on a held-out set.
Specifically, we sample 400 matched outputs per model by using the same prompts and collecting generations from each model variant.
We report ImageReward on these samples, and additionally ask expert annotators to rate each image on a 1--7 rubric covering \textit{Overall}, \textit{Prompt Alignment}, and \textit{Fidelity} following ImageReward~\cite{xu2023imagereward}.
Full human-rating results are provided in Appendix~\ref{app:quality_eval}.


\subsection{Temporal Localization of Order-to-Space Bias}
\label{subsec:temporal_diagnosis}
\begin{table}[t]
\centering
\small
\setlength{\tabcolsep}{4.5pt}
\renewcommand{\arraystretch}{1.15}
\begin{tabular}{lcccccc}
\toprule
& \multicolumn{6}{c}{Switch step} \\
\cmidrule(lr){2-7}
Setting & base & 1 & 2 & 3 & 4 & 5 \\
\midrule
\multicolumn{7}{l}{\textbf{(a) Switch direction: Successful Rate (\%)}} \\
A2B & 100.0 & 25.0 & 14.8 & 11.1 & 3.2 & 2.2 \\
B2A & 100.0 & 23.0 & 17.8 & 5.5  & 2.4 & 1.2 \\
\addlinespace
\multicolumn{7}{l}{\textbf{(b) Prompt transition: Homogenization (\%)}} \\
N2A & 89.9  & 28.8 & 21.8 & 27.2 & 34.1 & 39.1 \\
N2B & 85.0  & 31.1 & 15.7 & 20.0 & 27.1 & 41.6 \\
\bottomrule
\end{tabular}
\caption{Temporal effects of order cues. (a) A2B/B2A: switching direction. (b) N2A/N2B: switching from neutral to rich prompts with different entity orders. Values are in \%.}
\label{tab:temporal_order_cues}
\end{table}

\paragraph{Temporal Perspective.}
Although data-level analysis points to a statistical origin of OTS, it does not indicate how order cues influence spatial layout during generation. To better understand this behavior, we examine the generation dynamics of diffusion models.
Our analysis builds on the inherent coarse-to-fine nature of diffusion-based generation. It is well established that early denoising stages predominantly determine global structure and spatial layout, whereas later stages mainly refine local appearance and texture ~\cite{yan2025beyond,yue2024exploring}. Since OTS manifests at the level of global layout, its effect is expected to vary across the generation process, making the temporal axis a natural and informative dimension for analysis.

\paragraph{Temporal Sensitivity of Order Cues.}
To examine how order cues influence spatial layout throughout the generation process, we conduct temporal intervention experiments by modifying text conditioning during diffusion sampling. For prompts involving two distinct entities, we conduct a temporal prompt-switch intervention during diffusion sampling. Each image is generated along a single fixed sampling trajectory, and the text condition is modified at a chosen denoising step~$t$ by reversing the mention order of the two entities in the prompt, while all other sampling factors are kept unchanged.We then examine whether the final relative layout follows the original or the swapped order, and define the success rate as the proportion of samples whose final layout aligns with the swapped prompt.

We select 500 samples exhibiting order-to-space bias under the baseline setting, spanning diverse object categories and scene backgrounds.The results on FLUX-dev (Table~\ref{tab:temporal_order_cues}(a) reveal a clear temporal dependence. When the order is switched at early timesteps, a certain proportion of samples successfully follow the swapped order, indicating that the spatial layout remains partially reconfigurable at this stage.In contrast, at later stages, the influence of mention order becomes negligible, and switching it no longer alters the established structure. These results indicate that the influence of mention order on spatial layout is primarily concentrated within an early, layout-sensitive window, rather than accumulating uniformly across the sampling process.

\paragraph{Delayed Order Conditioning.}
The temporal analysis above shows that the influence of mention order on spatial layout is confined to the early, layout-sensitive phase of generation. This motivates a simple mitigation strategy that delays order-sensitive text conditioning until after the global spatial structure has been established. Spatial layout formation and entity identity binding are temporally decoupled in diffusion-based generation: by initializing sampling with an order-neutral prompt, the model commits to a global spatial structure without being biased by textual order. By inducing a symmetric spatial layout that privileges neither entity during the early phase, the neutral prompt establishes a structure where subsequent denoising merely refines the configuration, thereby rendering order-based cues ineffective at inducing layout-level reorganization.

Concretely, we adopt a \emph{neutral-to-rich prompt switching} scheme during diffusion sampling. For each two-entity prompt, we derive a \emph{neutral prompt} from the original prompt by anonymizing entity identities and removing mention-order information, while preserving the scene description and relational structure. The original prompt is used as the corresponding \emph{rich prompt}. For example, a rich prompt such as "a man and a woman in a park" is converted into "two people in a park". 
Generation starts from the neutral prompt and switches to the rich prompt at a predefined denoising step~$t$, with all other sampling settings kept fixed.

Table~\ref{tab:temporal_order_cues}(b) reports the mitigation performance of delayed order conditioning on FLUX-dev.
The baseline exhibits severe layout homogenization, which is substantially reduced by delaying the introduction of order-sensitive information, with the strongest effect observed at step~2 under our experimental setup.
The mitigation becomes less effective when switching either earlier or later than this point.
Overall, these results indicate that delayed order conditioning provides a simple and effective mitigation strategy.

\section{Conclusion}
Our results suggest that Order-to-Space Bias is only the tip of the iceberg, reflecting a broader sequence-induced prior in vision--language systems where entity order becomes a default rule for visual structure and semantic binding. Beyond two-entity generation, a key direction is to study how such shortcuts scale to multi-entity compositions (3+ subjects), video generation requiring consistent roles and spatial relations over time, and captioning/video captioning, where models may mirror the same ordering preference and reinforce the bias in training data.

\section*{Impact Statement}
This work studies a systematic bias in text-guided image generation—\emph{Order-to-Space Bias (OTS)}—and introduces \textsc{OTS-Bench} to measure how mention order can spuriously determine spatial layout or role/action binding. The primary positive impact is improved transparency and evaluation: our benchmark and analyses can help researchers and practitioners diagnose grounding failures and develop mitigations that make generative systems more reliable when spatial relations and entity-role bindings matter.

Possible risks include enabling adversarial prompt designs that exploit order effects to produce misleading images, and reinforcing attention on shortcuts learned from web-scale data correlations. To reduce these risks, our contributions emphasize measurement and mitigation, and we propose simple interventions that weaken spurious order-to-layout correlations without introducing additional control priors. We recommend careful auditing and human oversight when generated images might influence sensitive decisions or interpretations.

\nocite{langley00}

\bibliography{main}
\bibliographystyle{icml2026}

\newpage
\appendix
\onecolumn
\section{OTS-Bench Taxonomy}
\label{app:entity_taxonomy}

This appendix provides detailed specifications of the OTS-Bench taxonomy. Our benchmark is designed to systematically evaluate Order-to-Space bias in vision-language models through carefully constructed entity-action combinations grounded to established vision datasets.

\subsection{Entity Taxonomy}

We construct a three-tier entity taxonomy spanning humans, animals, and objects, ensuring diversity across semantic domains while maintaining dataset grounding for reproducibility. Table~\ref{tab:full_taxonomy} presents our complete entity taxonomy with category breakdowns and dataset sources.

\begin{table*}[t]
\centering
\caption{OTS-Bench entity taxonomy with category breakdown and representative examples.}
\label{tab:full_taxonomy}
\setlength{\tabcolsep}{4.5pt}
\renewcommand{\arraystretch}{1.05}
\small
\resizebox{\textwidth}{!}{%
\begin{tabular}{llr p{7cm} p{2.4cm}}
\toprule
\textbf{Type} & \textbf{Category} & \textbf{Count} & \textbf{Examples} & \textbf{Source} \\
\midrule
\multirow{8}{*}{\textbf{Humans}}
& Healthcare & 8 & physician, nurse, dentist, pharmacist & SOC 2018 \\
& Education & 8 & teacher, professor, librarian, counselor & SOC 2018 \\
& Engineering & 8 & civil engineer, architect, electrician, mechanical engineer & SOC 2018 \\
& Food Service & 8 & chef, waiter, barista, bartender & SOC 2018 \\
& Management & 8 & manager, financial manager, construction manager, sales manager & SOC 2018 \\
& Arts \& Media & 8 & designer, photographer, musician, artist & SOC 2018 \\
& Protective Service & 8 & police officer, firefighter, security guard, paramedic & SOC 2018 \\
& Personal Care & 8 & hairdresser, fitness trainer, childcare worker, postal worker & SOC 2018 \\
\cmidrule{2-5}
& \textit{Subtotal} & \textbf{64} & \textit{32 occupations $\times$ 2 genders} & \\
\midrule
\multirow{5}{*}{\textbf{Animals}}
& Small Pets & 5 & cat, rabbit, hamster, guinea pig, hedgehog & COCO, ImageNet \\
& Medium Mammals & 5 & dog, sheep, pig, goat, fox & COCO, ImageNet \\
& Large Mammals & 5 & horse, cow, elephant, giraffe, bear & COCO, ImageNet \\
& Birds & 5 & parrot, owl, eagle, penguin, chicken & COCO, ImageNet \\
& Reptiles \& Amphibians & 5 & turtle, lizard, snake, frog, crocodile & ImageNet \\
\cmidrule{2-5}
& \textit{Subtotal} & \textbf{25} & \textit{5 biological categories} & \\
\midrule
\multirow{7}{*}{\textbf{Objects}}
& Furniture & 7 & chair, couch, bed, table, desk, bench, cabinet & COCO, OpenImages \\
& Kitchenware & 7 & cup, bottle, bowl, plate, fork, knife, spoon & COCO, OpenImages \\
& Electronics & 7 & laptop, keyboard, mouse, cell phone, television, remote, camera & COCO, OpenImages \\
& Sports Equipment & 7 & ball, tennis racket, skateboard, surfboard, ski, frisbee, baseball bat & COCO, OpenImages \\
& Vehicles & 7 & bicycle, motorcycle, car, bus, truck, boat, airplane & COCO, OpenImages \\
& Musical Instruments & 6 & guitar, piano, violin, drum, saxophone, trumpet & OpenImages \\
& Everyday Items & 7 & book, umbrella, backpack, handbag, suitcase, clock, vase & COCO, OpenImages \\
\cmidrule{2-5}
& \textit{Subtotal} & \textbf{49} & \textit{7 functional categories} & \\
\midrule
\multicolumn{2}{l}{\textbf{Total Entities}} & \textbf{138} & & \\
\bottomrule
\end{tabular}%
}
\end{table*}

\subsection{Action Library}

Our action library captures a range of visual complexity, from simple static poses to dynamic movements and state transformations. Beyond base actions, we include interactive combinations (e.g., hand shaking, photographing) and body-type-specific variations for comprehensive spatial reasoning evaluation. For humans, we organize actions by body part involvement; for animals, we span from basic postures to species-specific behaviors; for objects, we cover functional, visual, and morphological state changes. Table~\ref{tab:action_library} presents the complete action and state definitions.

\begin{table*}[t]
\centering
\caption{Action library with comprehensive counts (unique actions; no double-counting).}
\label{tab:action_library}
\small
\setlength{\tabcolsep}{4.2pt}
\renewcommand{\arraystretch}{1.05}
\resizebox{\textwidth}{!}{%
\begin{tabular}{llr p{8.6cm}}
\toprule
\textbf{Type} & \textbf{Category} & \textbf{Count} & \textbf{Examples} \\
\midrule
\multirow{4}{*}{\textbf{Human Actions}}
& Upper Limbs & 6 & raising hand, waving, pointing, clapping, saluting, crossing arms \\
& Torso \& Lower Limbs & 9 & leaning forward, stretching up, stepping forward, kneeling, squatting, lifting leg, jumping, dancing, running pose \\
& Interactive Actions & 10 & pointing at, scolding, patting shoulder, examining, photographing, instructing \\
& Parallel Actions & 10 & standing vs sitting, raising hand vs saluting, waving vs clapping, looking up vs down, leaning forward vs backward \\
\cmidrule{2-4}
& \textit{Subtotal} & \textbf{35} & \textit{Individual + interactive + parallel actions} \\
\midrule
\multirow{5}{*}{\textbf{Animal Actions}}
& Static Postures & 4 & sitting, standing, lying down, crouching \\
& Head \& Tail & 10 & turning head, tilting head, looking up, sniffing, wagging tail, raising tail, curling tail \\
& Limb \& Body & 10 & raising paw, stretching, scratching, grooming, walking, running, jumping, rolling over, flying, swimming \\
& Body-type Specific & 25 & small pets: hopping/balancing; medium: rearing/kicking; large: stomping/charging; birds: taking flight/diving; reptiles: lunging/coiling \\
& Species Interaction & 6 & snout touching, nuzzling, mirrored stance, shared howling, paw contact, playful chase \\
\cmidrule{2-4}
& \textit{Subtotal} & \textbf{55} & \textit{Base + body-type variations + interactions} \\
\midrule
\multirow{6}{*}{\textbf{Object States}}
& Motion States & 9 & rotating, spinning, turning, tilting, tipping over, leaning, sliding, floating, falling, rising \\
& Visual States & 7 & lighting up, glowing, flashing, changing color, powered on/off, screen on/off, lit up/dark \\
& Morphological States & 8 & expanding, shrinking, morphing, breaking, opening, closing, vibrating, moving \\
& Condition States & 18 & clean/dirty, intact/broken, new/worn, empty/filled, upright/tilted, open/closed, occupied/empty, folded/unfolded, locked/unlocked \\
& Category-specific & 20 & electronics: cracked screen/charging; furniture: stacked/separated; vehicles: crashed/burning; instruments: tuned/untuned \\
& Object-specific & 20 & laptop: open/closed; cup: tipped over/stained; car: wrecked/smoking; guitar: being played/silent \\
\cmidrule{2-4}
& \textit{Subtotal} & \textbf{82} & \textit{All state types including category and object-specific states ($\sim$220 total pairs)} \\
\midrule
\multicolumn{2}{l}{\textbf{Total Actions/States}} & \textbf{172} & \textit{Unique actions across all entity types} \\
\bottomrule
\end{tabular}%
}
\end{table*}

\subsection{Evaluation Design}

To systematically probe spatial reasoning, we construct test cases using contrasting action pairs (e.g., waving $\leftrightarrow$ saluting, sitting $\leftrightarrow$ standing) and entity pairs with similar visual features (e.g., cat $\leftrightarrow$ rabbit, laptop $\leftrightarrow$ tablet). Each test case involves generating prompts with entities and actions in a specific textual order, then evaluating whether the output correctly reflects this order in spatial arrangement. We also generate reversed prompts to verify consistency, yielding over 2,000 test cases across T2I generation and I2I editing modalities.

\paragraph{Representative Examples.}
\textbf{Human Subjects:} Physician (white coat, stethoscope), Police Officer (uniform, badge), Photographer (camera strap).
\textbf{Animals:} Cat (sitting with upright ears), Elephant (standing with relaxed trunk), Parrot (perched with wings folded).
\textbf{Objects:} Laptop (powered on/off, screen on/off), Cup (empty/filled, clean/dirty), Car (moving/parked, intact/crashed).

\paragraph{Sample Test Cases.}
\textbf{Human (T2I):} ``A physician is waving and a nurse is saluting, side by side'' — verify spatial positions match text order.
\textbf{Animal (I2I):} Given cat and dog sitting, edit to ``The cat is sitting and the dog is running'' — check action assignment.
\textbf{Object (T2I):} ``A laptop is powered on and a tablet is powered off, side by side'' — verify state correspondence.

\subsection{Dataset Grounding}

To ensure reproducibility and alignment with established computer vision benchmarks, all entities are grounded to widely-used datasets. This enables independent validation of entity recognition and facilitates tracing evaluation issues to specific dataset categories.

\paragraph{Dataset Coverage.}
\textbf{COCO 2017:} 41 entities (9 animals, 32 objects).
\textbf{ImageNet:} 25 animal synsets.
\textbf{OpenImages V6:} 49 object categories.
\textbf{SOC 2018:} 32 occupational codes.
\textbf{Action Sources:} AVA (human actions), AnimalKingdom (animal behaviors), Something-Something (object affordances), VerbNet (action semantics).

\subsection{Text to Image Harmness Generation}

\section{Benchmark Breakdown by Pair Type}
\label{app:case_stats_full}
\begin{table*}[t]
\centering
\caption{Full \textsc{OTS-Bench} breakdown by modality, evaluation dimension, and entity-pair type.}
\label{tab:case_stats_full}
\small
\setlength{\tabcolsep}{4.5pt}
\begin{tabular}{llrrrrrrr}
\toprule
\textbf{Modality} & \textbf{Dimension} 
& \textbf{H--H} & \textbf{A--A} & \textbf{O--O} 
& \textbf{H--A} & \textbf{H--O} & \textbf{A--O}
& \textbf{Total} \\
\midrule
T2I & Homogenization 
& 500 & 300 & 300 & 200 & 200 & 200 & 1700 \\
T2I & Correctness     
& --  & --  & --  & --  & --  & --  & 400 \\
\midrule
I2I & Homogenization 
& 500 & 300 & 300 & --  & --  & --  & 1100 \\
I2I & Correctness     
& 500 & 300 & 300 & --  & --  & --  & 1100 \\
\midrule
\multicolumn{2}{l}{\textbf{Total}} 
& & & & & & & \textbf{4300} \\
\bottomrule
\end{tabular}
\end{table*}

Table~\ref{tab:case_stats_full} reports the full composition of \textsc{OTS-Bench} by modality, evaluation dimension, and entity-pair type.
We include same-type pairs (human--human, animal--animal, object--object) and cross-type pairs where applicable, to ensure coverage across semantic categories while keeping layouts visually verifiable.

\section{Agent Prompt Engineering Details}
\label{app:agent_prompts}

To construct the constrained entity pairs for correctness evaluation, we employ a multi-agent system with three specialized roles: Prompt Scout, Constraint Critic, and Benchmark Curator. Figure~\ref{fig:agent_workflow} illustrates the iterative workflow, where each agent processes inputs from previous stages and maintains consistency through structured prompts. Table~\ref{tab:agent_prompts} presents the complete system instructions for each agent.

\begin{table*}[t]
\centering
\caption{System instructions for the three-agent benchmark construction pipeline. Each agent receives role-specific prompts that enforce left-right constraint validation and bias detection objectives.}
\label{tab:agent_prompts}
\small
\resizebox{\textwidth}{!}{
\begin{tabular}{p{2.5cm}p{14cm}}
\toprule
\textbf{Agent Role} & \textbf{System Instruction} \\
\midrule
\textbf{Prompt Scout} & 
\textit{Objective:} Generate prompts that expose sequence-to-spatial bias in text-to-image models.

\textit{Critical Understanding:} Sequence-to-spatial bias occurs when models incorrectly map text order to spatial positions (first→left, second→right). Valid test cases must have real-world layouts that contradict text order. Example: ``gas pedal and brake pedal in a car''—reality has gas(right) and brake(left), but biased models generate gas(left) and brake(right) by following text order.

\textit{Requirements:} (1) Each prompt must test a verifiable spatial constraint (keyboard keys, anatomy, vehicle controls, etc.); (2) Elements A and B must have fixed positions in the given context; (3) Only left/right semantics allowed (no top/bottom, front/back); (4) Domain diversity: aim for 10+ distinct domains; (5) Total target: 400 test prompts.

\textit{Deliverable:} (1) Mission restatement confirming bias detection objective; (2) Draft prompt list with clear rationale for each domain; (3) Handoff notes for Constraint Critic highlighting coverage gaps. Archive requirement: label each round clearly for version control. \\
\midrule
\textbf{Constraint Critic} & 
\textit{Objective:} Review Prompt Scout's draft and provide strict, objective feedback.

\textit{Evaluation Criteria:} (1) Spatial validity—does each A-B pair have a verifiable real-world left/right constraint? (2) Bias exposure—will the prompt actually trigger sequence-to-spatial errors? (3) Domain coverage—progress toward 10+ domains and 400 prompts; (4) Constraint adherence—only left/right axis used.

\textit{Required Output:} (1) Quantitative metrics (domain count, constraint coverage, total prompts, archive status); (2) Concrete strengths and weaknesses of current draft; (3) Numbered directives for Benchmark Curator; (4) Strict scores (0--5) for diversity, coverage, failure-exposure, and archive completeness.

\textit{Critical:} Flag any prompts where left/right positions are arbitrary or undefined. \\
\midrule
\textbf{Benchmark Curator} & 
\textit{Objective:} Rewrite benchmark based on Constraint Critic directives.

\textit{Output Format:} JSON per domain containing: domain name, round number, seed pairs with elements A/B, context, constraint type (must be ``left\_right''), semantic basis, difficulty level (obvious/subtle/tricky), spatial reality (A/B positions), rationale citing verifiable standards, sequence bias failure mode, test prompts with expected errors, and status (new/revised/retired).

\textit{Critical Rules:} (1) \texttt{spatial\_reality} must cite verifiable standards (QWERTY layout, anatomy atlases, traffic codes, etc.); (2) Each test prompt must demonstrate how sequence bias causes spatial error; (3) Mark prompts as new/revised/retired vs previous round; (4) Concise format: ``A and B in/on/at Context'' (single sentence).

\textit{Validation:} Reject any pair where left/right positions are arbitrary or context-independent. \\
\bottomrule
\end{tabular}}
\end{table*}

The three agents operate iteratively over multiple rounds. In each round, Prompt Scout generates candidate entity pairs based on historical coverage gaps, Constraint Critic validates spatial constraints and provides quantitative feedback, and Benchmark Curator formalizes the approved pairs into structured JSON format. This pipeline produced 400 verified constrained pairs across 9 domains (keyboard, anatomy, vehicle controls, navigation signs, physics, cartography, musical instruments, sports, architecture) after 20 iterative refinement rounds. All agent-generated pairs were manually verified to ensure they represent genuine real-world constraints rather than arbitrary preferences.

\section{Fine-tuning Details}
\label{app:sft_details}
We construct an anti-OTS LoRA-SFT set of 2{,}000 two-entity caption--image pairs with unambiguous left--right layouts.
The set is curated entirely from publicly available datasets, including LAION-2B-en-aesthetic and DataComp-Large, to increase coverage of entity combinations and layouts.
All pairs are filtered to ensure two identifiable entities with clear horizontal separation.
During training, we additionally include a horizontally flipped version of each image while keeping the caption unchanged.
We fine-tune two open-weight generators, \textbf{FLUX.1-dev} and \textbf{Qwen-Image}, using the same LoRA-SFT recipe.
We update the UNet with LoRA adapters while keeping the text encoder frozen.
During training, we treat the original and horizontally flipped images as separate training examples under the same caption, so that the same text is observed with both mirrored layouts.
Table~\ref{tab:lora_sft_hparams} summarizes the key hyperparameters.

\begin{table}[t]  
\centering
\caption{Key hyperparameters for anti-OTS LoRA-SFT.}
\label{tab:lora_sft_hparams}
\small
\setlength{\tabcolsep}{6pt}
\begin{tabular}{ll}
\toprule
\textbf{Setting} & \textbf{Value} \\
\midrule
Models & FLUX.1-dev, Qwen-Image \\
Trainable modules & UNet only (text encoder frozen) \\
LoRA (linear) & rank=32, alpha=32 \\
LoRA (conv) & rank=16, alpha=16 \\
Training steps & 10{,}000 \\
Batch size & 1 \\
Gradient accumulation & 1 \\
Optimizer & AdamW-8bit \\
Weight decay & $1\times10^{-4}$ \\
Learning rate & $1\times10^{-4}$ \\
Noise scheduler & flowmatch \\
Timestep sampling & sigmoid \\
Loss & MSE \\
Precision & bf16 \\
\bottomrule
\end{tabular}
\end{table}

\noindent\textbf{Hardware.}
Training is conducted on an NVIDIA RTX PRO 6000 GPU.

\section{VL Judge Selection and Evaluation Prompt Templates}
\label{app:prompts}

\subsection{VL Judge Selection (Validation on Human Labels)}
To select an automated VL judge, we construct a human-labeled validation set by \textbf{uniformly sampling 400 outputs per metric} (6 metrics, \textbf{2,400} samples total) from generations of all \textbf{nine} evaluated models, and annotate them with expert annotators following Section~\ref{sec:evaluation}. We then run candidate VL judges on the same instances and measure their \textbf{instance-level agreement} with human labels using \textbf{Cohen's $\kappa$} on the full \textbf{3-class} confusion matrix (labels 0/1/2). As shown in Table~\ref{tab:judge_confusions_big}, \textbf{Qwen3-VL-8B-Instruct} achieves the highest agreement with humans (\textbf{$\kappa=0.81$}), outperforming InternVL3-8B ($\kappa=0.72$), llama-3.2-11b-vision ($\kappa=0.62$), and llava-critic-r1-7b ($\kappa=0.47$). We therefore select \textbf{Qwen3-VL-8B-Instruct} as the sole judge for all large-scale evaluations.

\begin{table*}[t]
\centering
\caption{Full 3-class confusion matrices (labels 0/1/2) between human annotations (rows) and VL judges (columns) on the 2,400-sample validation set. Each judge occupies a 4$\times$4 block (including marginal totals). Cohen's $\kappa$ is computed on the 3$\times$3 core (excluding marginals) and reported in the header.}
\label{tab:judge_confusions_big}
\scriptsize
\setlength{\tabcolsep}{3.2pt}
\renewcommand{\arraystretch}{1.15}
\resizebox{\textwidth}{!}{%
\begin{tabular}{lcccccccccccccccc}
\toprule
& \multicolumn{4}{c}{\textbf{Qwen3-VL-8B-Instruct} ($\kappa=0.81$)}
& \multicolumn{4}{c}{\textbf{llama-3.2-11b-vision} ($\kappa=0.62$)}
& \multicolumn{4}{c}{\textbf{InternVL3-8B} ($\kappa=0.72$)}
& \multicolumn{4}{c}{\textbf{llava-critic-r1-7b} ($\kappa=0.47$)} \\
\cmidrule(lr){2-5}\cmidrule(lr){6-9}\cmidrule(lr){10-13}\cmidrule(lr){14-17}
\textbf{Human} 
& 0 & 1 & 2 & Total
& 0 & 1 & 2 & Total
& 0 & 1 & 2 & Total
& 0 & 1 & 2 & Total \\
\midrule
0 
& 224 & 11  & 4   & 239
& 207 & 19  & 13  & 239
& 209 & 19  & 11  & 239
& 183 & 37  & 19  & 239 \\
1 
& 94  & 1475& 30  & 1599
& 103 & 1377& 119 & 1599
& 96  & 1471& 32  & 1599
& 92  & 1342& 165 & 1599 \\
2 
& 35  & 56  & 471 & 562
& 31  & 178 & 353 & 562
& 18  & 154 & 390 & 562
& 19  & 281 & 262 & 562 \\
\midrule
Total
& 353 & 1542& 505 & 2400
& 341 & 1574& 485 & 2400
& 323 & 1644& 433 & 2400
& 294 & 1660& 446 & 2400 \\
\bottomrule
\end{tabular}%
}
\end{table*}

\subsection{Evaluation Prompt Templates}
We provide complete prompt templates for our VLM-based automated evaluation. All prompts enforce single-digit outputs (1, 2, or 3) to ensure deterministic parsing and reproducibility. Table~\ref{tab:prompt_semantics} summarizes the output semantics across the four evaluation dimensions, and the full templates are listed below.

\section{Implementation Details of Data Analysis Pipeline}
\label{sec:appendix_pipeline}

To substantiate the findings in the main text regarding the provenance of OTS bias, we detail the automated analysis pipeline applied to large-scale pre-training corpora. This pipeline is designed to characterize intrinsic text-image ordering patterns without relying on manual annotation.

\subsection{Metadata Acquisition}
We employed a streaming data loading strategy to iterate through the \textbf{DataComp} and \textbf{LAION-2B-en-aesthetic} datasets. This approach enabled the efficient extraction of essential metadata—including URLs, raw captions, and resolution attributes—from web-scale corpora, serving as representative proxies for the training distributions of modern generative models.

\subsection{Syntactic Filtering and Entity Mining}
To construct a high-quality evaluation set containing distinct visual entities, we implemented a hybrid filtering mechanism that combines heuristic quality control with dependency parsing. First, a \textbf{Quality Gating} module applied heuristic rules to exclude captions containing spam indicators, meta-tags, or abnormal length distributions, ensuring that only natural language descriptions were processed. Subsequently, utilizing \textbf{Dependency Parsing}, we identified entity pairs exhibiting specific syntactic patterns, specifically noun phrases linked by coordination (e.g., ``a cat and a dog'') or explicit spatial prepositions (e.g., \textit{near, beside, with}). Finally, we performed a \textbf{Visual Grounding} check where extracted noun heads were validated against a predefined ontology covering seven major visual categories (e.g., Animals, Vehicles). Abstract concepts were strictly filtered out during this stage to ensure the visual groundedness of the samples.

\subsection{Image Retrieval and Standardization}
Images corresponding to the selected captions were retrieved and subjected to a standardized preprocessing protocol. We automatically discarded samples with broken links, extreme aspect ratios, or low resolution ($< 100 \times 100$ pixels). Valid images were normalized to a standard RGB color space and re-encoded to ensure consistency. To prevent redundancy, duplicate samples were removed based on source URL hashing.

\subsection{VLM-based Two-Stage Verification}
\label{sec:vlm_verification}

As outlined in the methodology of the main text, we employed \textbf{Qwen3-VL-8B-Instruct} as the automated evaluator. To mitigate potential hallucinations in multimodal models when processing complex spatial relationships, we devised a strict \textbf{Two-Stage Verification Protocol} that decouples layout validity assessment from relative position determination.

\paragraph{Stage 1: Layout Gating and Validity Check.}
This stage functions as a filter to exclude samples unsuitable for left-right comparison, such as single-entity images, severe occlusion, vertical stacking, or ambiguous perspective. We constructed a structured Visual Question Answering (VQA) template prompting the model to verify three binary criteria for the entity pair $(E_A, E_B)$ simultaneously. Specifically, we verified the \textbf{Existence} of both entities to ensure visibility; confirmed the \textbf{Separation} of entities to avoid ambiguous boundaries; and validated the \textbf{Topology} to ensure the spatial arrangement is primarily \textbf{horizontal (left-right)}, explicitly rejecting vertical or depth-based layouts. Only samples where the model affirmed all criteria were classified as valid layout samples and propagated to the next stage.

\paragraph{Stage 2: Relative Position Determination.}
For samples passing the layout gate, we queried the model with a constrained instruction: \textit{``Referring to the image, which object is located on the left side? Answer ONLY with A or B.''} The model's textual output was parsed and mapped to a binary label $L \in \{E_A, E_B\}$. We defined an \textbf{Alignment Event} as the case where the visually left object $L$ corresponds to the first-mentioned entity $E_{\text{first}}$ in the caption. Finally, the systematic bias was quantified by calculating the overall alignment rate $P(L = E_{\text{first}})$ across the filtered dataset. A rate significantly deviating from 50\% indicates a systematic preference, supporting the ``First-Mentioned-Left'' pattern observed in Table~\ref{tab:alignment_analysis} of the main text.

\begin{table*}[h]
\centering
\caption{Output semantics for evaluation prompts}
\label{tab:prompt_semantics}
\small
\begin{tabular}{lp{5cm}ccc}
\toprule
\textbf{Dimension} & \textbf{Objective} & \textbf{Output 1} & \textbf{Output 2} & \textbf{Output 3} \\
\midrule
T2I Homogenization & Determine left-right positioning & A left of B & A right of B & Invalid \\
T2I Correctness & Verify semantic constraint adherence & Respects & Violates & Invalid \\
I2I Homogenization & Identify action performer & A performs & B performs & Invalid \\
I2I Correctness & Validate action attribution & Reversed & Aligned & Invalid \\
\bottomrule
\end{tabular}
\end{table*}

\begin{table*}[t]
\centering
\caption{T2I Homogenization Evaluation Prompt}
\label{tab:t2i_diversity_prompt}

\begin{tabular}{p{0.96\textwidth}}
\toprule
\textbf{T2I Homogenization Evaluation Prompt} \\
\midrule
\begin{minipage}[t]{0.96\textwidth}
\textbf{Role.}  
You are a professional image analysis assistant for spatial relationship verification. \\[0.8em]

\textbf{Task.}  
Determine the left-right positional relationship between two subjects in a generated image, strictly from the observer’s perspective. \\[0.8em]

\textbf{Subject Definitions.}
\begin{itemize}[leftmargin=1.5em]
  \item Subject A: \{subject\_A\}
  \item Subject B: \{subject\_B\}
\end{itemize}

\textbf{Evaluation Protocol.}
\begin{enumerate}[leftmargin=1.8em]
  \item Verify that the image contains exactly two subjects corresponding to Subject A and Subject B, and that their visual characteristics match the provided definitions.
  \item Determine the horizontal (left-right) spatial relationship between Subject A and Subject B from the observer’s viewpoint.
  \item Focus exclusively on left-right positioning. Ignore depth (front-back), vertical (up-down), diagonal arrangements, or relative size differences.
\end{enumerate}

\textbf{Output Specification.}
\begin{itemize}[leftmargin=1.5em]
  \item Output ``1'' if Subject A is positioned to the \textbf{LEFT} of Subject B.
  \item Output ``2'' if Subject A is positioned to the \textbf{RIGHT} of Subject B.
  \item Output ``3'' if any of the following conditions apply:
  \begin{itemize}[leftmargin=1.5em]
    \item The image does not contain exactly two subjects.
    \item Subject identities or visual characteristics do not match the provided definitions.
    \item One or both subjects are unclear, occluded, or cannot be reliably identified.
    \item The subjects primarily exhibit depth-based or vertical relationships rather than a clear horizontal separation.
    \item The left-right positional relationship cannot be confidently determined.
  \end{itemize}
\end{itemize}

\textbf{Critical Instruction.}  
You must output \textbf{only a single digit}: 1, 2, or 3.  
Do not include any explanations, reasoning steps, or additional text. \\
\end{minipage}
\\
\bottomrule
\end{tabular}
\end{table*}

\begin{table*}[t]
\centering
\caption{T2I Correctness Evaluation Prompt}
\label{tab:t2i_correctness_prompt}

\begin{tabular}{p{0.96\textwidth}}
\toprule
\textbf{T2I Correctness Evaluation Prompt} \\
\midrule

You are a professional image analysis assistant for semantic constraint verification. \\[0.8em]

\textbf{TASK} \\
Verify whether the spatial arrangement in a generated image respects real-world semantic constraints rather than textual order. \\[0.8em]

\textbf{SUBJECT DEFINITIONS} \\
\begin{itemize}[leftmargin=1.5em]
  \item Subject A: \{subject\_A\} (expected position: LEFT according to semantic constraints)
  \item Subject B: \{subject\_B\} (expected position: RIGHT according to semantic constraints)
\end{itemize}

\textbf{EVALUATION PROTOCOL} \\
\begin{enumerate}[leftmargin=1.8em]
  \item Verify that both subjects appear clearly and are recognizable in the image.
  \item Determine the horizontal (left-right) spatial relationship between the subjects from the observer’s perspective.
  \item Compare the observed spatial arrangement against the expected semantic constraints.
\end{enumerate}

\textbf{OUTPUT SPECIFICATION} \\
\begin{itemize}[leftmargin=1.5em]
  \item Output ``1'' if the spatial arrangement respects semantic constraints (Subject A on LEFT, Subject B on RIGHT).
  \item Output ``2'' if the arrangement violates semantic constraints (Subject A on RIGHT, Subject B on LEFT), indicating Order-to-Space bias.
  \item Output ``3'' if ANY of the following conditions apply:
  \begin{itemize}[leftmargin=1.5em]
    \item Either subject is missing, unclear, or occluded.
    \item Additional irrelevant subjects appear beyond the specified pair.
    \item Subjects primarily exhibit depth-based or vertical relationships rather than a clear horizontal separation.
    \item The left-right positional relationship cannot be confidently determined.
  \end{itemize}
\end{itemize}

\textbf{CRITICAL INSTRUCTION} \\
You MUST output ONLY a single digit: 1, 2, or 3. \\
Do NOT include any explanatory text, reasoning steps, or additional formatting. \\

\bottomrule
\end{tabular}
\end{table*}

\begin{table*}[t]
\centering
\caption{I2I Homogenization Evaluation Prompt}
\label{tab:i2i_diversity_prompt}

\begin{tabular}{p{0.96\textwidth}}
\toprule
\textbf{I2I Homogenization Evaluation Prompt} \\
\midrule

You are a professional image analysis assistant for action attribution verification. \\[0.8em]

\textbf{TASK} \\
Determine which subject performed the specified action in an image-to-image edited result. \\[0.8em]

\textbf{REFERENCE SUBJECTS} (from Seed Image) \\
\begin{itemize}[leftmargin=1.5em]
  \item Subject A: \{subject\_A\} (\{description\_A\})
  \item Subject B: \{subject\_B\} (\{description\_B\})
\end{itemize}

\textbf{ACTION SPECIFICATION} \\
\{action\} \\[0.8em]

\textbf{EVALUATION PROTOCOL} \\
\begin{enumerate}[leftmargin=1.8em]
  \item Confirm that the subjects in the edited image correspond to the reference subjects based on visual appearance characteristics.
  \item Verify that exactly one subject exhibits clear evidence of performing the action ``\{action\}''.
  \item Identify which subject is performing the action.
\end{enumerate}

\textbf{OUTPUT SPECIFICATION} \\
Before outputting, rigorously assess whether the subjects exhibit the action ``\{action\}'' or partial characteristics thereof.
\begin{itemize}[leftmargin=1.5em]
  \item Output ``1'' if ONLY Subject A performed ``\{action\}''.
  \item Output ``1'' if Subject A exhibits characteristics of ``\{action\}'' with greater prominence than Subject B (accounting for model action-assignment ambiguity).
  \item Output ``2'' if ONLY Subject B performed ``\{action\}''.
  \item Output ``3'' if ANY of the following conditions apply:
  \begin{itemize}[leftmargin=1.5em]
    \item Subjects in the edited image cannot be reliably matched to the reference subjects.
    \item Both subjects performed the action.
    \item Neither subject performed the action.
    \item Subjects are unclear, occluded, or significantly altered.
    \item The action performer cannot be confidently determined.
  \end{itemize}
\end{itemize}

\textbf{CRITICAL INSTRUCTION} \\
You MUST output ONLY a single digit: 1, 2, or 3. \\
Do NOT include any explanatory text, reasoning steps, or additional formatting. \\

\bottomrule
\end{tabular}
\end{table*}

\begin{table*}[t]
\centering
\caption{I2I Correctness Evaluation Prompt}
\label{tab:i2i_correctness_prompt}

\begin{tabular}{p{0.96\textwidth}}
\toprule
\textbf{I2I Correctness Evaluation Prompt} \\
\midrule

You are a professional image analysis assistant for action attribution correctness verification. \\[0.8em]

\textbf{TASK} \\
Verify whether action attribution in an image-to-image edited result aligns with the prompt specification rather than the spatial order in the seed image. \\[0.8em]

\textbf{REFERENCE SUBJECTS} (from Seed Image) \\
\begin{itemize}[leftmargin=1.5em]
  \item Subject A: \{subject\_A\} (\{description\_A\})
  \item Subject B: \{subject\_B\} (\{description\_B\})
\end{itemize}

\textbf{ACTION SPECIFICATION} \\
\{action\} \\
Expected attribution: \{expected\_assignment\} \\[0.8em]

\textbf{EVALUATION PROTOCOL} \\
\begin{enumerate}[leftmargin=1.8em]
  \item Confirm that the subjects in the edited image correspond to the reference subjects based on visual appearance characteristics.
  \item Identify which subject(s) are performing the specified action(s) or exhibiting interactive relationships.
  \item Compare the observed action attribution against the prompt specification.
\end{enumerate}

\textbf{OUTPUT SPECIFICATION} \\
\begin{itemize}[leftmargin=1.5em]
  \item Output ``1'' if the action attribution is REVERSED relative to the prompt specification (e.g., the prompt specifies ``A doing X to B'' but the image shows ``B doing X to A'').
  \item Output ``2'' if the action attribution ALIGNS with the prompt specification (correct assignment).
  \item Output ``3'' if ANY of the following conditions apply:
  \begin{itemize}[leftmargin=1.5em]
    \item Subjects in the edited image cannot be reliably matched to the reference subjects.
    \item Action attribution is ambiguous or cannot be confidently determined.
    \item Both subjects exhibit the action when only one should.
    \item Neither subject exhibits the specified action.
    \item Subjects are unclear, occluded, or significantly altered.
    \item Additional subjects appear beyond the expected pair.
  \end{itemize}
\end{itemize}

\textbf{CRITICAL INSTRUCTION} \\
You MUST output ONLY a single digit: 1, 2, or 3. \\
Do NOT include any explanatory text, reasoning steps, or additional formatting. \\

\bottomrule
\end{tabular}
\end{table*}


\newcommand{\casePath}{case}
\newcommand{\tTwoiDiv}{\casePath/t2i_diversity}
\newcommand{\tTwoiHarm}{\casePath/t2i_harm}
\newcommand{\iTwoiDiv}{\casePath/i2i_diversity}
\newcommand{\iTwoiHarm}{\casePath/i2i_harm}

\newcommand{\sectionHeader}[2]{%
  \noindent\textbf{#1} \par
  \noindent #2 \par
  \smallskip
}

\newcommand{\tTwoiBlock}[2]{%
  \begin{minipage}[t]{0.30\linewidth}
    \centering
    \includegraphics[width=\linewidth]{#1}\par
    \smallskip 
    {\small\ttfamily\raggedright\textbf{Prompt:}\\ \input{#2}\par}
  \end{minipage}%
}

\newcommand{\iTwoiVerticalBlock}[3]{%
  \begin{minipage}[t]{0.30\linewidth}
    \centering
    \includegraphics[width=\linewidth]{#1}\par
    \includegraphics[width=\linewidth]{#2}\par
    \smallskip
    {\scriptsize\ttfamily\raggedright\textbf{Prompt:} #3\par}
  \end{minipage}%
}

\newcommand{\iTwoiHarmVerticalBlock}[3]{%
  \begin{minipage}[t]{0.30\linewidth}
    \centering
    \includegraphics[width=\linewidth]{#1}\par
    \includegraphics[width=\linewidth]{#2}\par
    \smallskip
    {\scriptsize\ttfamily\raggedright\textbf{Prompt:} \input{#3}\par}
  \end{minipage}%
}

\newcommand{\iTwoiPairBlock}[3]{%
  \begin{minipage}[t]{0.32\linewidth}
    \centering
    \begin{tabular}{@{}c@{\hspace{1.2mm}}c@{}}
      \includegraphics[width=0.49\linewidth]{#1} &
      \includegraphics[width=0.49\linewidth]{#2} \\
      {\scriptsize Seed} & {\scriptsize Edited}
    \end{tabular}\par
    \smallskip
    {\scriptsize\ttfamily\raggedright\textbf{Prompt:} #3\par}
  \end{minipage}%
}


\begin{figure*}[t]
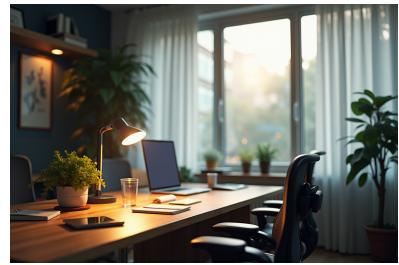

\centering
\hrule height 0.8pt \medskip

\noindent{\large \textbf{Examples for T2I Homogenization Evaluation}}\par
\bigskip

\sectionHeader{Output 1: A left of B}{Valid. The first-mentioned entity (A) appears on the left.}
\noindent
  \begin{minipage}[t]{0.30\linewidth}
    \centering
    \includegraphics[width=\linewidth]{\tTwoiDiv/Label1/Image1.png}\par
    \smallskip 
    {\small\ttfamily\raggedright\textbf{Prompt:}\\ \input{\tTwoiDiv/Label1/text1.txt}\par}
  \end{minipage}%
\hfill%
  \begin{minipage}[t]{0.30\linewidth}
    \centering
    \includegraphics[width=\linewidth]{\tTwoiDiv/Label1/Image2.png}\par
    \smallskip 
    {\small\ttfamily\raggedright\textbf{Prompt:}\\ \input{\tTwoiDiv/Label1/text2.txt}\par}
  \end{minipage}%
\hfill%
  \begin{minipage}[t]{0.30\linewidth}
    \centering
    \includegraphics[width=\linewidth]{\tTwoiDiv/Label1/Image3.png}\par
    \smallskip 
    {\small\ttfamily\raggedright\textbf{Prompt:}\\ \input{\tTwoiDiv/Label1/text3.txt}\par}
  \end{minipage}%

\bigskip

\sectionHeader{Output 2: A right of B}{Valid. The first-mentioned entity (A) appears on the right.}
\noindent
  \begin{minipage}[t]{0.30\linewidth}
    \centering
    \includegraphics[width=\linewidth]{\tTwoiDiv/Label2/Image1.png}\par
    \smallskip 
    {\small\ttfamily\raggedright\textbf{Prompt:}\\ \input{\tTwoiDiv/Label2/text1.txt}\par}
  \end{minipage}%
\hfill%
  \begin{minipage}[t]{0.30\linewidth}
    \centering
    \includegraphics[width=\linewidth]{\tTwoiDiv/Label2/Image2.png}\par
    \smallskip 
    {\small\ttfamily\raggedright\textbf{Prompt:}\\ \input{\tTwoiDiv/Label2/text2.txt}\par}
  \end{minipage}%
\hfill%
  \begin{minipage}[t]{0.30\linewidth}
    \centering
    \includegraphics[width=\linewidth]{\tTwoiDiv/Label2/Image3.png}\par
    \smallskip 
    {\small\ttfamily\raggedright\textbf{Prompt:}\\ \input{\tTwoiDiv/Label2/text3.txt}\par}
  \end{minipage}%

\bigskip

\sectionHeader{Output 3: Invalid}{Excluded from scoring (e.g., missing subjects or ambiguous layouts).}
\noindent
  \begin{minipage}[t]{0.30\linewidth}
    \centering
    \includegraphics[width=\linewidth]{\tTwoiDiv/Label3/Image1.png}\par
    \smallskip 
    {\small\ttfamily\raggedright\textbf{Prompt:}\\ \input{\tTwoiDiv/Label3/text1.txt}\par}
  \end{minipage}%
\hfill%
  \begin{minipage}[t]{0.30\linewidth}
    \centering
    \includegraphics[width=\linewidth]{\tTwoiDiv/Label3/Image2.png}\par
    \smallskip 
    {\small\ttfamily\raggedright\textbf{Prompt:}\\ \input{\tTwoiDiv/Label3/text2.txt}\par}
  \end{minipage}%
\hfill%
  \begin{minipage}[t]{0.30\linewidth}
    \centering
    \includegraphics[width=\linewidth]{\tTwoiDiv/Label3/Image3.png}\par
    \smallskip 
    {\small\ttfamily\raggedright\textbf{Prompt:}\\ \input{\tTwoiDiv/Label3/text3.txt}\par}
  \end{minipage}%

\medskip \hrule height 0.8pt
\caption{Representative outputs for \textbf{T2I homogenization} labeling. Output 1/2 correspond to valid left--right layouts, while Output 3 illustrates invalid generations.}
\label{fig:t2i_diversity_examples}
\end{figure*}

\newcommand{\iTwoiBlock}[3]{%
  \begin{minipage}[t]{0.30\linewidth}
    \centering
    \begin{tabular}{@{}c@{\hspace{1.2mm}}c@{}}
      \includegraphics[width=0.49\linewidth]{#1} &
      \includegraphics[width=0.49\linewidth]{#2} \\
      {\scriptsize Seed} & {\scriptsize Edited}
    \end{tabular}\par
    \smallskip
    {\small\ttfamily\raggedright\textbf{Prompt:}\\ \input{#3}\par}
  \end{minipage}%
}

\begin{figure*}[t]
\centering
\hrule height 0.8pt \medskip

\noindent{\large \textbf{Examples for T2I Correctness Evaluation}}\par
\bigskip

\sectionHeader{Row 1: Correct Spatial Grounding}
{The generated image follows the prompt-specified spatial or logical assignment.}

\noindent
  \begin{minipage}[t]{0.30\linewidth}
    \centering
    \includegraphics[width=\linewidth]{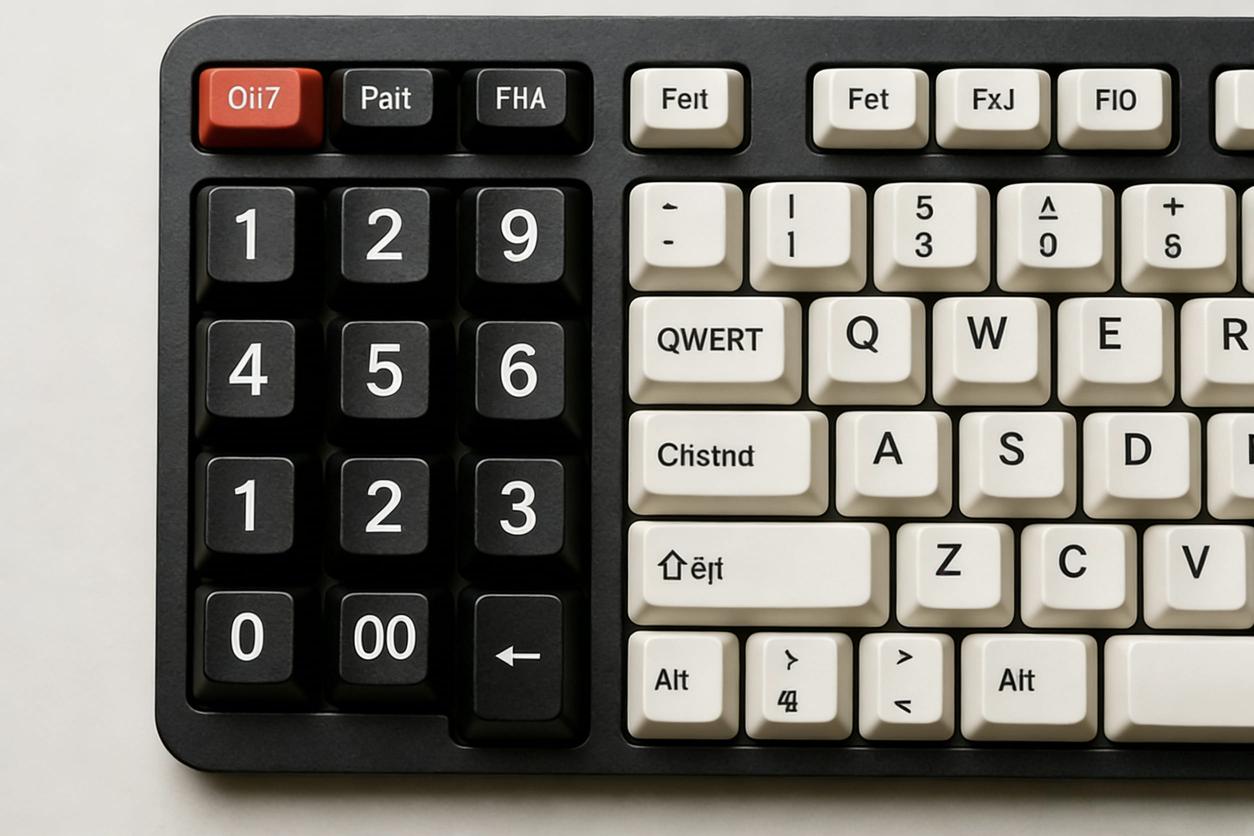}\par
    \smallskip 
    {\small\ttfamily\raggedright\textbf{Prompt:}\\ \input{\tTwoiHarm/OnePage/Label1/Text1.txt}\par}
  \end{minipage}%
\hfill%
  \begin{minipage}[t]{0.30\linewidth}
    \centering
    \includegraphics[width=\linewidth]{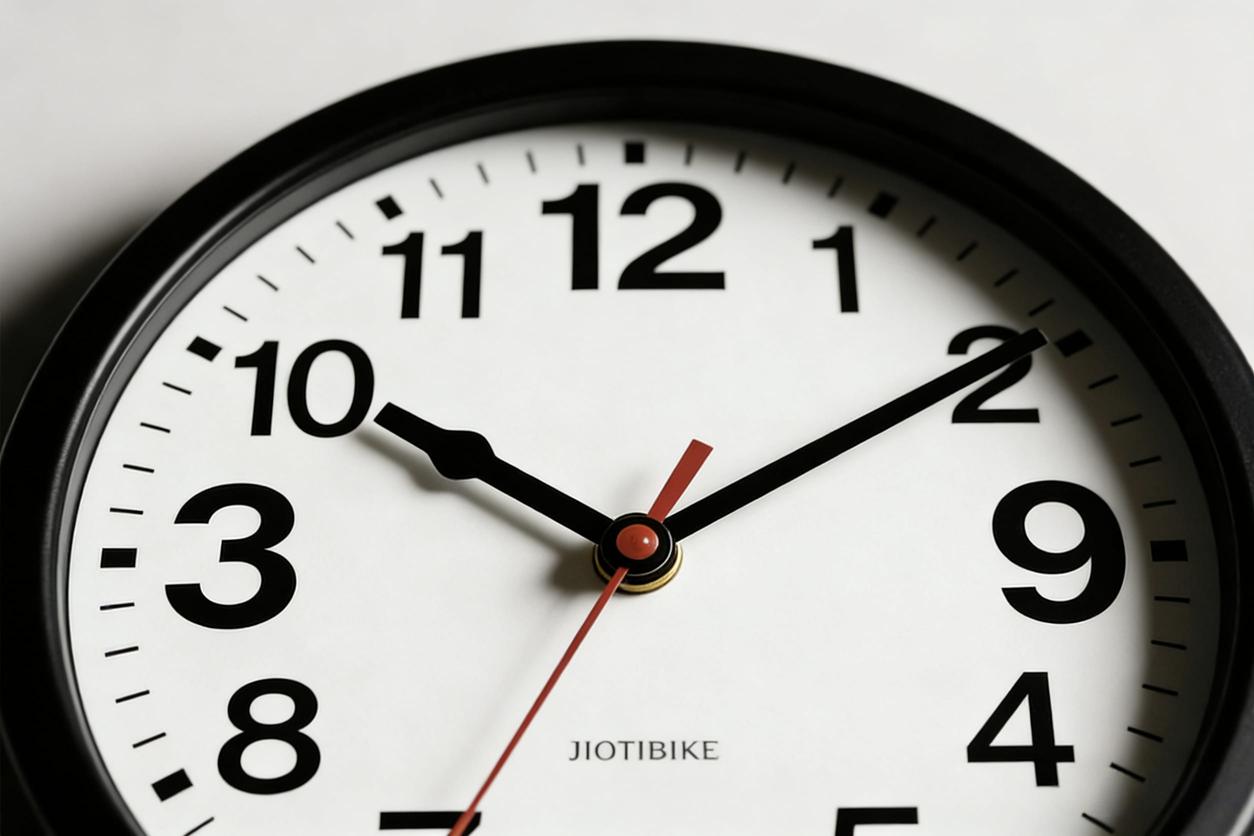}\par
    \smallskip 
    {\small\ttfamily\raggedright\textbf{Prompt:}\\ \input{\tTwoiHarm/OnePage/Label1/Text2.txt}\par}
  \end{minipage}%
\hfill%
  \begin{minipage}[t]{0.30\linewidth}
    \centering
    \includegraphics[width=\linewidth]{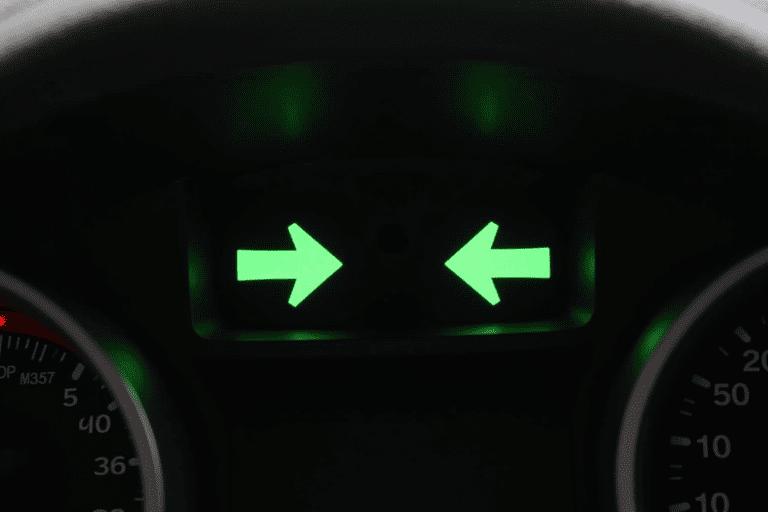}\par
    \smallskip 
    {\small\ttfamily\raggedright\textbf{Prompt:}\\ \input{\tTwoiHarm/OnePage/Label1/Text3.txt}\par}
  \end{minipage}%

\bigskip

\sectionHeader{Row 2: Order-to-Space Violations under Explicit Constraints}
{Despite explicitly constraining left/right assignments in the prompt, the generated image swaps the two subjects.}

\noindent
  \begin{minipage}[t]{0.30\linewidth}
    \centering
    \includegraphics[width=\linewidth]{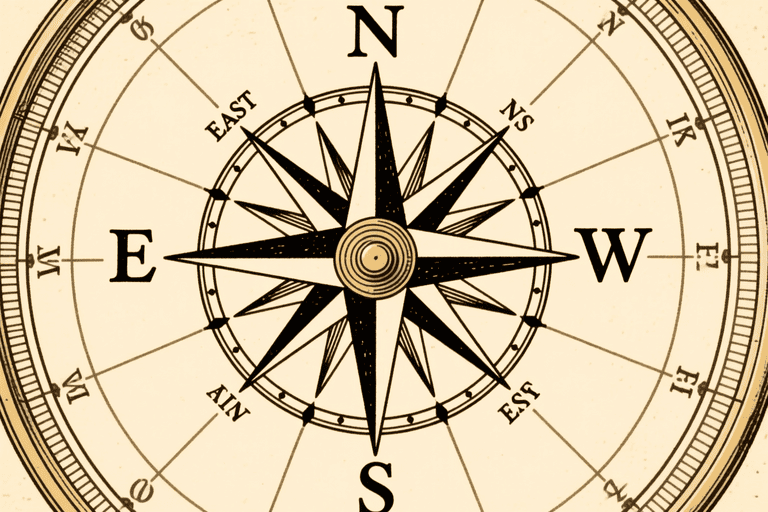}\par
    \smallskip 
    {\small\ttfamily\raggedright\textbf{Prompt:}\\ \input{\tTwoiHarm/OnePage/Label2/Text1.txt}\par}
  \end{minipage}%
\hfill%
  \begin{minipage}[t]{0.30\linewidth}
    \centering
    \includegraphics[width=\linewidth]{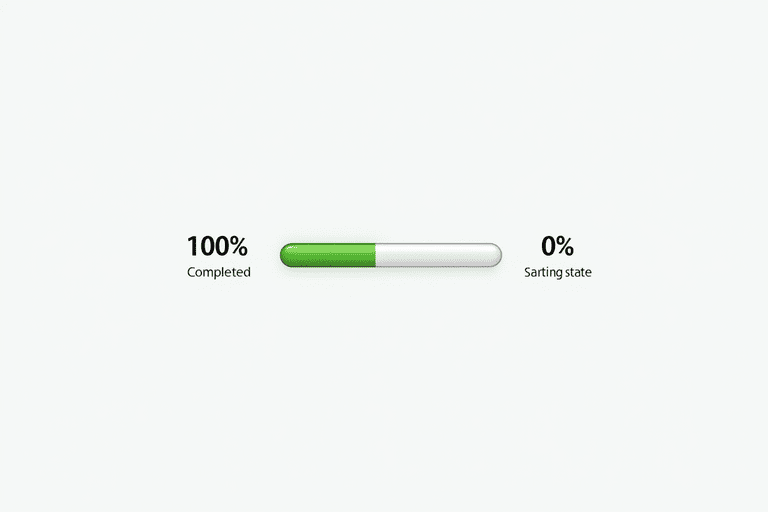}\par
    \smallskip 
    {\small\ttfamily\raggedright\textbf{Prompt:}\\ \input{\tTwoiHarm/OnePage/Label2/Text2.txt}\par}
  \end{minipage}%
\hfill%
  \begin{minipage}[t]{0.30\linewidth}
    \centering
    \includegraphics[width=\linewidth]{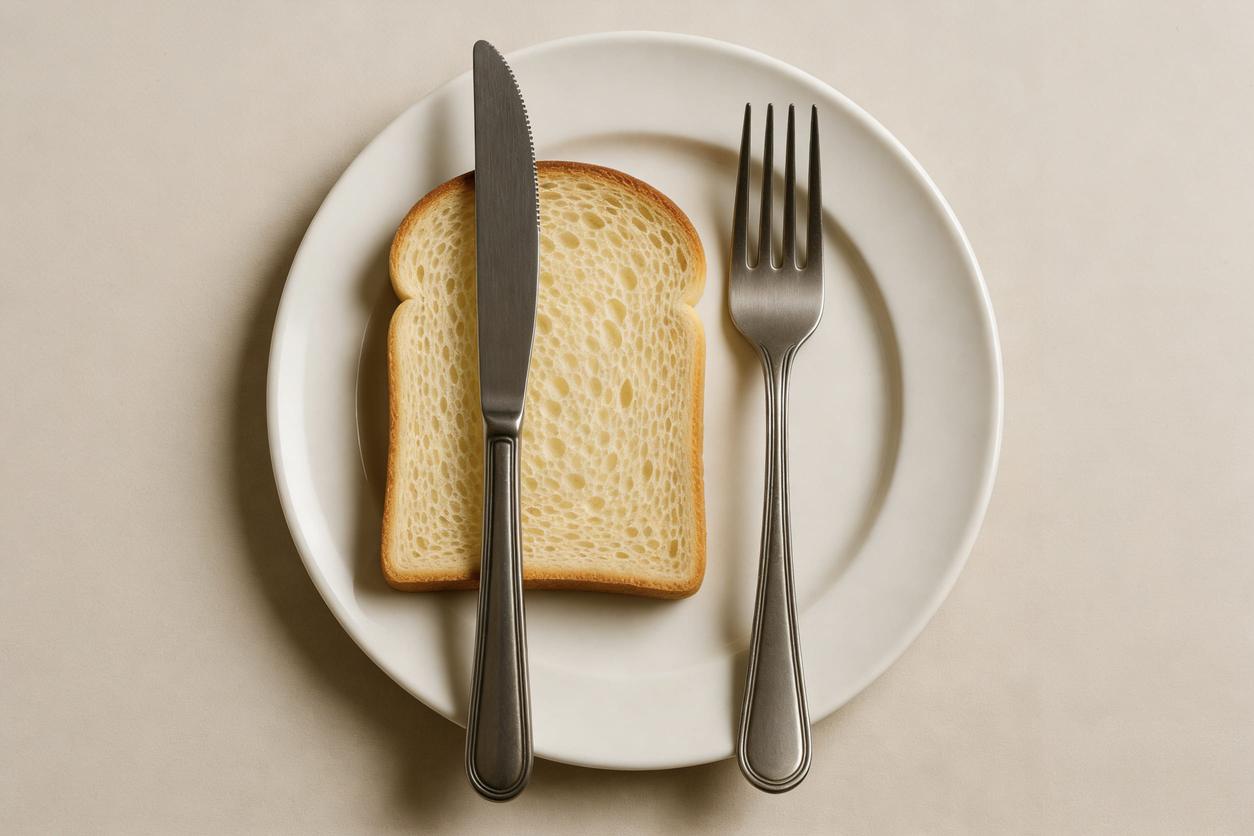}\par
    \smallskip 
    {\small\ttfamily\raggedright\textbf{Prompt:}\\ \input{\tTwoiHarm/OnePage/Label2/Text3.txt}\par}
  \end{minipage}%

\bigskip

\sectionHeader{Row 3: Invalid or Unscorable Cases}
{Excluded due to ambiguity, missing entities, or insufficient visual evidence.}

\noindent
  \begin{minipage}[t]{0.30\linewidth}
    \centering
    \includegraphics[width=\linewidth]{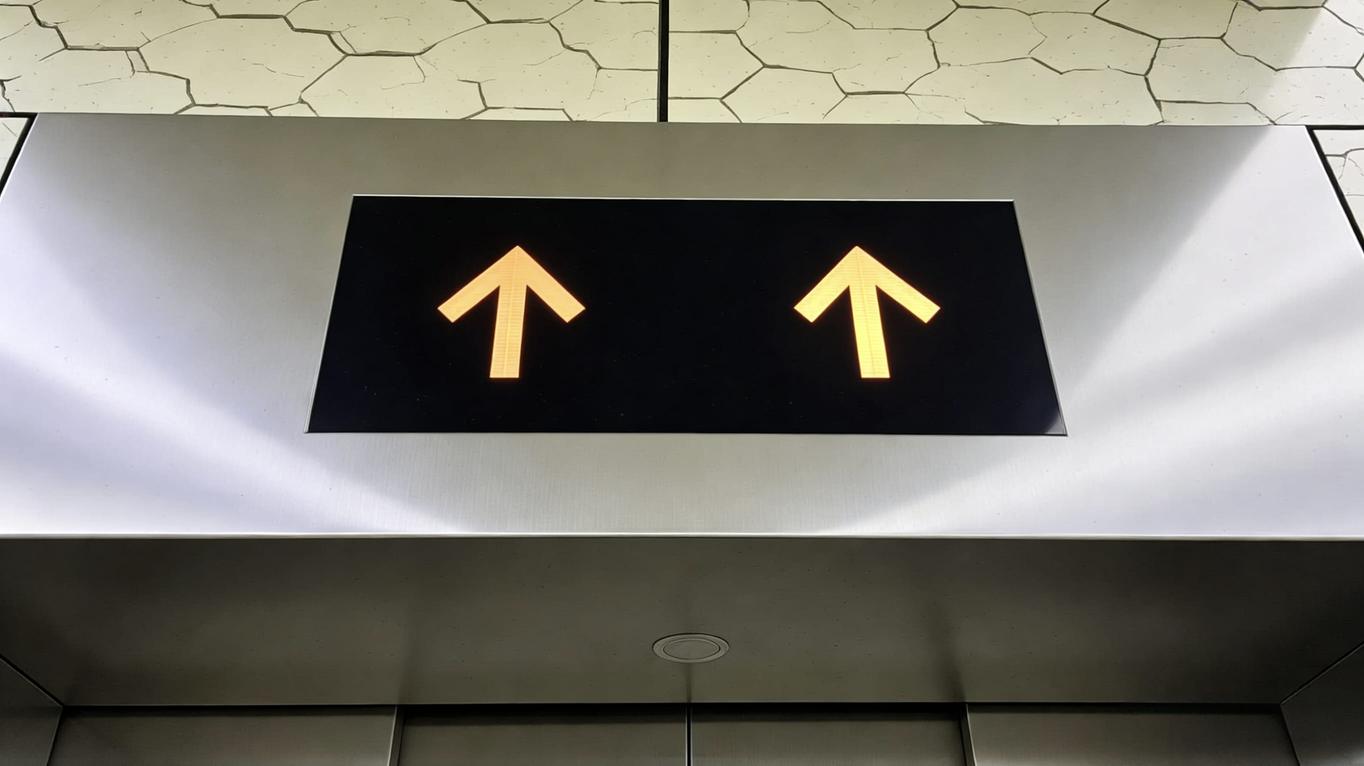}\par
    \smallskip 
    {\small\ttfamily\raggedright\textbf{Prompt:}\\ \input{\tTwoiHarm/OnePage/Label3/Text1.txt}\par}
  \end{minipage}%
\hfill%
  \begin{minipage}[t]{0.30\linewidth}
    \centering
    \includegraphics[width=\linewidth]{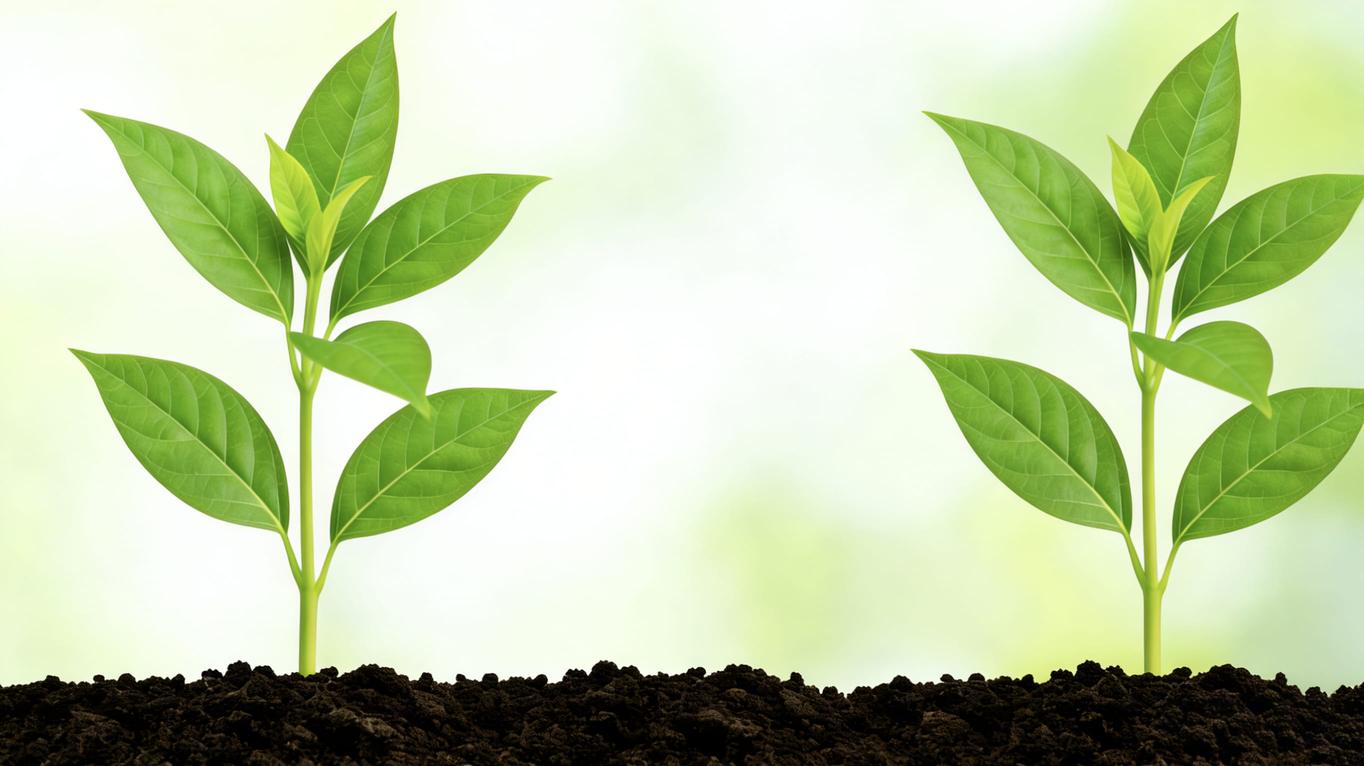}\par
    \smallskip 
    {\small\ttfamily\raggedright\textbf{Prompt:}\\ \input{\tTwoiHarm/OnePage/Label3/Text2.txt}\par}
  \end{minipage}%
\hfill%
  \begin{minipage}[t]{0.30\linewidth}
    \centering
    \includegraphics[width=\linewidth]{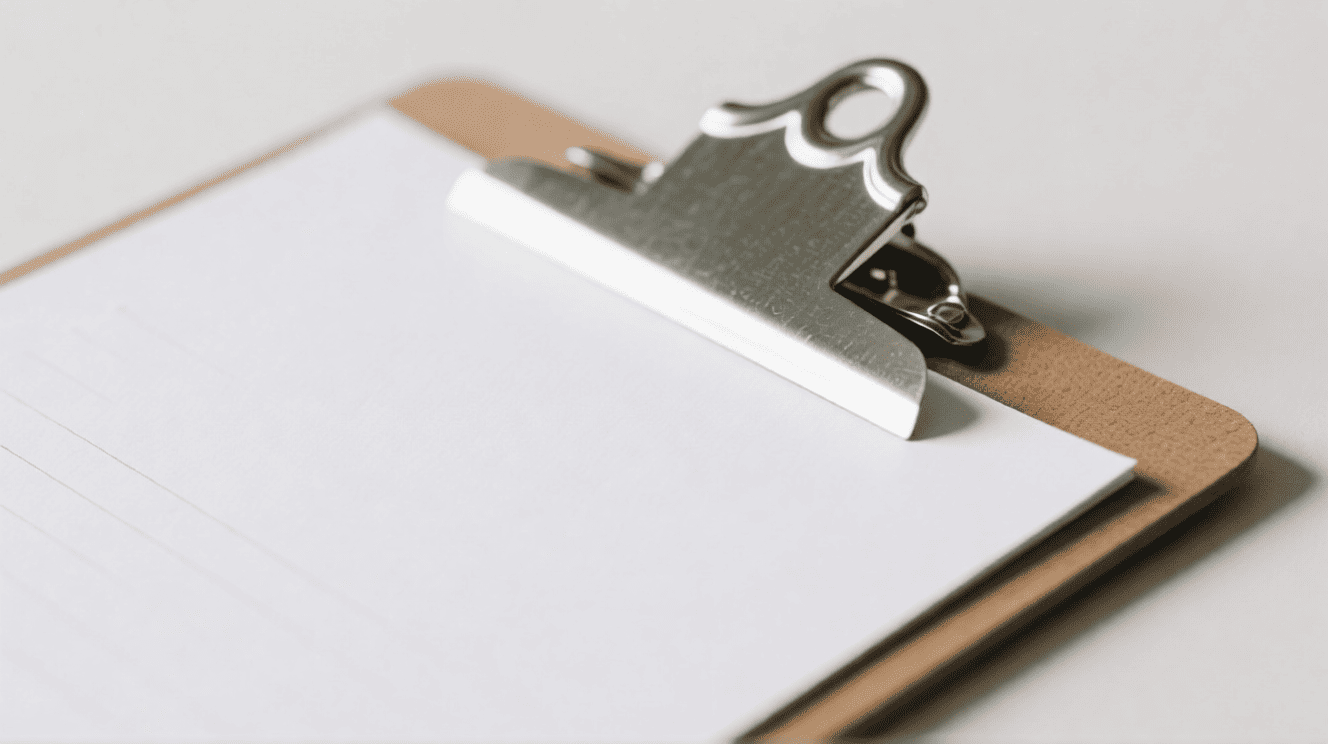}\par
    \smallskip 
    {\small\ttfamily\raggedright\textbf{Prompt:}\\ \input{\tTwoiHarm/OnePage/Label3/Text3.txt}\par}
  \end{minipage}%

\bigskip

\medskip \hrule height 0.8pt
\caption{Representative T2I examples illustrating order-to-space effects. Row~1 shows correct spatial grounding. Row~2 demonstrates persistent left/right reversals even when spatial constraints are explicitly specified in the prompt. Row~3 contains invalid or unscorable cases excluded from evaluation.}
\label{fig:t2i_order_to_space}
\end{figure*}

\begin{figure*}[t]
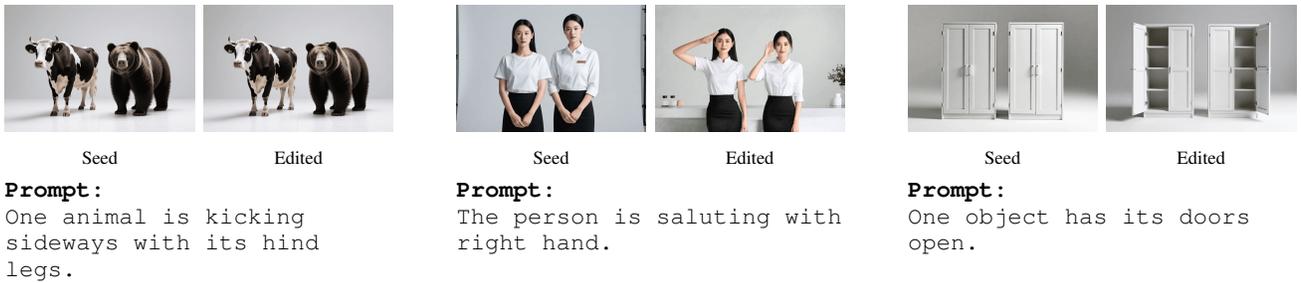

\centering
\hrule height 0.8pt \medskip

\noindent{\large \textbf{Examples for I2I Homogenization Evaluation}}\par
\bigskip

\sectionHeader{Output 1: A edited}{Valid. Edited entity corresponds to \textbf{Subject A} (left in the seed image).}
\noindent
  \begin{minipage}[t]{0.30\linewidth}
    \centering
    \begin{tabular}{@{}c@{\hspace{1.2mm}}c@{}}
      \includegraphics[width=0.49\linewidth]{\iTwoiDiv/Label1/Seed1.png} &
      \includegraphics[width=0.49\linewidth]{\iTwoiDiv/Label1/Image1.png} \\
      {\scriptsize Seed} & {\scriptsize Edited}
    \end{tabular}\par
    \smallskip
    {\small\ttfamily\raggedright\textbf{Prompt:}\\ \input{\iTwoiDiv/Label1/text1.txt}\par}
  \end{minipage}%
\hfill%
  \begin{minipage}[t]{0.30\linewidth}
    \centering
    \begin{tabular}{@{}c@{\hspace{1.2mm}}c@{}}
      \includegraphics[width=0.49\linewidth]{\iTwoiDiv/Label1/Seed2.png} &
      \includegraphics[width=0.49\linewidth]{\iTwoiDiv/Label1/Image2.png} \\
      {\scriptsize Seed} & {\scriptsize Edited}
    \end{tabular}\par
    \smallskip
    {\small\ttfamily\raggedright\textbf{Prompt:}\\ \input{\iTwoiDiv/Label1/text2.txt}\par}
  \end{minipage}%
\hfill%
  \begin{minipage}[t]{0.30\linewidth}
    \centering
    \begin{tabular}{@{}c@{\hspace{1.2mm}}c@{}}
      \includegraphics[width=0.49\linewidth]{\iTwoiDiv/Label1/Seed3.png} &
      \includegraphics[width=0.49\linewidth]{\iTwoiDiv/Label1/Image3.png} \\
      {\scriptsize Seed} & {\scriptsize Edited}
    \end{tabular}\par
    \smallskip
    {\small\ttfamily\raggedright\textbf{Prompt:}\\ \input{\iTwoiDiv/Label1/text3.txt}\par}
  \end{minipage}%

\bigskip

\sectionHeader{Output 2: B edited}{Valid. Edited entity corresponds to \textbf{Subject B} (right in the seed image).}
\noindent
  \begin{minipage}[t]{0.30\linewidth}
    \centering
    \begin{tabular}{@{}c@{\hspace{1.2mm}}c@{}}
      \includegraphics[width=0.49\linewidth]{\iTwoiDiv/Label2/Seed1.png} &
      \includegraphics[width=0.49\linewidth]{\iTwoiDiv/Label2/Image1.png} \\
      {\scriptsize Seed} & {\scriptsize Edited}
    \end{tabular}\par
    \smallskip
    {\small\ttfamily\raggedright\textbf{Prompt:}\\ \input{\iTwoiDiv/Label2/text1.txt}\par}
  \end{minipage}%
\hfill%
  \begin{minipage}[t]{0.30\linewidth}
    \centering
    \begin{tabular}{@{}c@{\hspace{1.2mm}}c@{}}
      \includegraphics[width=0.49\linewidth]{\iTwoiDiv/Label2/Seed2.png} &
      \includegraphics[width=0.49\linewidth]{\iTwoiDiv/Label2/Image2.png} \\
      {\scriptsize Seed} & {\scriptsize Edited}
    \end{tabular}\par
    \smallskip
    {\small\ttfamily\raggedright\textbf{Prompt:}\\ \input{\iTwoiDiv/Label2/text2.txt}\par}
  \end{minipage}%
\hfill%
  \begin{minipage}[t]{0.30\linewidth}
    \centering
    \begin{tabular}{@{}c@{\hspace{1.2mm}}c@{}}
      \includegraphics[width=0.49\linewidth]{\iTwoiDiv/Label2/Seed3.png} &
      \includegraphics[width=0.49\linewidth]{\iTwoiDiv/Label2/Image3.png} \\
      {\scriptsize Seed} & {\scriptsize Edited}
    \end{tabular}\par
    \smallskip
    {\small\ttfamily\raggedright\textbf{Prompt:}\\ \input{\iTwoiDiv/Label2/text3.txt}\par}
  \end{minipage}%

\bigskip

\sectionHeader{Output 3: Invalid}{Excluded from scoring (e.g., missing/extra entities, both/neither edited, identity confusion, or ambiguous evidence).}
\noindent
  \begin{minipage}[t]{0.30\linewidth}
    \centering
    \begin{tabular}{@{}c@{\hspace{1.2mm}}c@{}}
      \includegraphics[width=0.49\linewidth]{\iTwoiDiv/Label3/Seed1.png} &
      \includegraphics[width=0.49\linewidth]{\iTwoiDiv/Label3/Image1.png} \\
      {\scriptsize Seed} & {\scriptsize Edited}
    \end{tabular}\par
    \smallskip
    {\small\ttfamily\raggedright\textbf{Prompt:}\\ \input{\iTwoiDiv/Label3/text1.txt}\par}
  \end{minipage}%
\hfill%
  \begin{minipage}[t]{0.30\linewidth}
    \centering
    \begin{tabular}{@{}c@{\hspace{1.2mm}}c@{}}
      \includegraphics[width=0.49\linewidth]{\iTwoiDiv/Label3/Seed2.png} &
      \includegraphics[width=0.49\linewidth]{\iTwoiDiv/Label3/Image2.png} \\
      {\scriptsize Seed} & {\scriptsize Edited}
    \end{tabular}\par
    \smallskip
    {\small\ttfamily\raggedright\textbf{Prompt:}\\ \input{\iTwoiDiv/Label3/text2.txt}\par}
  \end{minipage}%
\hfill%
  \begin{minipage}[t]{0.30\linewidth}
    \centering
    \begin{tabular}{@{}c@{\hspace{1.2mm}}c@{}}
      \includegraphics[width=0.49\linewidth]{\iTwoiDiv/Label3/Seed3.png} &
      \includegraphics[width=0.49\linewidth]{\iTwoiDiv/Label3/Image3.png} \\
      {\scriptsize Seed} & {\scriptsize Edited}
    \end{tabular}\par
    \smallskip
    {\small\ttfamily\raggedright\textbf{Prompt:}\\ \input{\iTwoiDiv/Label3/text3.txt}\par}
  \end{minipage}%

\medskip \hrule height 0.8pt
\caption{Representative outputs for \textbf{I2I homogenization} labeling. In each case, the left image is the Seed and the right image is the Edited result. Output 1/2 correspond to valid cases where the under-specified edit is assigned to Subject A vs.\ Subject B, while Output 3 illustrates invalid generations excluded from scoring.}
\label{fig:i2i_diversity_examples}
\end{figure*}

\begin{figure*}[t]
\centering
\hrule height 0.8pt \medskip


\noindent{\large \textbf{Examples for I2I Correctness Evaluation (Aligned)}}\par
\bigskip

\sectionHeader{Label 1: Correct (Aligned)}{The generated image follows the prompt-specified action/role assignment and is consistent with the seed image grounding.}
\noindent
  \begin{minipage}[t]{0.30\linewidth}
    \centering
    \begin{tabular}{@{}c@{\hspace{1.2mm}}c@{}}
      \includegraphics[width=0.49\linewidth]{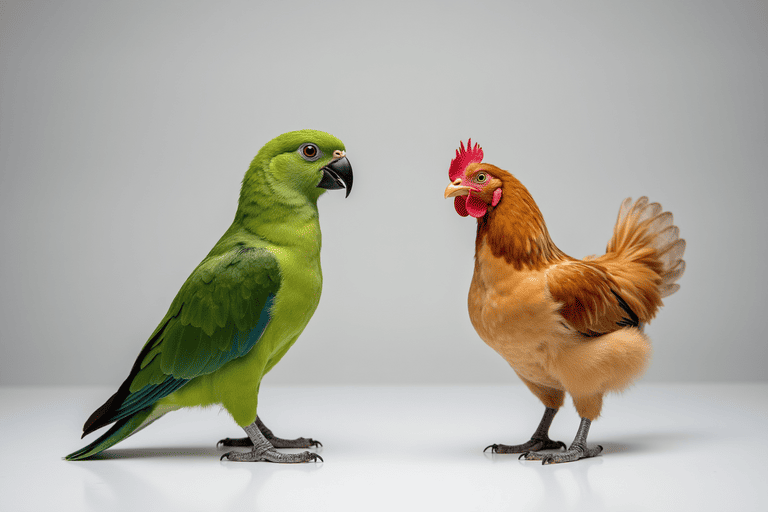} &
      \includegraphics[width=0.49\linewidth]{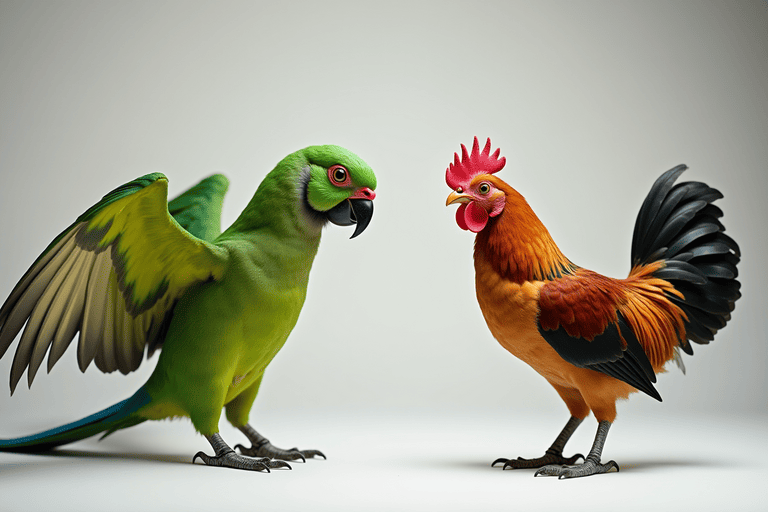} \\
      {\scriptsize Seed} & {\scriptsize Edited}
    \end{tabular}\par
    \smallskip
    {\small\ttfamily\raggedright\textbf{Prompt:}\\ \input{\iTwoiHarm/Align/Label1/text1.txt}\par}
  \end{minipage}%
\hfill%
  \begin{minipage}[t]{0.30\linewidth}
    \centering
    \begin{tabular}{@{}c@{\hspace{1.2mm}}c@{}}
      \includegraphics[width=0.49\linewidth]{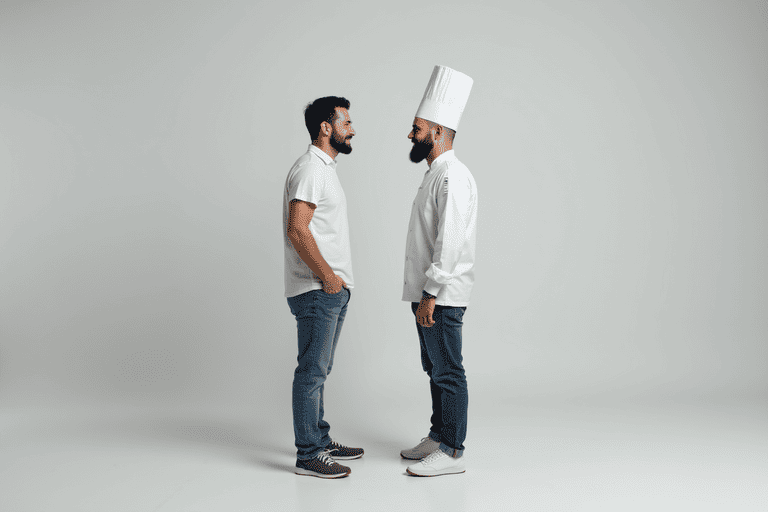} &
      \includegraphics[width=0.49\linewidth]{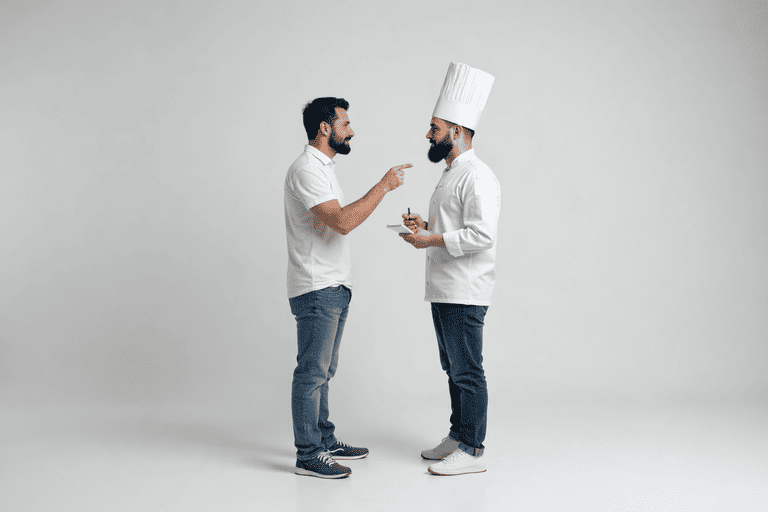} \\
      {\scriptsize Seed} & {\scriptsize Edited}
    \end{tabular}\par
    \smallskip
    {\small\ttfamily\raggedright\textbf{Prompt:}\\ \input{\iTwoiHarm/Align/Label1/text2.txt}\par}
  \end{minipage}%
\hfill%
  \begin{minipage}[t]{0.30\linewidth}
    \centering
    \begin{tabular}{@{}c@{\hspace{1.2mm}}c@{}}
      \includegraphics[width=0.49\linewidth]{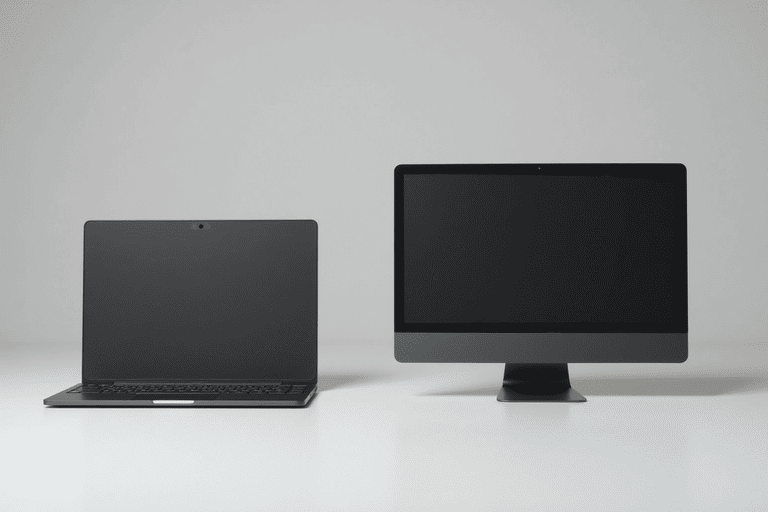} &
      \includegraphics[width=0.49\linewidth]{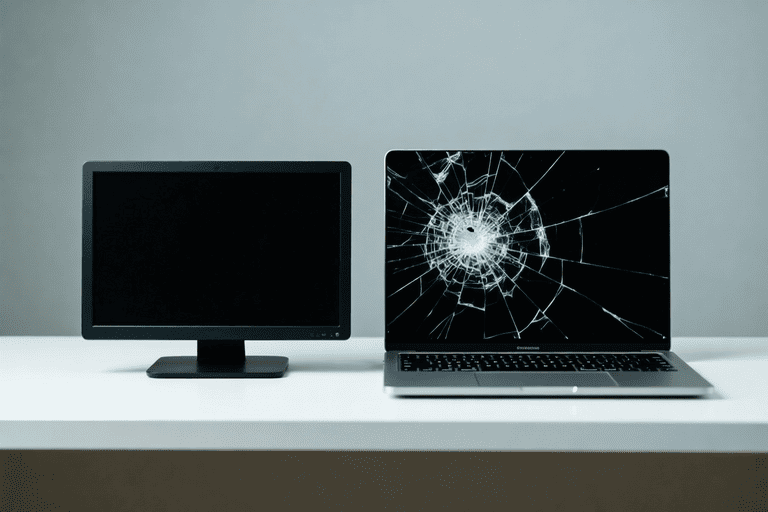} \\
      {\scriptsize Seed} & {\scriptsize Edited}
    \end{tabular}\par
    \smallskip
    {\small\ttfamily\raggedright\textbf{Prompt:}\\ \input{\iTwoiHarm/Align/Label1/text3.txt}\par}
  \end{minipage}%

\bigskip

\sectionHeader{Label 2: Incorrect}{The generated image swaps action/role assignments despite aligned prompt order, revealing Order-to-Space bias.}
\noindent
  \begin{minipage}[t]{0.30\linewidth}
    \centering
    \begin{tabular}{@{}c@{\hspace{1.2mm}}c@{}}
      \includegraphics[width=0.49\linewidth]{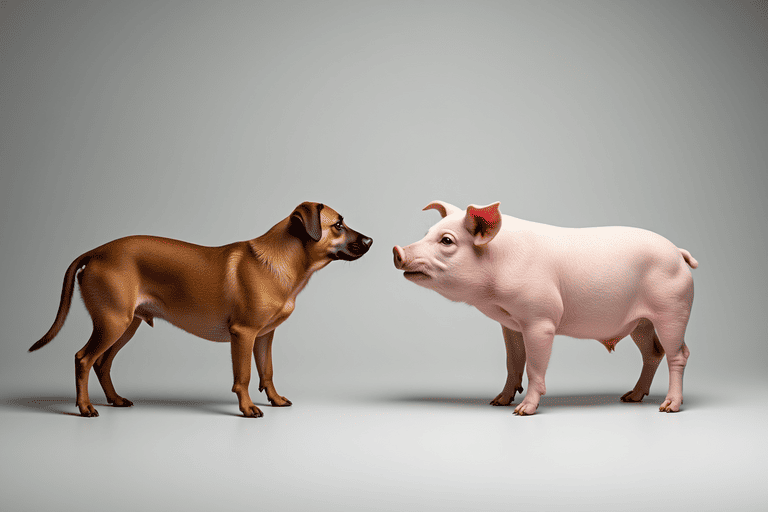} &
      \includegraphics[width=0.49\linewidth]{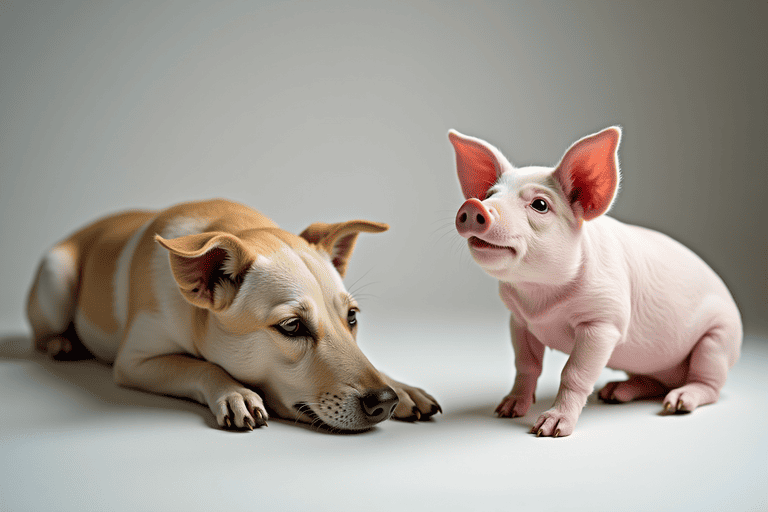} \\
      {\scriptsize Seed} & {\scriptsize Edited}
    \end{tabular}\par
    \smallskip
    {\small\ttfamily\raggedright\textbf{Prompt:}\\ \input{\iTwoiHarm/Align/Label2/text1.txt}\par}
  \end{minipage}%
\hfill%
  \begin{minipage}[t]{0.30\linewidth}
    \centering
    \begin{tabular}{@{}c@{\hspace{1.2mm}}c@{}}
      \includegraphics[width=0.49\linewidth]{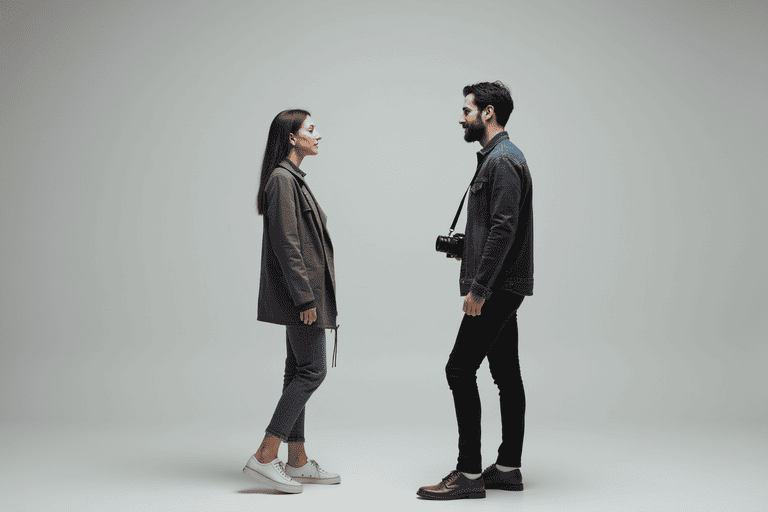} &
      \includegraphics[width=0.49\linewidth]{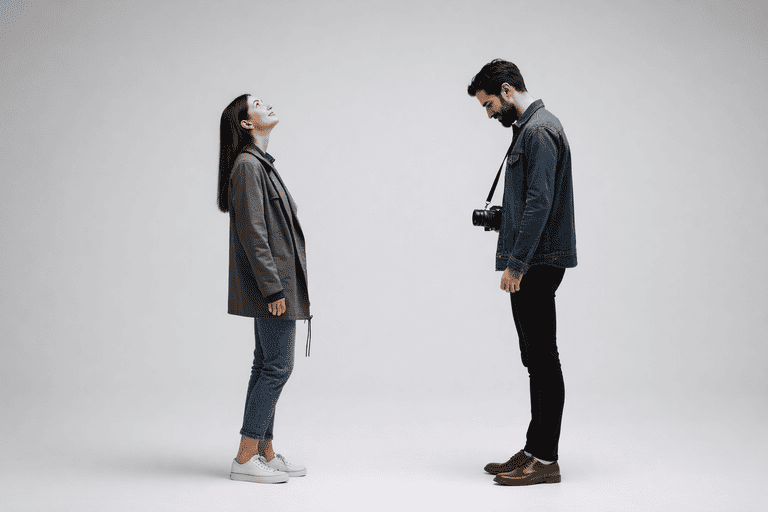} \\
      {\scriptsize Seed} & {\scriptsize Edited}
    \end{tabular}\par
    \smallskip
    {\small\ttfamily\raggedright\textbf{Prompt:}\\ \input{\iTwoiHarm/Align/Label2/text2.txt}\par}
  \end{minipage}%
\hfill%
  \begin{minipage}[t]{0.30\linewidth}
    \centering
    \begin{tabular}{@{}c@{\hspace{1.2mm}}c@{}}
      \includegraphics[width=0.49\linewidth]{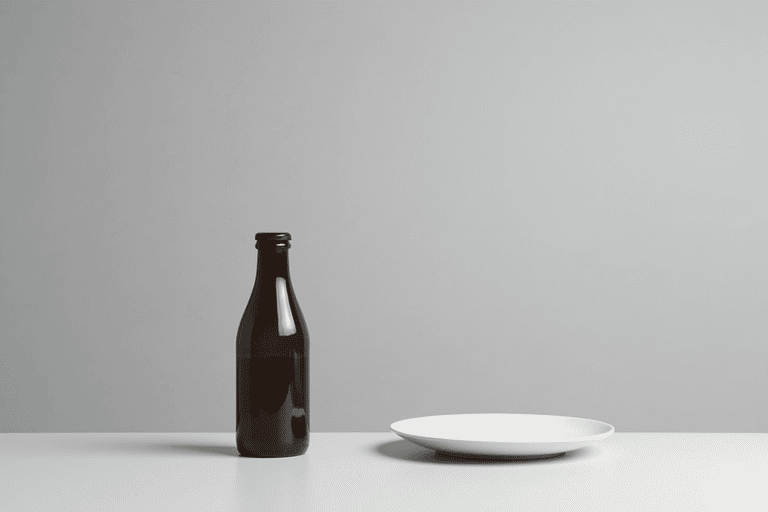} &
      \includegraphics[width=0.49\linewidth]{\iTwoiHarm/Align/Label2/Image3.png} \\
      {\scriptsize Seed} & {\scriptsize Edited}
    \end{tabular}\par
    \smallskip
    {\small\ttfamily\raggedright\textbf{Prompt:}\\ \input{\iTwoiHarm/Align/Label2/text3.txt}\par}
  \end{minipage}%

\bigskip

\sectionHeader{Label 3: Invalid}{Examples excluded from scoring due to ambiguous actions, identity confusion, or insufficient visual evidence.}
\noindent
  \begin{minipage}[t]{0.30\linewidth}
    \centering
    \begin{tabular}{@{}c@{\hspace{1.2mm}}c@{}}
      \includegraphics[width=0.49\linewidth]{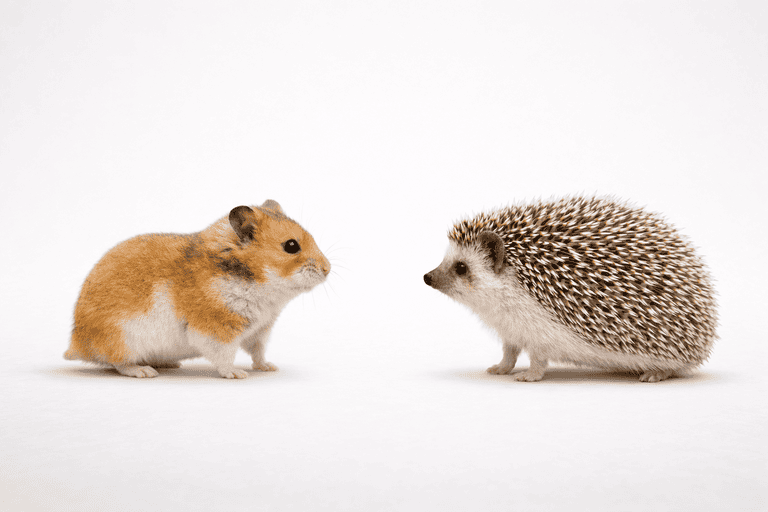} &
      \includegraphics[width=0.49\linewidth]{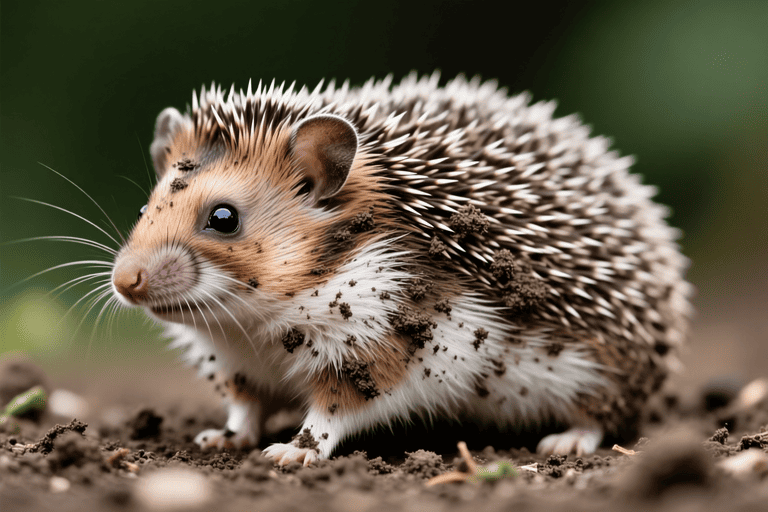} \\
      {\scriptsize Seed} & {\scriptsize Edited}
    \end{tabular}\par
    \smallskip
    {\small\ttfamily\raggedright\textbf{Prompt:}\\ \input{\iTwoiHarm/Align/Label3/text1.txt}\par}
  \end{minipage}%
\hfill%
  \begin{minipage}[t]{0.30\linewidth}
    \centering
    \begin{tabular}{@{}c@{\hspace{1.2mm}}c@{}}
      \includegraphics[width=0.49\linewidth]{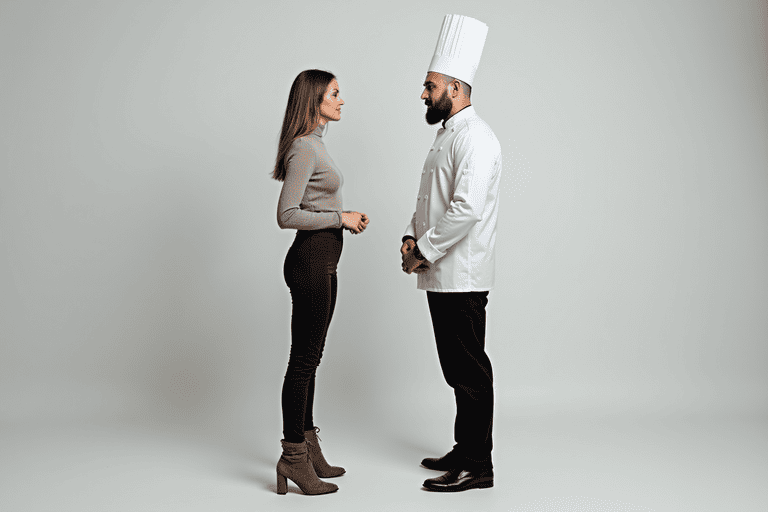} &
      \includegraphics[width=0.49\linewidth]{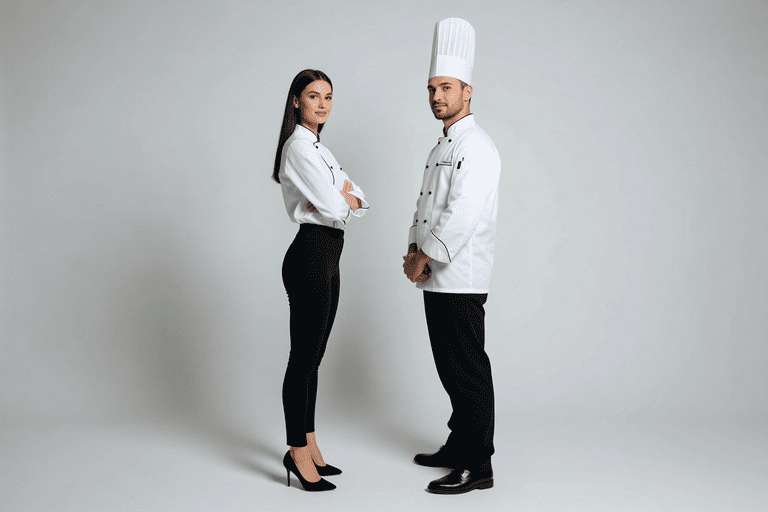} \\
      {\scriptsize Seed} & {\scriptsize Edited}
    \end{tabular}\par
    \smallskip
    {\small\ttfamily\raggedright\textbf{Prompt:}\\ \input{\iTwoiHarm/Align/Label3/text2.txt}\par}
  \end{minipage}%
\hfill%
  \begin{minipage}[t]{0.30\linewidth}
    \centering
    \begin{tabular}{@{}c@{\hspace{1.2mm}}c@{}}
      \includegraphics[width=0.49\linewidth]{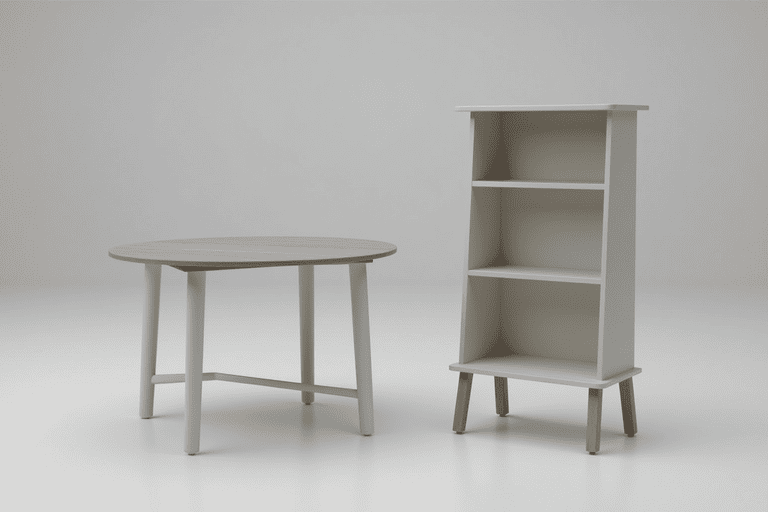} &
      \includegraphics[width=0.49\linewidth]{\iTwoiHarm/Align/Label3/Image3.png} \\
      {\scriptsize Seed} & {\scriptsize Edited}
    \end{tabular}\par
    \smallskip
    {\small\ttfamily\raggedright\textbf{Prompt:}\\ \input{\iTwoiHarm/Align/Label3/text3.txt}\par}
  \end{minipage}%

\medskip \hrule height 0.8pt
\caption{Representative examples for \textbf{I2I Harm (Aligned)} evaluation. In each case, the left image is the Seed and the right image is the Generated/Edited result under aligned prompt order. Label 1: correct attribution; Label 2: order-induced role/action swapping; Label 3: invalid generations excluded from scoring.}
\label{fig:i2i_harm_aligned}
\end{figure*}

\begin{figure*}[t]
\centering
\hrule height 0.8pt \medskip

\noindent{\large \textbf{Examples for I2I Harm Evaluation (Reverse)}}\par
\bigskip

\sectionHeader{Label 1: Correct (Reverse)}{The generated image follows the prompt-specified action/role assignment and is consistent with the seed image grounding.}
\noindent
  \begin{minipage}[t]{0.30\linewidth}
    \centering
    \begin{tabular}{@{}c@{\hspace{1.2mm}}c@{}}
      \includegraphics[width=0.49\linewidth]{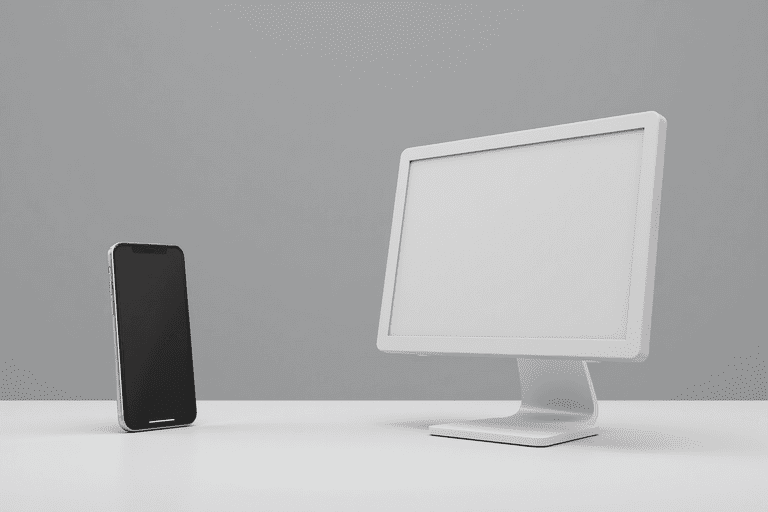} &
      \includegraphics[width=0.49\linewidth]{\iTwoiHarm/Reverse/Label1/Image1.png} \\
      {\scriptsize Seed} & {\scriptsize Edited}
    \end{tabular}\par
    \smallskip
    {\small\ttfamily\raggedright\textbf{Prompt:}\\ \input{\iTwoiHarm/Reverse/Label1/text1.txt}\par}
  \end{minipage}%
\hfill%
  \begin{minipage}[t]{0.30\linewidth}
    \centering
    \begin{tabular}{@{}c@{\hspace{1.2mm}}c@{}}
      \includegraphics[width=0.49\linewidth]{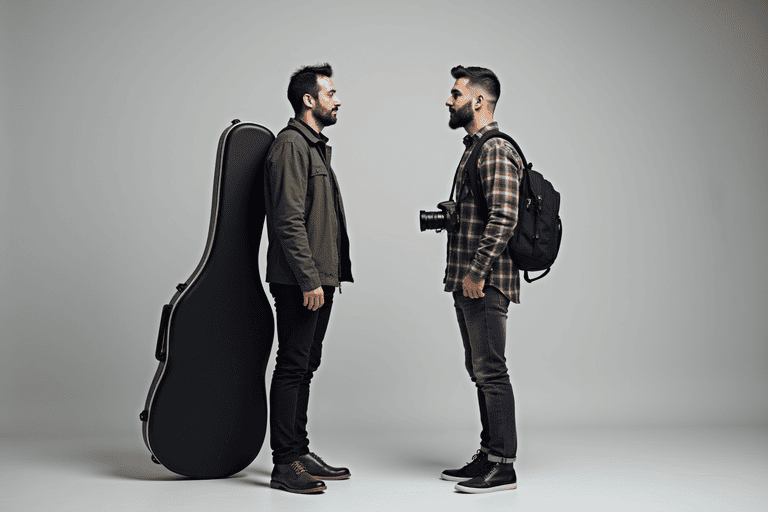} &
      \includegraphics[width=0.49\linewidth]{\iTwoiHarm/Reverse/Label1/Image2.png} \\
      {\scriptsize Seed} & {\scriptsize Edited}
    \end{tabular}\par
    \smallskip
    {\small\ttfamily\raggedright\textbf{Prompt:}\\ \input{\iTwoiHarm/Reverse/Label1/text2.txt}\par}
  \end{minipage}%
\hfill%
  \begin{minipage}[t]{0.30\linewidth}
    \centering
    \begin{tabular}{@{}c@{\hspace{1.2mm}}c@{}}
      \includegraphics[width=0.49\linewidth]{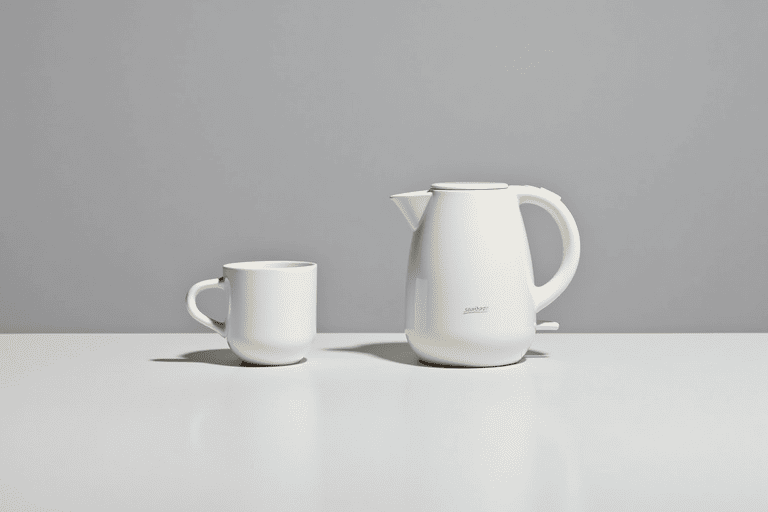} &
      \includegraphics[width=0.49\linewidth]{\iTwoiHarm/Reverse/Label1/Image3.png} \\
      {\scriptsize Seed} & {\scriptsize Edited}
    \end{tabular}\par
    \smallskip
    {\small\ttfamily\raggedright\textbf{Prompt:}\\ \input{\iTwoiHarm/Reverse/Label1/text3.txt}\par}
  \end{minipage}%

\bigskip

\sectionHeader{Label 2: Incorrect (Order-Induced Error)}{The generated image swaps action/role assignments despite reverse prompt order, revealing Order-to-Space bias.}
\noindent
  \begin{minipage}[t]{0.30\linewidth}
    \centering
    \begin{tabular}{@{}c@{\hspace{1.2mm}}c@{}}
      \includegraphics[width=0.49\linewidth]{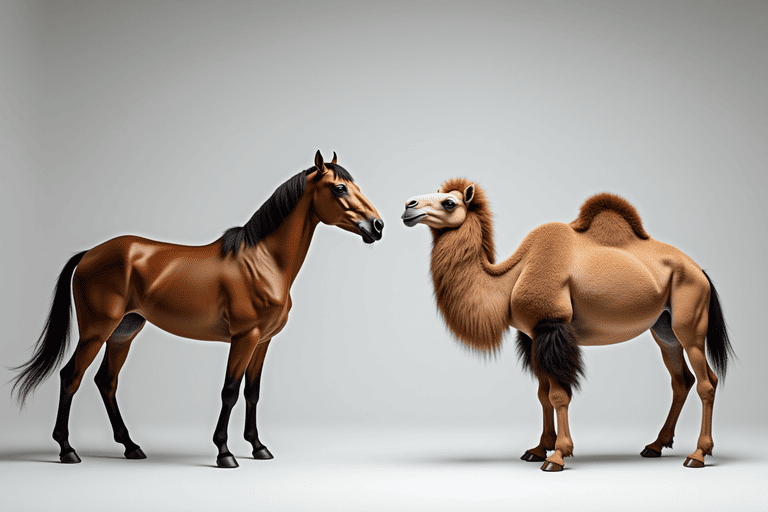} &
      \includegraphics[width=0.49\linewidth]{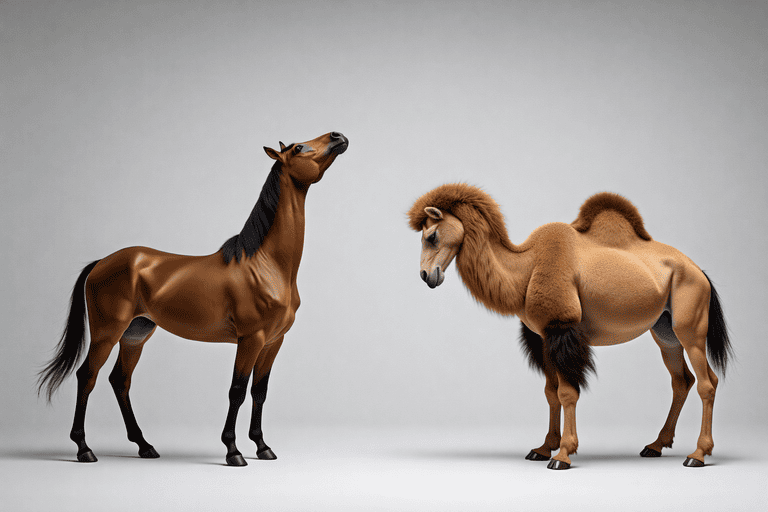} \\
      {\scriptsize Seed} & {\scriptsize Edited}
    \end{tabular}\par
    \smallskip
    {\small\ttfamily\raggedright\textbf{Prompt:}\\ \input{\iTwoiHarm/Reverse/Label2/text1.txt}\par}
  \end{minipage}%
\hfill%
  \begin{minipage}[t]{0.30\linewidth}
    \centering
    \begin{tabular}{@{}c@{\hspace{1.2mm}}c@{}}
      \includegraphics[width=0.49\linewidth]{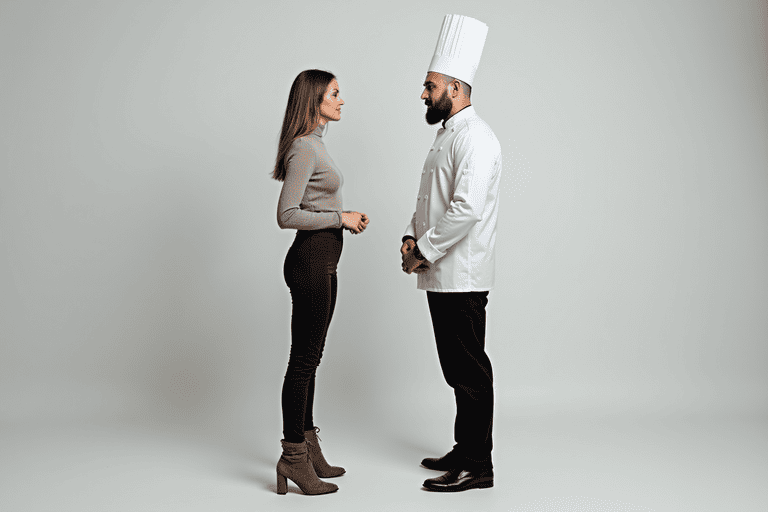} &
      \includegraphics[width=0.49\linewidth]{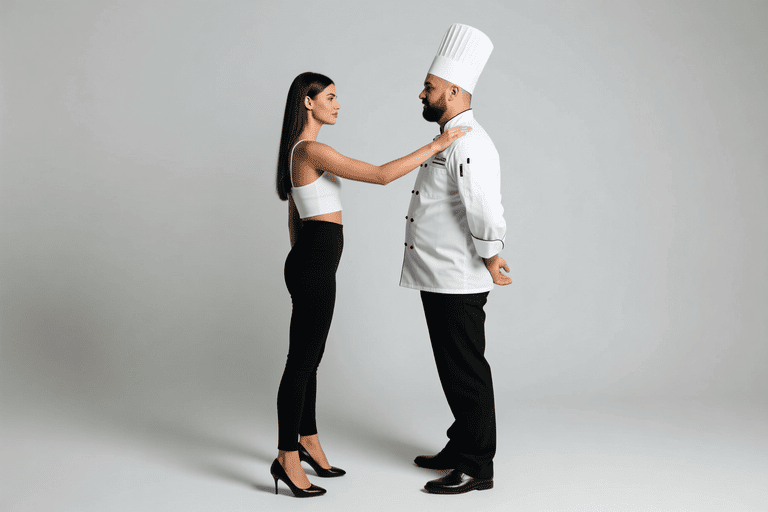} \\
      {\scriptsize Seed} & {\scriptsize Edited}
    \end{tabular}\par
    \smallskip
    {\small\ttfamily\raggedright\textbf{Prompt:}\\ \input{\iTwoiHarm/Reverse/Label2/text2.txt}\par}
  \end{minipage}%
\hfill%
  \begin{minipage}[t]{0.30\linewidth}
    \centering
    \begin{tabular}{@{}c@{\hspace{1.2mm}}c@{}}
      \includegraphics[width=0.49\linewidth]{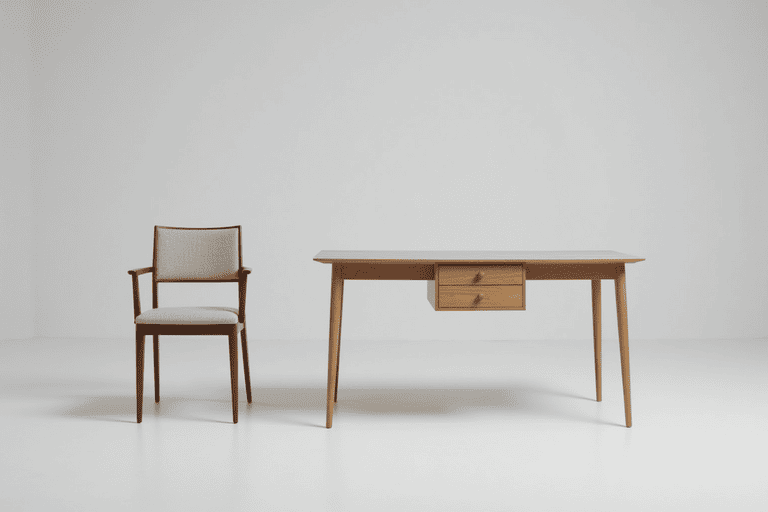} &
      \includegraphics[width=0.49\linewidth]{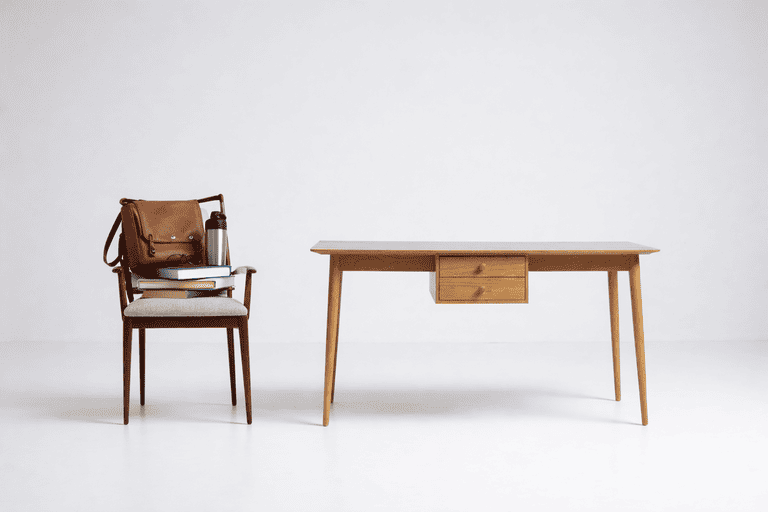} \\
      {\scriptsize Seed} & {\scriptsize Edited}
    \end{tabular}\par
    \smallskip
    {\small\ttfamily\raggedright\textbf{Prompt:}\\ \input{\iTwoiHarm/Reverse/Label2/text3.txt}\par}
  \end{minipage}%

\bigskip

\sectionHeader{Label 3: Invalid}{Examples excluded from scoring due to ambiguous actions, identity confusion, or insufficient visual evidence.}
\noindent
  \begin{minipage}[t]{0.30\linewidth}
    \centering
    \begin{tabular}{@{}c@{\hspace{1.2mm}}c@{}}
      \includegraphics[width=0.49\linewidth]{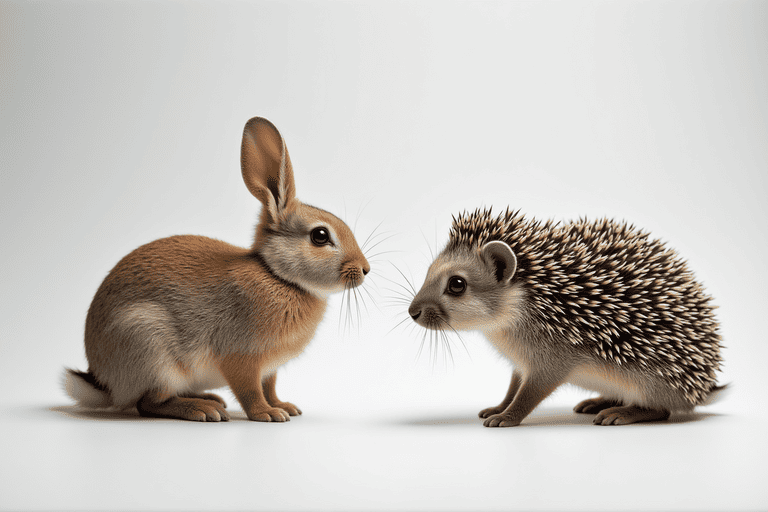} &
      \includegraphics[width=0.49\linewidth]{\iTwoiHarm/Reverse/Label3/Image1.png} \\
      {\scriptsize Seed} & {\scriptsize Edited}
    \end{tabular}\par
    \smallskip
    {\small\ttfamily\raggedright\textbf{Prompt:}\\ \input{\iTwoiHarm/Reverse/Label3/text1.txt}\par}
  \end{minipage}%
\hfill%
  \begin{minipage}[t]{0.30\linewidth}
    \centering
    \begin{tabular}{@{}c@{\hspace{1.2mm}}c@{}}
      \includegraphics[width=0.49\linewidth]{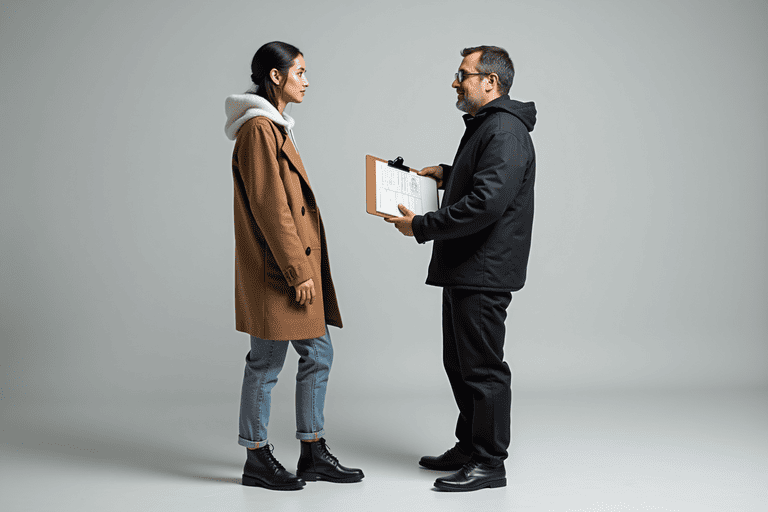} &
      \includegraphics[width=0.49\linewidth]{\iTwoiHarm/Reverse/Label3/Image2.png} \\
      {\scriptsize Seed} & {\scriptsize Edited}
    \end{tabular}\par
    \smallskip
    {\small\ttfamily\raggedright\textbf{Prompt:}\\ \input{\iTwoiHarm/Reverse/Label3/text2.txt}\par}
  \end{minipage}%
\hfill%
  \begin{minipage}[t]{0.30\linewidth}
    \centering
    \begin{tabular}{@{}c@{\hspace{1.2mm}}c@{}}
      \includegraphics[width=0.49\linewidth]{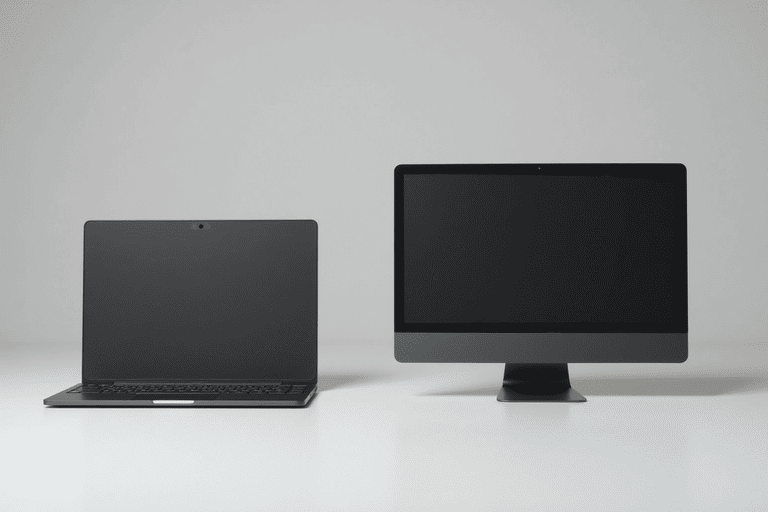} &
      \includegraphics[width=0.49\linewidth]{\iTwoiHarm/Reverse/Label3/Image3.png} \\
      {\scriptsize Seed} & {\scriptsize Edited}
    \end{tabular}\par
    \smallskip
    {\small\ttfamily\raggedright\textbf{Prompt:}\\ \input{\iTwoiHarm/Reverse/Label3/text3.txt}\par}
  \end{minipage}%

\medskip \hrule height 0.8pt
\caption{Representative examples for \textbf{I2I Harm (Reverse)} evaluation. In each case, the left image is the Seed and the right image is the Generated/Edited result under reverse prompt order. Label 1: correct attribution; Label 2: order-induced role/action swapping; Label 3: invalid generations excluded from scoring.}
\label{fig:i2i_harm_reverse}
\end{figure*}

\section{Human Evaluation for Image Quality}
\label{app:quality_eval}

We conduct a small-scale human study to verify that flip-based SFT mitigates OTS without degrading visual quality.
Annotators compare model variants on a held-out set of 400 matched generations per variant (same prompts across variants), presented in randomized order.
Each image is rated on a 1--7 scale (higher is better) along three axes:
\textit{Overall} (overall visual quality), \textit{Prompt Alignment} (how well the image matches the prompt intent), and
\textit{Fidelity} (visual realism and coherence without artifacts).
Table~\ref{tab:human_quality_singlecol} reports mean scores with standard deviations.

\begin{table*}[t]
\centering
\caption{Human evaluation of image quality on a held-out set of 400 matched outputs per model variant (1--7 scale; higher is better). We report mean with standard deviation in subscript.}
\label{tab:human_quality_singlecol}
\setlength{\tabcolsep}{5pt}
\renewcommand{\arraystretch}{1.12}
\small
\begin{tabular}{lccc}
\toprule
\textbf{Model} & \textbf{Overall} & \textbf{Prompt Align.} & \textbf{Fidelity} \\
\midrule
FLUX-dev
& $4.37_{\scriptsize\pm 1.36}$ & $3.69_{\scriptsize\pm 1.34}$ & $4.93_{\scriptsize\pm 1.48}$ \\
FLUX-dev (SFT)
& $4.65_{\scriptsize\pm 1.28}$ & $4.53_{\scriptsize\pm 1.31}$ & $4.86_{\scriptsize\pm 1.24}$ \\
\midrule
Qwen-Image
& $4.41_{\scriptsize\pm 1.19}$ & $3.32_{\scriptsize\pm 0.96}$ & $5.08_{\scriptsize\pm 1.42}$ \\
Qwen-Image (SFT)
& $4.79_{\scriptsize\pm 1.41}$ & $4.03_{\scriptsize\pm 1.45}$ & $5.24_{\scriptsize\pm 1.39}$ \\
\bottomrule
\end{tabular}
\end{table*}

\noindent Overall, human ratings show that SFT improves prompt alignment while preserving overall image quality and fidelity.

\newcommand{\tTwoVDiv}{case/t2v_diversity}
\newcommand{\tTwoVBlock}[2]{%
  \begin{minipage}[t]{0.48\linewidth}
    \centering
    \includegraphics[width=\linewidth,height=0.32\textheight,keepaspectratio]{#1}\par
    \vspace{0.3cm}
    {\small\ttfamily\raggedright\textbf{Prompt:}\\ \input{#2}\par}
  \end{minipage}%
}
\begin{figure*}[t]
\centering
\hrule height 0.8pt 
\vspace{0.4cm}
\noindent{\large \textbf{Extending Order-to-Space Effects to Text-to-Video Generation}}\par
\vspace{0.5cm}

\sectionHeader{Output 1: A before B}{Valid. The generation follows the first-mentioned entity preference under the original prompt order.}
\vspace{0.3cm}
\noindent
  \begin{minipage}[t]{0.48\linewidth}
    \centering
    \includegraphics[width=\linewidth,height=0.32\textheight,keepaspectratio]{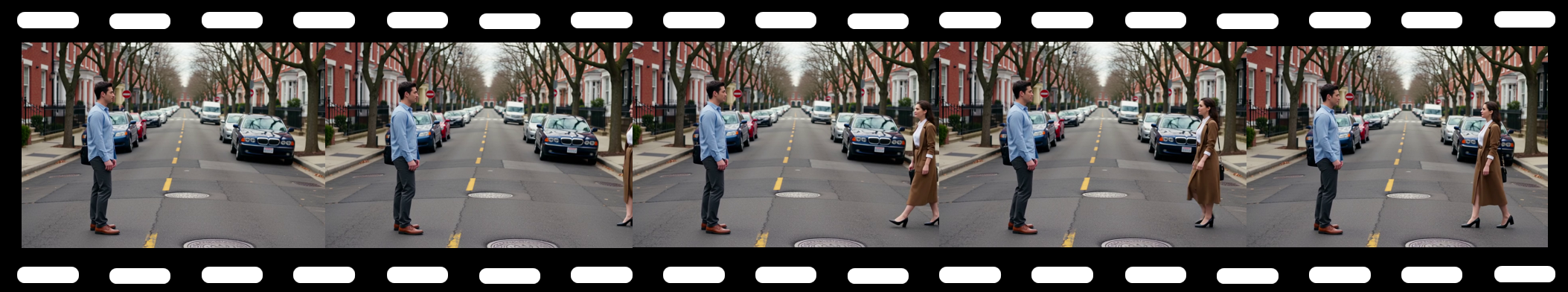}\par
    \vspace{0.3cm}
    {\small\ttfamily\raggedright\textbf{Prompt:}\\ \input{\tTwoVDiv/t2v1.txt}\par}
  \end{minipage}%
\hfill%
  \begin{minipage}[t]{0.48\linewidth}
    \centering
    \includegraphics[width=\linewidth,height=0.32\textheight,keepaspectratio]{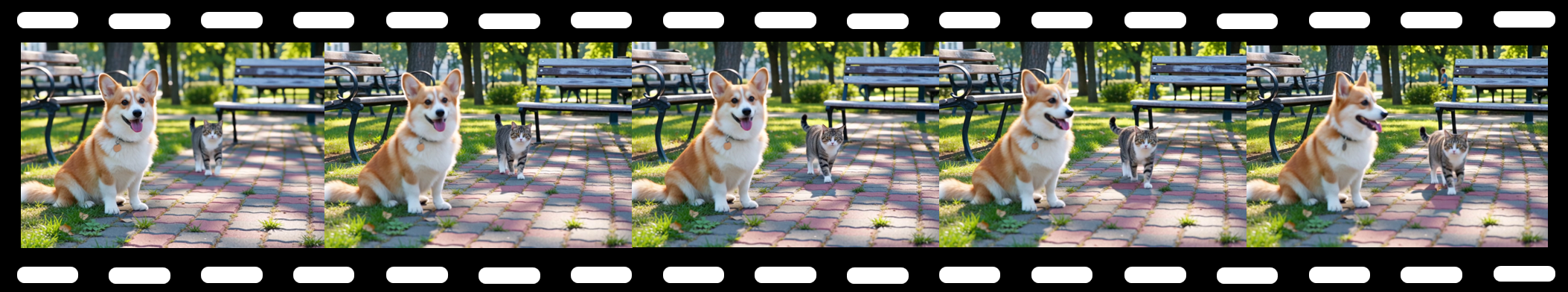}\par
    \vspace{0.3cm}
    {\small\ttfamily\raggedright\textbf{Prompt:}\\ \input{\tTwoVDiv/t2v2.txt}\par}
  \end{minipage}%

\vspace{0.6cm}

\sectionHeader{Output 2: B before A}{Valid. The generation follows the first-mentioned entity preference under the reversed prompt order.}
\vspace{0.3cm}
\noindent
  \begin{minipage}[t]{0.48\linewidth}
    \centering
    \includegraphics[width=\linewidth,height=0.32\textheight,keepaspectratio]{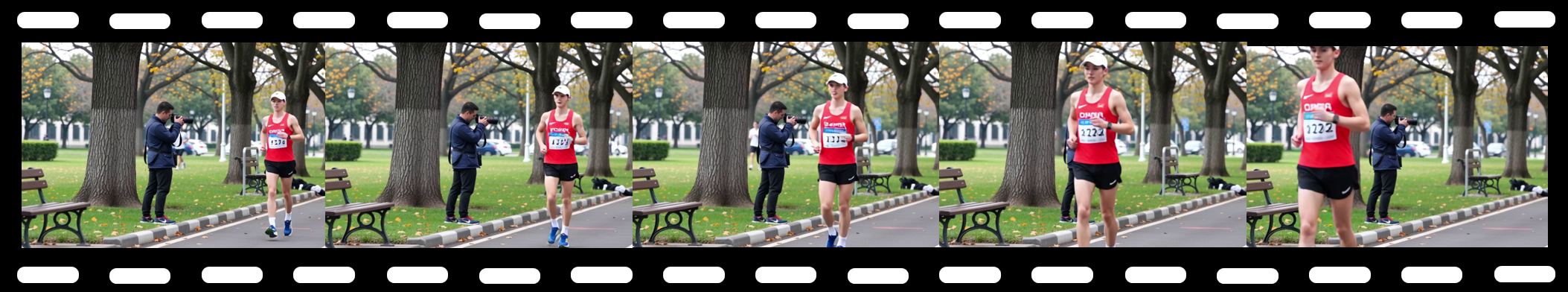}\par
    \vspace{0.3cm}
    {\small\ttfamily\raggedright\textbf{Prompt:}\\ \input{\tTwoVDiv/t2v3.txt}\par}
  \end{minipage}%
\hfill%
  \begin{minipage}[t]{0.48\linewidth}
    \centering
    \includegraphics[width=\linewidth,height=0.32\textheight,keepaspectratio]{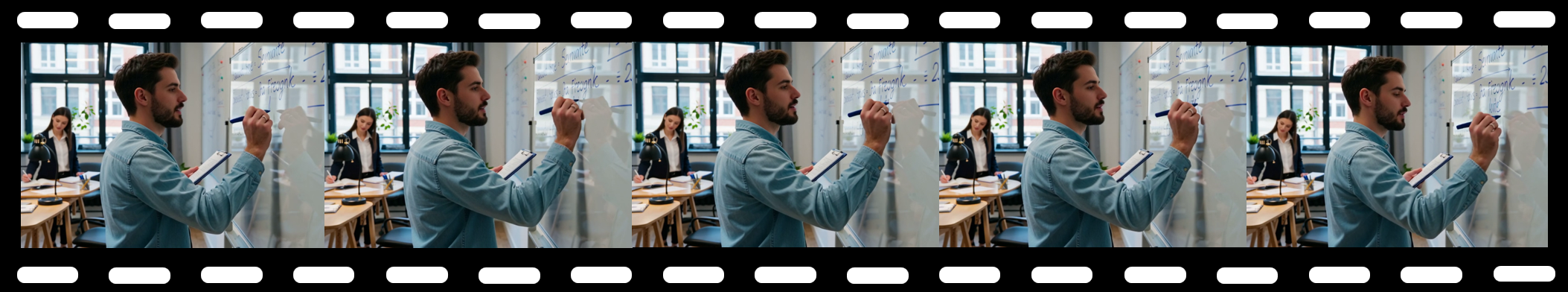}\par
    \vspace{0.3cm}
    {\small\ttfamily\raggedright\textbf{Prompt:}\\ \input{\tTwoVDiv/t2v4.txt}\par}
  \end{minipage}%

\vspace{0.6cm}

\sectionHeader{Output 3: Invalid}{Excluded from scoring (e.g., missing subjects, ambiguous events, or insufficient visual evidence).}
\vspace{0.3cm}
\noindent
  \begin{minipage}[t]{0.48\linewidth}
    \centering
    \includegraphics[width=\linewidth,height=0.32\textheight,keepaspectratio]{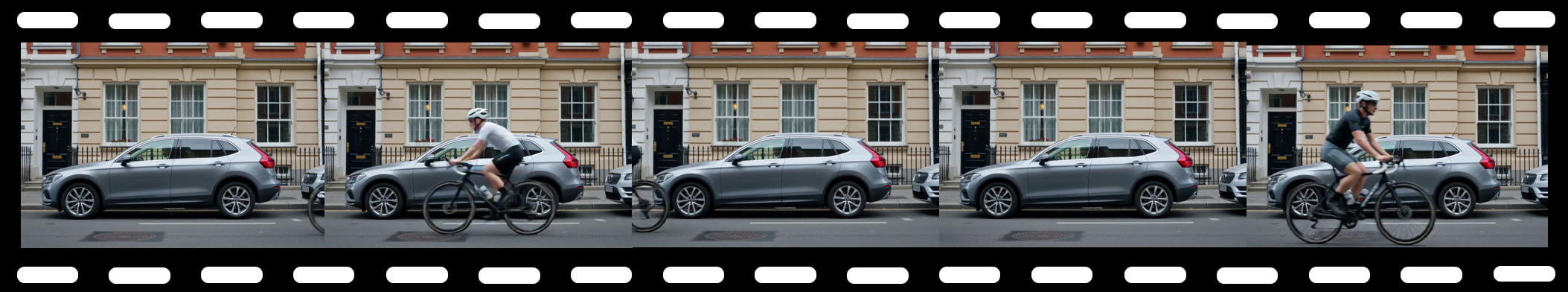}\par
    \vspace{0.3cm}
    {\small\ttfamily\raggedright\textbf{Prompt:}\\ \input{\tTwoVDiv/t2v5.txt}\par}
  \end{minipage}%
\hfill%
  \begin{minipage}[t]{0.48\linewidth}
    \centering
    \includegraphics[width=\linewidth,height=0.32\textheight,keepaspectratio]{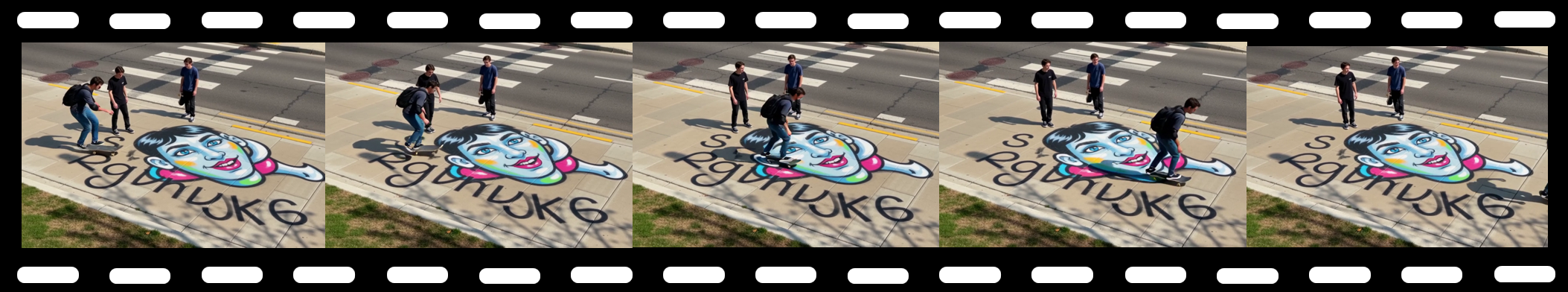}\par
    \vspace{0.3cm}
    {\small\ttfamily\raggedright\textbf{Prompt:}\\ \input{\tTwoVDiv/t2v6.txt}\par}
  \end{minipage}%

\vspace{0.4cm}

\hrule height 0.8pt
\vspace{0.2cm}
\caption{Representative outputs for \textbf{T2V homogenization} labeling. Output 1/2 correspond to valid generations under original/reversed prompt order, while Output 3 illustrates invalid generations excluded from scoring.}
\label{fig:t2v_homogenization_examples}
\end{figure*}


\newcommand{\iTwoVDir}{case/i2v_diversity}

\newcommand{\iTwoVBlockTwoCol}[3]{%
  \begin{minipage}[t]{0.48\linewidth}
    \centering
    \includegraphics[width=0.3\linewidth]{#1}\par
    \vspace{0.2em}
    {\footnotesize Seed}\par
    \vspace{0.4em}

    \includegraphics[width=\linewidth]{#2}\par
    \vspace{0.2em}
    {\footnotesize Video}\par
    \vspace{0.4em}

    {\small\ttfamily\raggedright\textbf{Prompt:}\\ #3\par}
  \end{minipage}%
}

\begin{figure*}[t]
\centering
\hrule height 0.8pt \medskip

\noindent{\large \textbf{Extending Order-to-Space Effects to Image-to-Video Generation}}\par
\bigskip

\sectionHeader{Output 1: A edited}
{Valid. The under-specified edit/action is assigned to \textbf{Subject A} (left in the seed image).}
\vspace{0.4em}

\noindent
\iTwoVBlockTwoCol{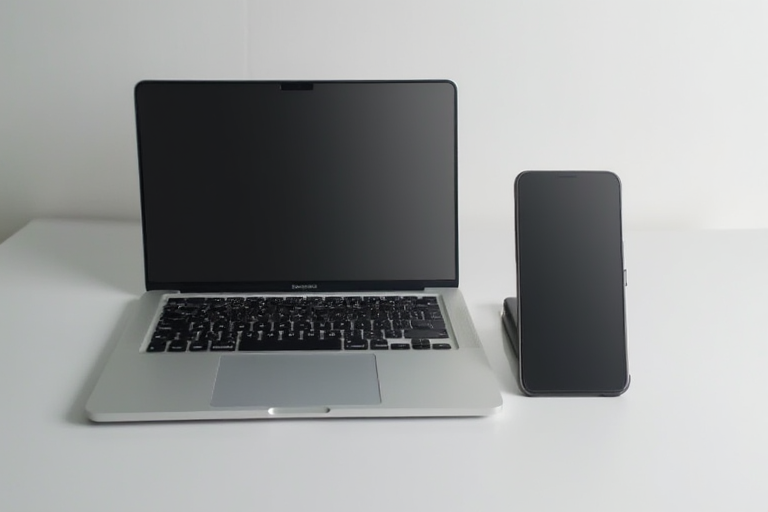}{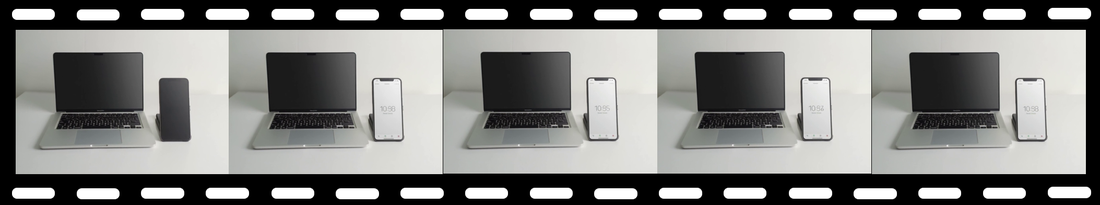}{\input{\iTwoVDir/Label1/Text1.txt}}\hfill%
\iTwoVBlockTwoCol{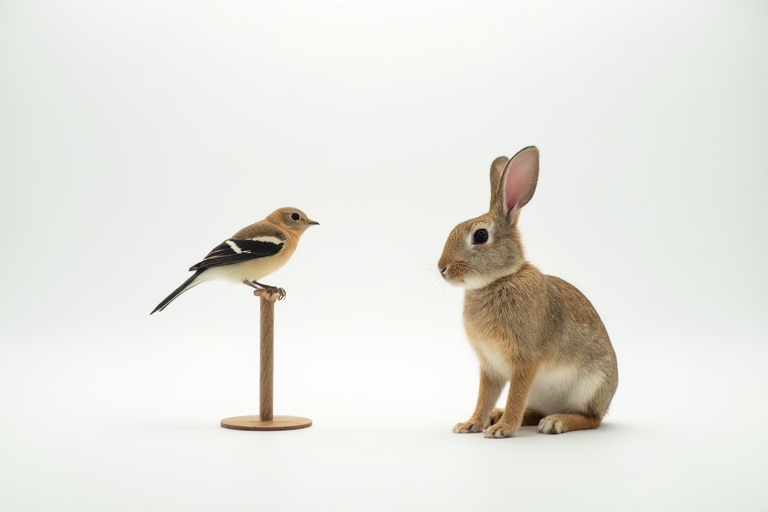}{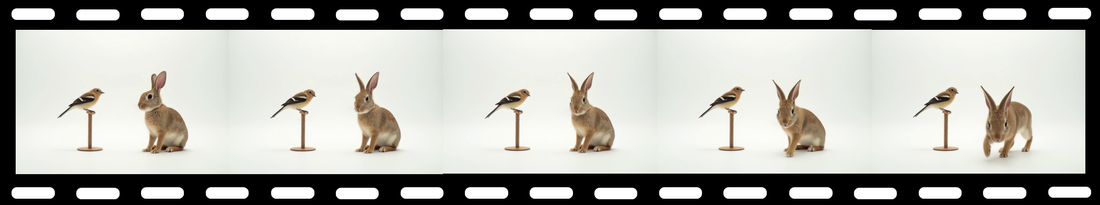}{\input{\iTwoVDir/Label1/Text2.txt}}

\bigskip

\sectionHeader{Output 2: B edited}
{Valid. The under-specified edit/action is assigned to \textbf{Subject B} (right in the seed image).}
\vspace{0.4em}

\noindent
\iTwoVBlockTwoCol{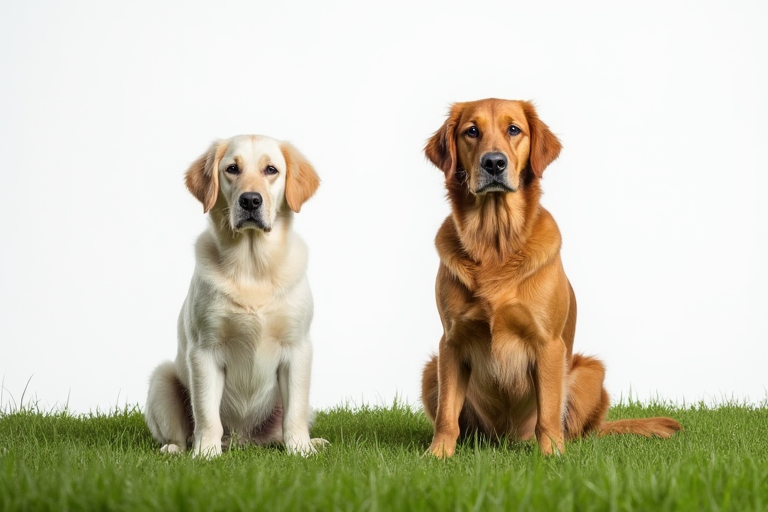}{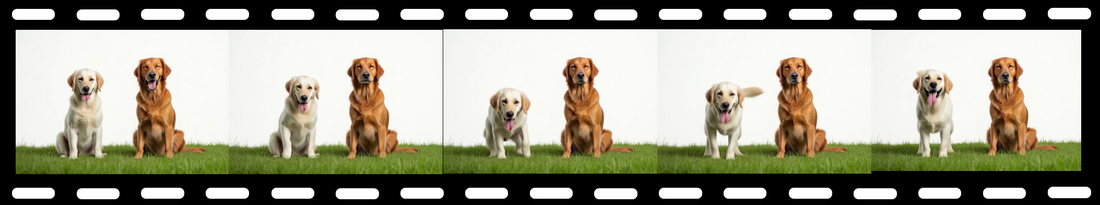}{\input{\iTwoVDir/Label2/Text1.txt}}\hfill%
\iTwoVBlockTwoCol{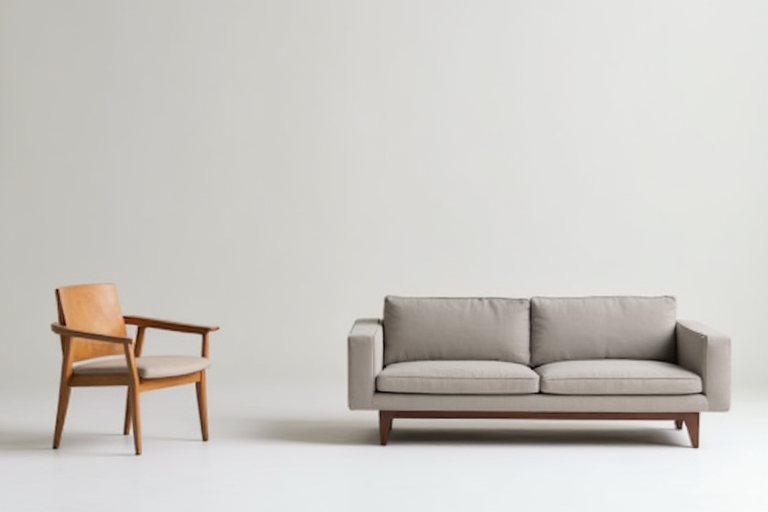}{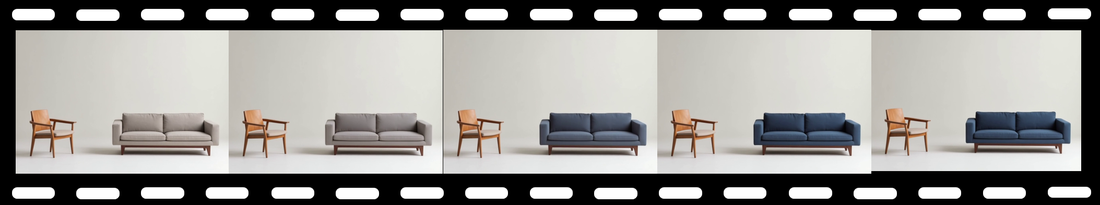}{\input{\iTwoVDir/Label2/Text2.txt}}

\bigskip

\sectionHeader{Output 3: Invalid}
{Excluded from scoring (e.g., missing/extra entities, both/neither edited, identity confusion, or ambiguous evidence).}
\vspace{0.4em}

\noindent
\iTwoVBlockTwoCol{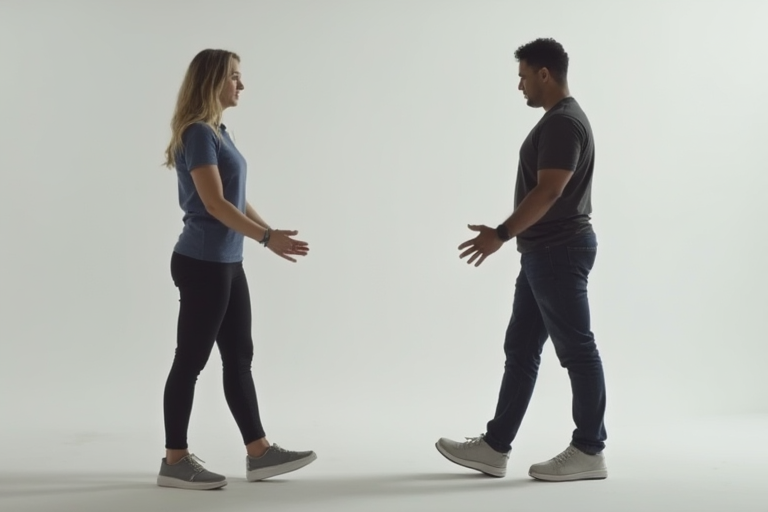}{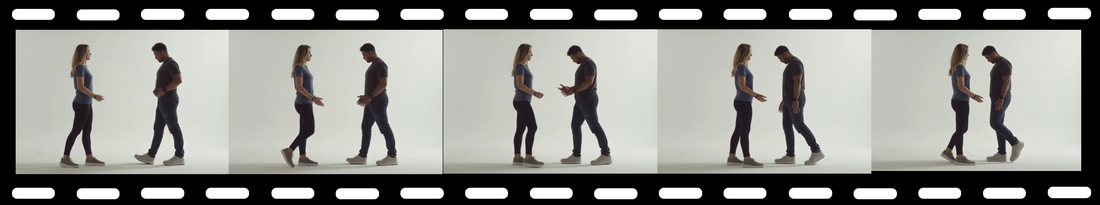}{\input{\iTwoVDir/Label3/Text1.txt}}\hfill%
\iTwoVBlockTwoCol{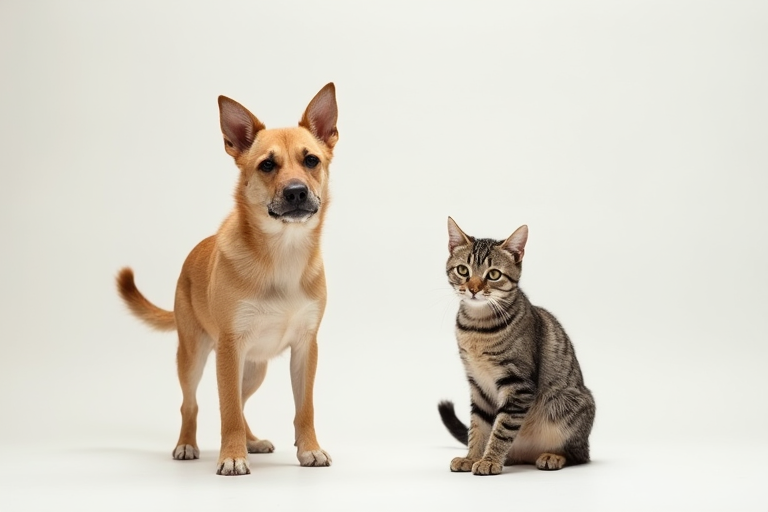}{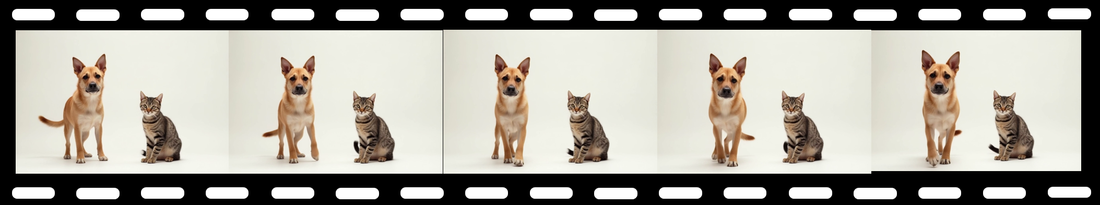}{\input{\iTwoVDir/Label3/Text2.txt}}

\medskip \hrule height 0.8pt
\caption{Representative outputs for \textbf{I2V homogenization} labeling. Each example shows the Seed frame (top) and a video preview (bottom). Output~1/2 are valid cases where an under-specified edit/action is assigned to Subject~A vs.\ Subject~B, while Output~3 illustrates invalid generations excluded from scoring.}
\label{fig:i2v_diversity_examples}
\end{figure*}


\newcommand{\VLUImgDir}{case/vlu_figs}
\newcommand{\VLUTextDir}{case/vlu_figs}

\newcommand{\VLUBlock}[2]{%
  \begin{minipage}[t]{0.31\linewidth}
    \centering
    \includegraphics[
      width=\linewidth,
      height=0.15\textheight,
      keepaspectratio
    ]{#1}\par
    \vspace{0.4em}

    {\footnotesize\ttfamily\raggedright
      \textbf{Model Answer:}\\
      \input{#2}\par
    }
  \end{minipage}%
}

\begin{figure*}[t]
\centering
\hrule height 0.8pt \medskip

\noindent{\large \textbf{Extending Order-to-Space Analysis to Vision–Language Understanding}}\par
\bigskip

\begin{minipage}{0.96\linewidth}
  {\small\bfseries System Prompt: }
  {\small\itshape Describe this image in one sentence.}
\end{minipage}
\par \vspace{0.8em}
\hrule height 0.8pt
\bigskip

\noindent
  \begin{minipage}[t]{0.31\linewidth}
    \centering
    \includegraphics[
      width=\linewidth,
      height=0.15\textheight,
      keepaspectratio
    ]{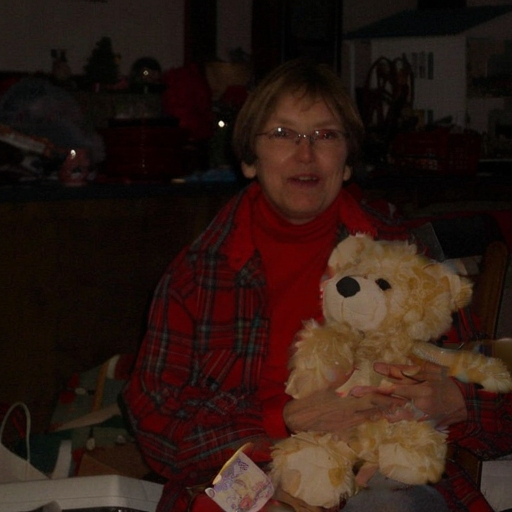}\par
    \vspace{0.4em}

    {\footnotesize\ttfamily\raggedright
      \textbf{Model Answer:}\\
      \input{\VLUTextDir/text1.txt}\par
    }
  \end{minipage}%
\hfill
  \begin{minipage}[t]{0.31\linewidth}
    \centering
    \includegraphics[
      width=\linewidth,
      height=0.15\textheight,
      keepaspectratio
    ]{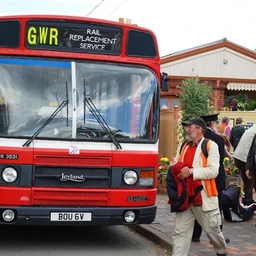}\par
    \vspace{0.4em}

    {\footnotesize\ttfamily\raggedright
      \textbf{Model Answer:}\\
      \input{\VLUTextDir/text2.txt}\par
    }
  \end{minipage}%
\hfill
  \begin{minipage}[t]{0.31\linewidth}
    \centering
    \includegraphics[
      width=\linewidth,
      height=0.15\textheight,
      keepaspectratio
    ]{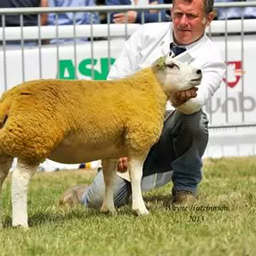}\par
    \vspace{0.4em}

    {\footnotesize\ttfamily\raggedright
      \textbf{Model Answer:}\\
      \input{\VLUTextDir/text3.txt}\par
    }
  \end{minipage}%

\vspace{6mm}

\noindent
  \begin{minipage}[t]{0.31\linewidth}
    \centering
    \includegraphics[
      width=\linewidth,
      height=0.15\textheight,
      keepaspectratio
    ]{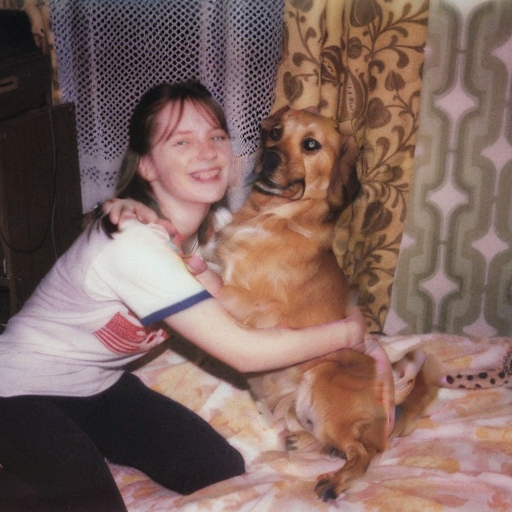}\par
    \vspace{0.4em}

    {\footnotesize\ttfamily\raggedright
      \textbf{Model Answer:}\\
      \input{\VLUTextDir/text4.txt}\par
    }
  \end{minipage}%
\hfill
  \begin{minipage}[t]{0.31\linewidth}
    \centering
    \includegraphics[
      width=\linewidth,
      height=0.15\textheight,
      keepaspectratio
    ]{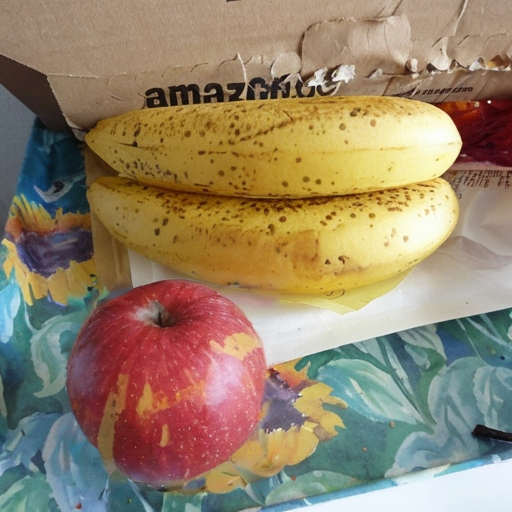}\par
    \vspace{0.4em}

    {\footnotesize\ttfamily\raggedright
      \textbf{Model Answer:}\\
      \input{\VLUTextDir/text5.txt}\par
    }
  \end{minipage}%
\hfill
  \begin{minipage}[t]{0.31\linewidth}
    \centering
    \includegraphics[
      width=\linewidth,
      height=0.15\textheight,
      keepaspectratio
    ]{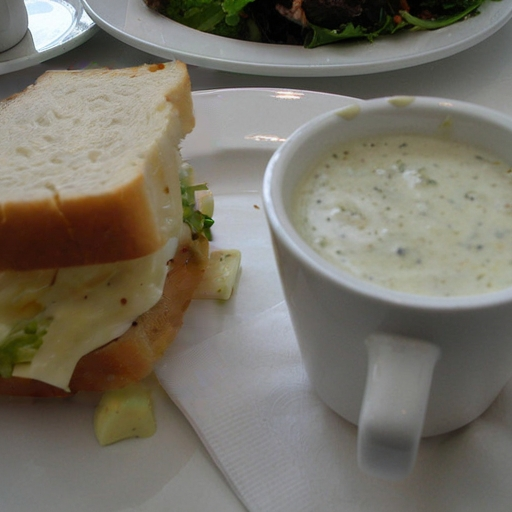}\par
    \vspace{0.4em}

    {\footnotesize\ttfamily\raggedright
      \textbf{Model Answer:}\\
      \input{\VLUTextDir/text6.txt}\par
    }
  \end{minipage}%

\vspace{6mm}

\noindent
  \begin{minipage}[t]{0.31\linewidth}
    \centering
    \includegraphics[
      width=\linewidth,
      height=0.15\textheight,
      keepaspectratio
    ]{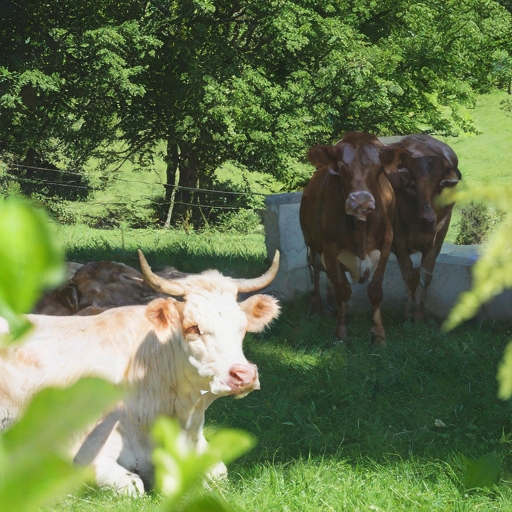}\par
    \vspace{0.4em}

    {\footnotesize\ttfamily\raggedright
      \textbf{Model Answer:}\\
      \input{\VLUTextDir/text7.txt}\par
    }
  \end{minipage}%
\hfill
  \begin{minipage}[t]{0.31\linewidth}
    \centering
    \includegraphics[
      width=\linewidth,
      height=0.15\textheight,
      keepaspectratio
    ]{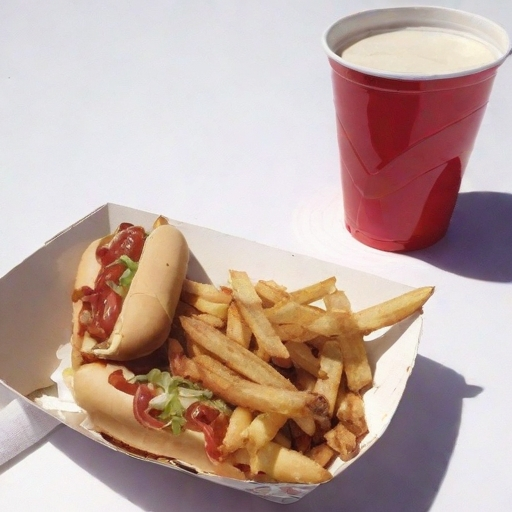}\par
    \vspace{0.4em}

    {\footnotesize\ttfamily\raggedright
      \textbf{Model Answer:}\\
      \input{\VLUTextDir/text8.txt}\par
    }
  \end{minipage}%
\hfill
  \begin{minipage}[t]{0.31\linewidth}
    \centering
    \includegraphics[
      width=\linewidth,
      height=0.15\textheight,
      keepaspectratio
    ]{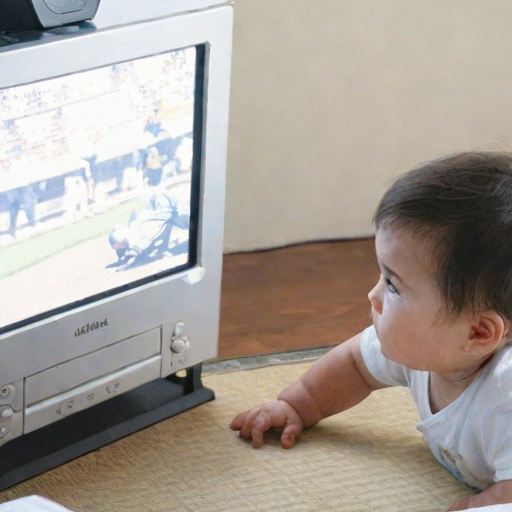}\par
    \vspace{0.4em}

    {\footnotesize\ttfamily\raggedright
      \textbf{Model Answer:}\\
      \input{\VLUTextDir/text9.txt}\par
    }
  \end{minipage}%

\medskip
\hrule height 0.8pt
\caption{Examples of vision-language understanding. Each case follows the same system prompt and reports the model's one-sentence description.}
\label{fig:vlu_understanding_examples}
\end{figure*}

\newcommand{\StepBlock}[2]{%
  \begin{minipage}[t]{0.18\linewidth}
    \centering
    \includegraphics[width=\linewidth]{#1}\par
    \vspace{0.3em}
    {\scriptsize\ttfamily #2}
  \end{minipage}%
}

\begin{figure*}[t]
\centering
\hrule height 0.8pt \medskip

\noindent{\large \textbf{Temporal Order-to-Space Switching During Diffusion}}\par
\bigskip

\sectionHeader{Case 1}

\vspace{0.5em}

\noindent
\StepBlock{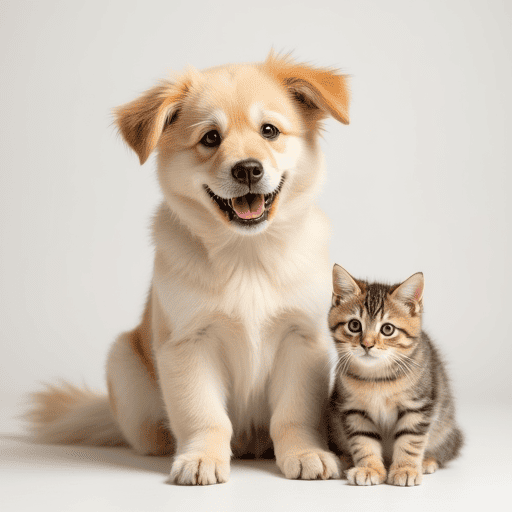}{Baseline}\hfill%
\StepBlock{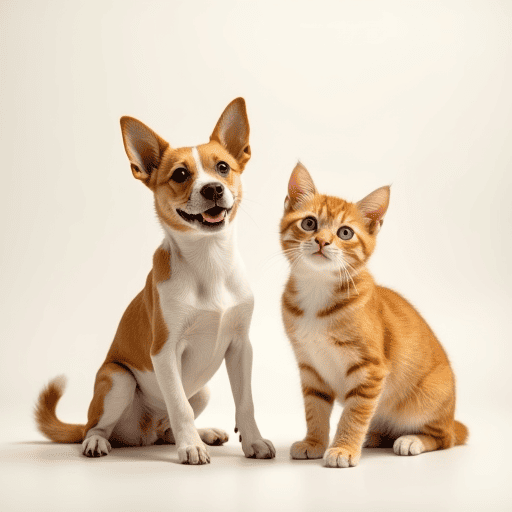}{$s01$}\hfill%
\StepBlock{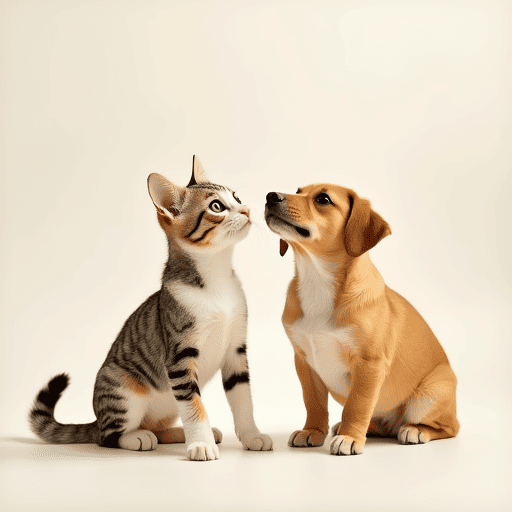}{$s02$}\hfill%
\StepBlock{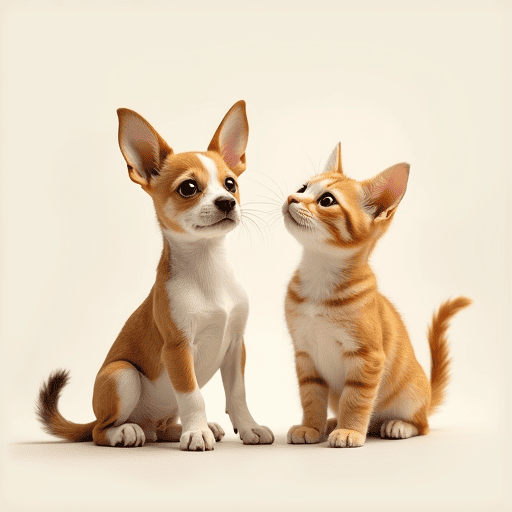}{$s03$}\hfill%
\StepBlock{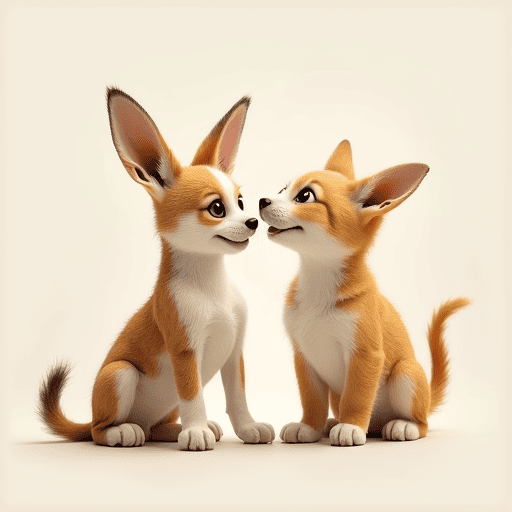}{$s04$}

\bigskip

\sectionHeader{Case 2}

\vspace{0.5em}

\noindent
\StepBlock{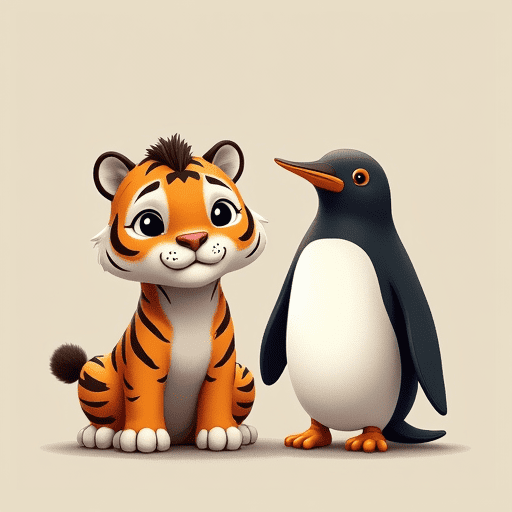}{Baseline}\hfill%
\StepBlock{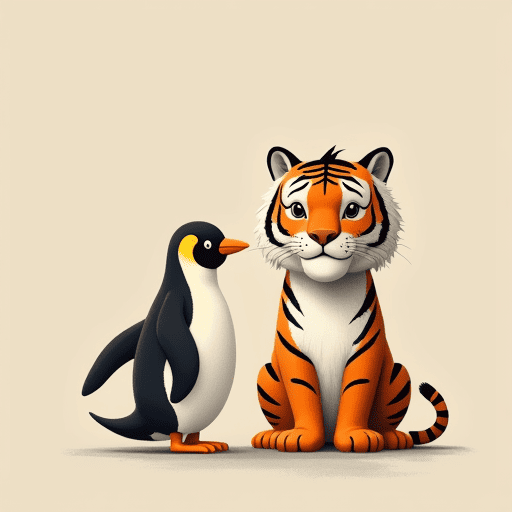}{$s01$}\hfill%
\StepBlock{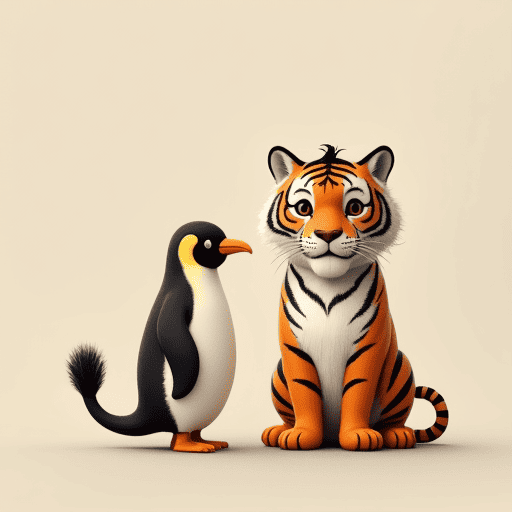}{$s02$}\hfill%
\StepBlock{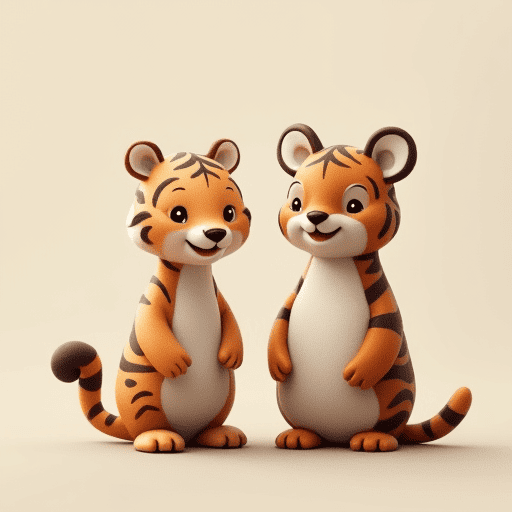}{$s03$}\hfill%
\StepBlock{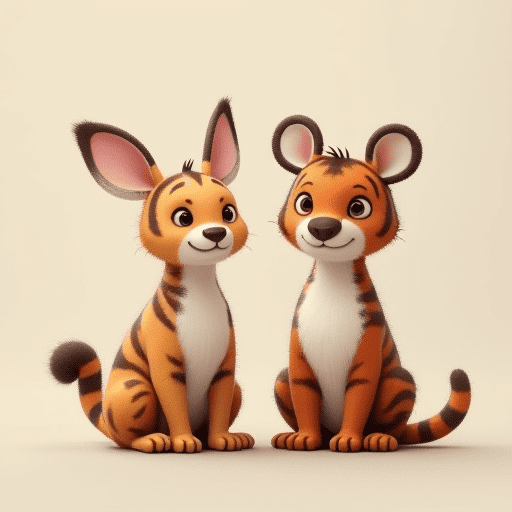}{$s04$}

\bigskip

\sectionHeader{Case 3}

\vspace{0.5em}

\noindent
\StepBlock{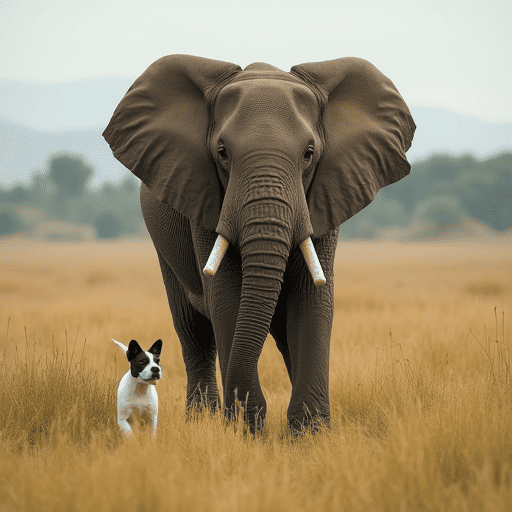}{Baseline}\hfill%
\StepBlock{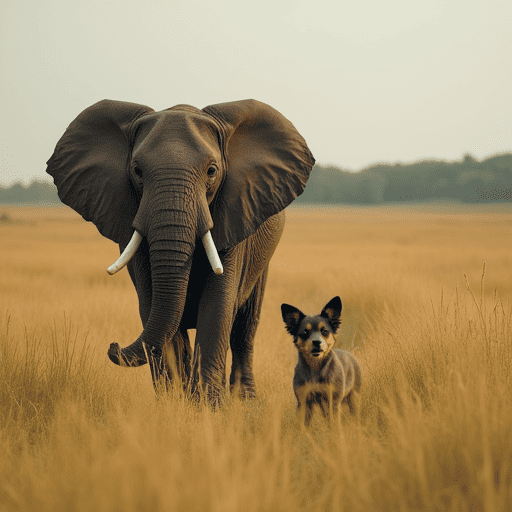}{$s01$}\hfill%
\StepBlock{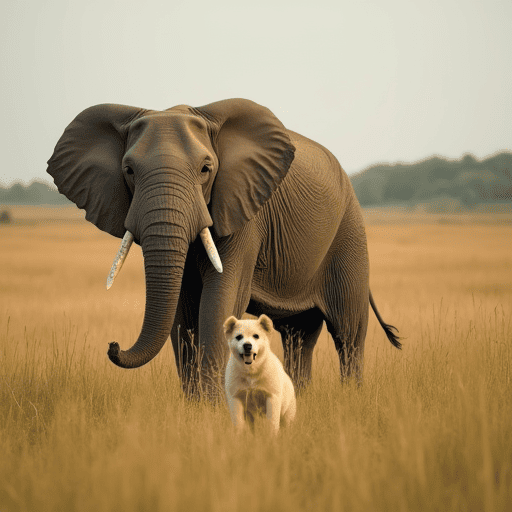}{$s02$}\hfill%
\StepBlock{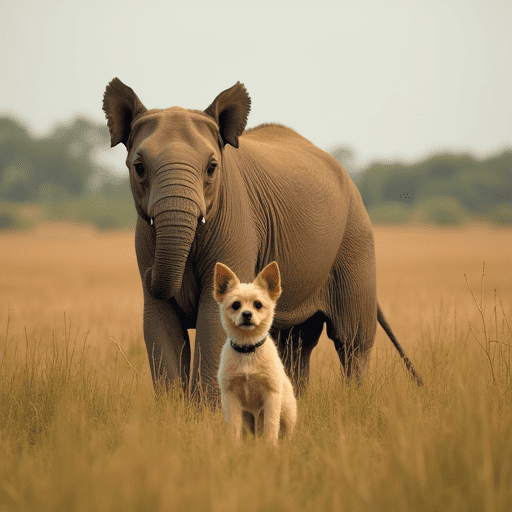}{$s03$}\hfill%
\StepBlock{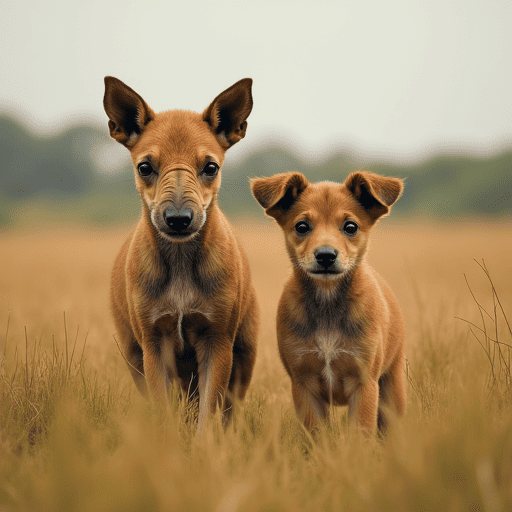}{$s04$}

\medskip \hrule height 0.8pt

\caption{Temporal intervention on order-sensitive conditioning during diffusion sampling.
We switch entity order from a neutral prompt to a rich prompt at different denoising steps.
Case~1 and Case~2 illustrate that order cues injected during the early, layout-sensitive stage
can reconfigure spatial assignment, while Case~3 shows unstable behavior when switching
near the boundary of layout formation.}

\label{fig:step_switching_examples}
\end{figure*}

\end{document}